\documentclass{article}
\PassOptionsToPackage{numbers, compress}{natbib}
\usepackage[final]{neurips_2024}
\usepackage{microtype}
\usepackage{graphicx}
\usepackage{booktabs} 
\usepackage{hyperref}

\usepackage{mathtools}
\usepackage[utf8]{inputenc} 
\usepackage[T1]{fontenc}    
\usepackage{url}            
\usepackage{amsfonts}       
\usepackage{nicefrac}       
\usepackage{xcolor}         
\usepackage{colortbl}
\usepackage{subcaption}
\usepackage{enumitem}
\usepackage{multirow}
\usepackage{capt-of}
\usepackage{wrapfig}
\usepackage{makecell}
\usepackage{bm}
\usepackage{arydshln}
\usepackage{placeins}
\definecolor{Blue9}{rgb}{0.1,0.3,0.95}
\hypersetup{
    colorlinks = true,
    citecolor = Blue9,
    linkcolor = Blue9
}
\usepackage{comment}
\usepackage{algorithm}
\usepackage{algorithmic}
\usepackage{eqparbox}
\usepackage{amsmath, amsfonts, amsmath, amssymb, amsthm, bm}
\usepackage{pifont} 

\theoremstyle{plain}

\theoremstyle{definition}

\theoremstyle{remark}

\renewcommand{\algorithmiccomment}[1]{\hfill{ $\rhd$ #1}}

\usepackage{xspace}

\usepackage{adjustbox}
\usepackage{listings} 
\usepackage{courier}
\definecolor{gg}{gray}{0.92}
\newcolumntype{a}{>{\columncolor{gg}}c}
\definecolor{figred}{RGB}{234, 21, 0}
\definecolor{figblue}{RGB}{48, 132, 194}
\definecolor{figgreen}{RGB}{99, 154, 63}

\usepackage[capitalize,noabbrev]{cleveref}
\usepackage[textsize=tiny]{todonotes}
\usepackage[hang,flushmargin]{footmisc} 
\newcommand{\metabbr}{MuDI\xspace}

\title{Identity Decoupling for Multi-Subject \\ Personalization of Text-to-Image Models}
\author{
    Sangwon Jang$^{\ast,1}$, \quad Jaehyeong Jo$^{\ast,1}$, \quad Kimin Lee$^{\dagger,1}$, \quad Sung Ju Hwang$^{\dagger,1,2}$ \\[3pt]
    $^{\ast}$Equal contribution \quad\quad\enspace $^{\dagger}$Equal advising  \\[3pt]
    KAIST$^{1}$, \enspace DeepAuto.ai$^{2}$ \\[3pt]
    \texttt{\{ sangwon.jang, harryjo97, kiminlee, sjhwang82 \}@kaist.ac.kr}
}

\begin{document}
\maketitle

\begin{figure*}[h!]
\vspace{-0.35in} 
\centering
    \includegraphics[width=0.95\linewidth]{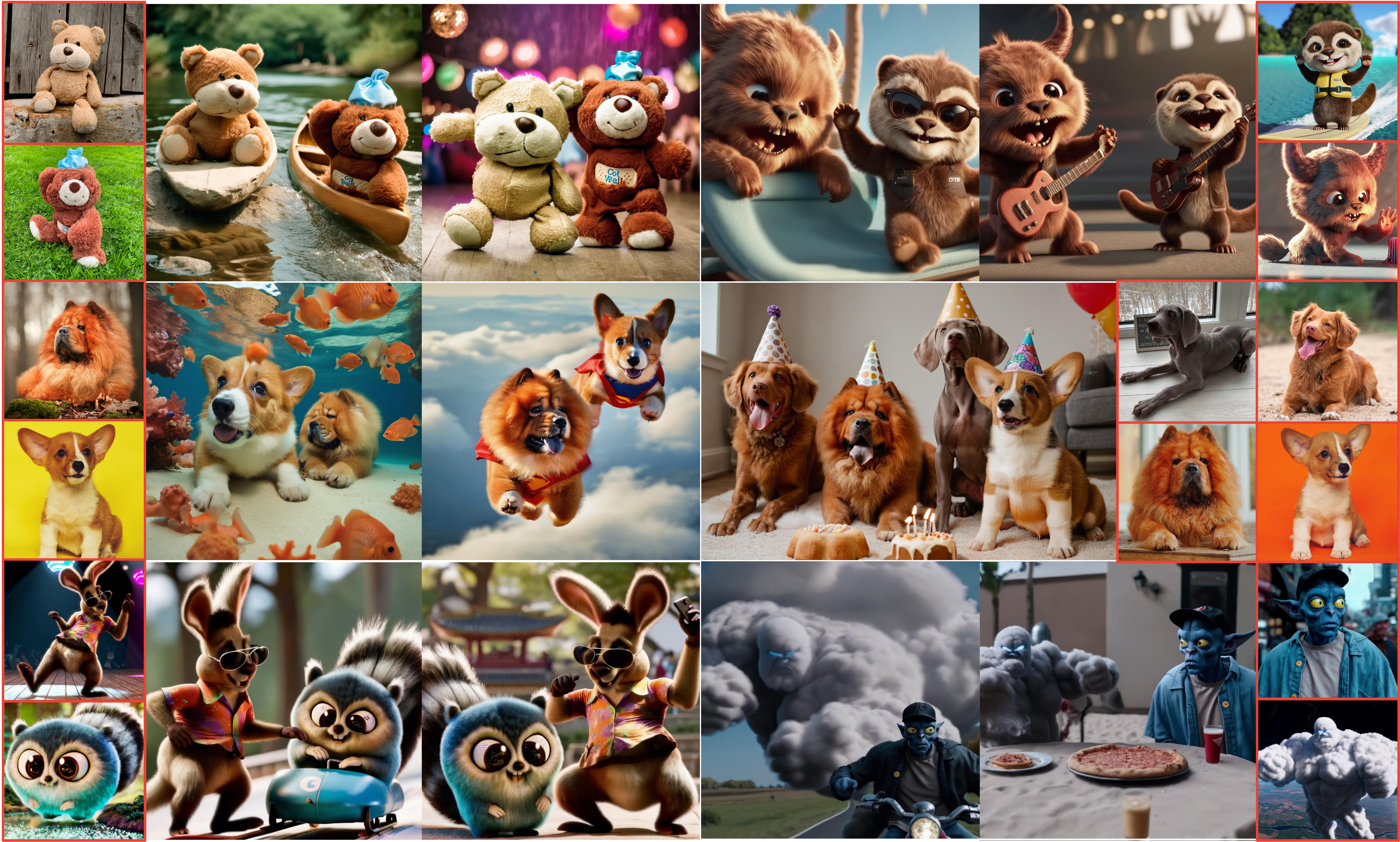}
\vspace{-0.05in} 
\caption{Given a few images of multiple subjects (red boxes), \metabbr can personalize a text-to-image model (e.g., SDXL~\citep{podell2023sdxl}) to generate multi-subject images without identity mixing. 
Some reference images (e.g., Cloud Man and Blue Alien) are created by Sora~\citep{Sora}, introducing novel concepts not previously encountered by SDXL.} 
    \label{fig:fig1}
\vspace{-0.1in} 
\end{figure*}

\begin{abstract}
Text-to-image diffusion models have shown remarkable success in generating personalized subjects based on a few reference images. However, current methods often fail when generating multiple subjects simultaneously, resulting in mixed identities with combined attributes from different subjects. In this work, we present \metabbr, a novel framework that enables multi-subject personalization by effectively decoupling identities from multiple subjects.
Our main idea is to utilize segmented subjects generated by a foundation model for segmentation (Segment Anything) for both training and inference, as a form of data augmentation for training and initialization for the generation process.
Moreover, we further introduce a new metric to better evaluate the performance of our method on multi-subject personalization. Experimental results show that our \metabbr can produce high-quality personalized images without identity mixing, even for highly similar subjects as shown in \autoref{fig:fig1}. Specifically, in human evaluation, MuDI obtains twice the success rate for personalizing multiple subjects without identity mixing over existing baselines and is preferred over 70\% against the strongest baseline. Our project page is at \url{https://mudi-t2i.github.io/}.
\end{abstract}
\section{Introduction}
\vspace{-0.05in}
\begin{figure*}[t!]
\vspace{-0.1in}
\centering
    \includegraphics[width=0.98\linewidth]{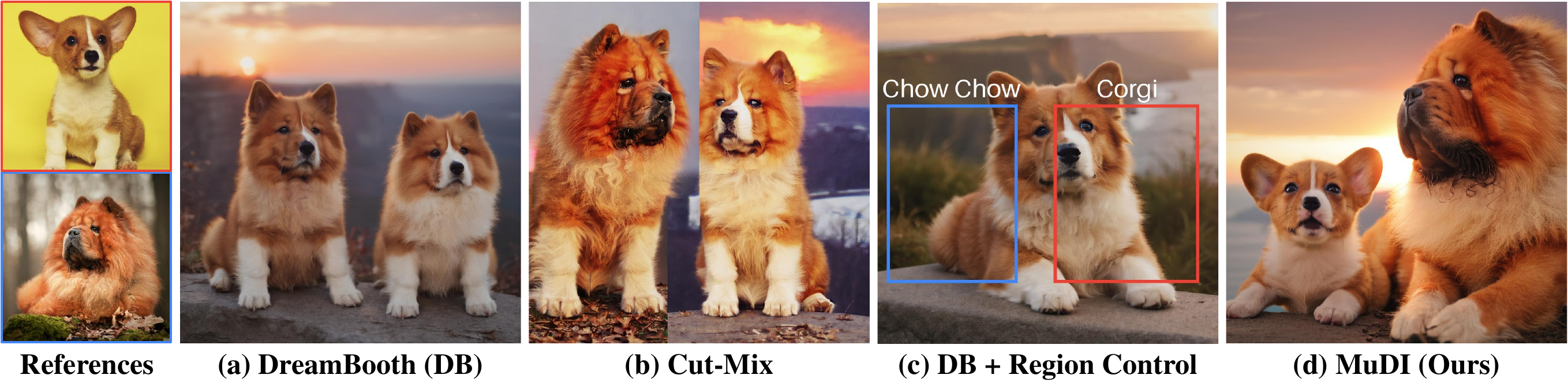}
\vspace{-0.05in}
    \caption{\textbf{Comparison of multi-subject personalization methods using Corgi and Chow Chow images (red boxes) using SDXL~\citep{podell2023sdxl}}. 
    DreamBooth~\citep{ruiz2023dreambooth} produces mixed identity dogs, such as a Corgi with Chow Chow ears\protect\footnotemark.
    Cut-Mix~\citep{han2023svdiff} often generates artifacts like unnatural vertical lines.
    Additionally, using layout conditioning like region control~\citep{gu2024mix} proves ineffective in preventing identity blending in recent advanced diffusion models such as SDXL.
    In contrast, ours successfully personalizes each dog, avoiding identity mixing and artifacts observed in prior methods.}
    \label{fig:fig2}
\vspace{-0.1in}
\end{figure*}
\footnotetext{Custom Diffusion~\citep{kumari2023multi} also results in identity mixing and we provide the examples in \autoref{fig:custom}.}

Text-to-image diffusion models, trained on large datasets of image and text pairs, have shown great success in generating high-quality images for given text prompts~\citep{ramesh2022hierarchical, saharia2022photorealistic, rombach2022ldm, betker2023improving}.
Building on this success, there is a growing interest in personalizing these text-to-image models. 
Specifically, given a few images of a single user-defined subject, several methods have been developed to enable these models to generate images of the subject in novel contexts~\citep{gal2022image, ruiz2023dreambooth, voynov2023p+, kumari2023multi}.
Furthermore, these personalization methods have been expanded to include the customization of style~\citep{sohn2023styledrop}, background~\citep{tang2023realfill}, and activity~\citep{huang2023action}, offering even greater flexibility and creativity in image generation.

Despite significant progress in personalizing text-to-image models for single subjects, current methods often struggle to handle multiple subjects simultaneously~\citep{kumari2023multi,han2023svdiff}.
While successful in rendering each subject individually, these methods suffer from identity mixing during the composition of subjects. 
For instance, as shown in \autoref{fig:fig2}(a), recent works, such as DreamBooth~\citep{ruiz2023dreambooth}, generate images with mixed identities when applied to two dogs.
The problem of identity mixing becomes more pronounced with semantically similar subjects that share attributes, such as colors or textures, which leads to greater confusion in maintaining distinct identities.

To address identity mixing in multi-subject personalization, \citet{han2023svdiff} proposed to use Cut-Mix~\citep{yun2019cutmix}, an augmentation technique that presents the models with cut-and-mixed images of the subjects during personalization.
However, using Cut-Mix-like images inevitably often results in the generation of unnatural images with stitching artifacts, such as vertical lines that separate the subjects.
Moreover, Cut-mix remains unsuccessful in decoupling similar subjects (see \autoref{fig:fig2}(b)).
There are alternative approaches~\citep{chen2023anydoor, liu2023cones2, gu2024mix} that rely on pre-defined conditioning, e.g., bounding boxes or ControlNet~\citep{zhang2023controlnet} to separate the identities spatially.
However, such auxiliary inputs like sketch~\citep{zhang2023controlnet} could be difficult to obtain, and we have observed that the layout conditioning is ineffective for recent diffusion models such as SDXL~\citep{podell2023sdxl} (see \autoref{fig:fig2}(c) and Appendix~\ref{app:add:region}).

In this work, we propose \metabbr, a multi-subject personalization framework that effectively addresses identity mixing, even for highly similar subjects. 
Our key idea is to leverage the segmented subjects obtained by a foundation model for image segmentation (Segment Anything Model (SAM)~\citep{kirillov2023sam}), enabling the decoupling of the identities among different subjects.
Specifically, we extract segmentation maps of the user-provided subjects using SAM and utilize them for both training and inference.
For training, we introduce a data augmentation method that randomly composes segmented subjects, which allows efficient personalization by removing identity-irrelevant information.
Additionally, we utilize the segmented subjects to initialize the generation process.
Instead of starting from Gaussian noise, we begin with a mean-shifted random noise created from segmented subjects. 
We find that this provides a helpful hint for the model to separate the identities and further reduces subject missing during generation.
Notably, our approach significantly mitigates identity mixing as shown in \autoref{fig:fig2}, without relying on preset auxiliary conditions such as bounding boxes or sketches.

We evaluate the effectiveness of the proposed framework using a new dataset composed of subjects prone to identity mixing, which includes a diverse range of categories from animals to objects and scenes. 
To facilitate this evaluation, we introduce a new metric specifically designed to assess the fidelity of multiple subjects in the images, taking into account the degree of identity mixing.
In our experiments, \metabbr successfully personalizes the subjects without mixed identities, significantly outperforming DreamBooth~\citep{ruiz2023dreambooth}, Cut-Mix~\citep{han2023svdiff}, and Textual Inversion~\citep{gal2022image}, in both qualitative and quantitative comparisons. 
Further human study with side-by-side comparisons of \metabbr over other methods shows that human raters prefer our method by more than 70\% over the strongest baseline.

\section{Related work}

\vspace{-0.05in}
\paragraph{Text-to-image personalization}
Personalized text-to-image diffusion models have shown impressive abilities to render a single user-specific subject in novel contexts from only a few images. Two representative classes of personalization methods have been proposed by \citet{gal2022image} and \citet{ruiz2023dreambooth}. Textual Inversion~\citep{gal2022image} optimizes new text embedding for representing the specified subjects and has been improved for learning on extended embedding spaces~\citep{voynov2023p+,alaluf2023neti}. On the other hand, DreamBooth~\citep{ruiz2023dreambooth} fine-tunes the weights of the pre-trained model to bind new concepts with unique identifiers and has been developed by recent works~\citep{kumari2023multi,han2023svdiff,tewel2023perfusion} for efficiently fine-tuning the models.

However, existing methods fall short of synthesizing multiple user-defined subjects together, suffering from identity mixing.
\citet{han2023svdiff} introduce Cut-Mix to address identity mixing by augmenting Cut-Mix-like images during training but fail to separate similar subjects and generates stitching artifacts.
Other lines of work~\citep{liu2023cones2, gu2024mix} compose personalized subjects using layout conditioning, which manipulates cross-attention maps with user-defined locations.
Yet such conditioning based on cross-attention maps is ineffective for recent diffusion models such as SDXL~\citep{podell2023sdxl}. 
In this work, we develop a novel framework that allows the personalization of multiple subjects without identity mixing even for subjects with similar appearances.

\vspace{-0.05in}
\paragraph{Modular customization}
Recent works~\citep{kumari2023multi,gu2024mix,po2023orthogonal} explore a different scenario for personalizing multiple subjects, namely \emph{modular customization}, where the subjects are independently learned by models and users mix and match the subjects during inference to compose them. 
Custom Diffusion~\citep{kumari2023multi} merges individually fine-tuned models by solving constrained optimization and Mix-of-Show~\citep{gu2024mix} introduces gradient fusion to merge single-concept LoRAs~\citep{hu2021lora}. When handling multiple subjects, these works also suffer from identity mixing, and they rely on preset spatial conditions such as ControlNet~\citep{zhang2023controlnet} and region control~\citep{gu2024mix} to address the problem. 
Notably, our method can be applied to this scenario to decouple subjects' identities without using such conditions.

\section{Preliminaries}
\vspace{-0.05in}

\paragraph{Text-to-image diffusion models}
Diffusion models~\cite{ho2020denoising,song2021sde} generate samples from noise by learning to reverse the perturbation, i.e., denoise, which can be modeled by a diffusion process. 
To be specific, at each step of the diffusion, the model predicts the random noise $\bm{\epsilon}\sim\mathcal{N}(0,\mathbf{I})$ that has been used to corrupt the sample. 
Text-to-image diffusion models~\citep{saharia2022photorealistic, rombach2022ldm} incorporate text conditions for the generation. 
Given the dataset $\mathcal{D}$ consisting of the image-text pairs $(\bm{x},\bm{c})$, text-to-image diffusion models parameterized by the noise prediction model $\bm{\epsilon}_{\theta}$ can be trained with the following objective:
\begin{align}
    \mathcal{L}_{DM}(\theta; \mathcal{D}) = \mathbb{E}_{(\bm{x},\bm{c})\sim\mathcal{D}, \bm{\epsilon}\sim\mathcal{N}(0,\mathbf{I}), t\sim\mathcal{U}(0,T)}\left[ \big\| \bm{\epsilon}_{\theta}(\bm{x}_t;\bm{c},t) - \bm{\epsilon} \big\|^2_2 \right], 
    \label{eq:diffusion}
\end{align}
where $\bm{\epsilon}$ is the random noise, time $t$ is sampled from the uniform distribution $\mathcal{U}(0,T)$, and $\bm{x}_t=\alpha_t\bm{x}+\sigma_t\bm{\epsilon}$ for the coefficients $\alpha_t$ and $\sigma_t$ that determine the noise schedule of the diffusion process.

\paragraph{Personalizing text-to-image models}
Given a few images of a single specific subject, DreamBooth~\citep{ruiz2023dreambooth} fine-tunes the weights of the diffusion model with a unique identifier for the subject, i.e., "a [identifier] [class noun]". 
The model weights are updated to learn the subject while preserving the visual prior, which can be achieved by minimizing the objective:
\begin{align}
    \mathcal{L}_{DB}(\theta) = \underbrace{\mathcal{L}_{DM}(\theta; \mathcal{D}_{ref})}_{\text{personalization loss}} + \lambda\underbrace{\mathcal{L}_{DM}(\theta; \mathcal{D}_{prior})}_{\text{prior preservation loss}} ,
    \label{eq:dreambooth}
\end{align}
where $\mathcal{L}_{DM}$ is the loss defined in Eq.~\eqref{eq:diffusion}, $\mathcal{D}_{ref}$ is the dataset consisting of reference images of the subject, $\mathcal{D}_{prior}$ is the dataset consisting of class-specific prior images, and $\lambda$ is a coefficient for the prior preservation loss. 
Similar to personalizing a single subject, existing works~\citep{kumari2023multi, han2023svdiff} jointly train for multiple subjects by combining the images from the set of user-specified subjects to construct $\mathcal{D}_{ref}$ and using different identifiers for each subject.

\begin{figure*}[t!]
\vspace{-0.1in}
\centering
    \includegraphics[width=1\linewidth]{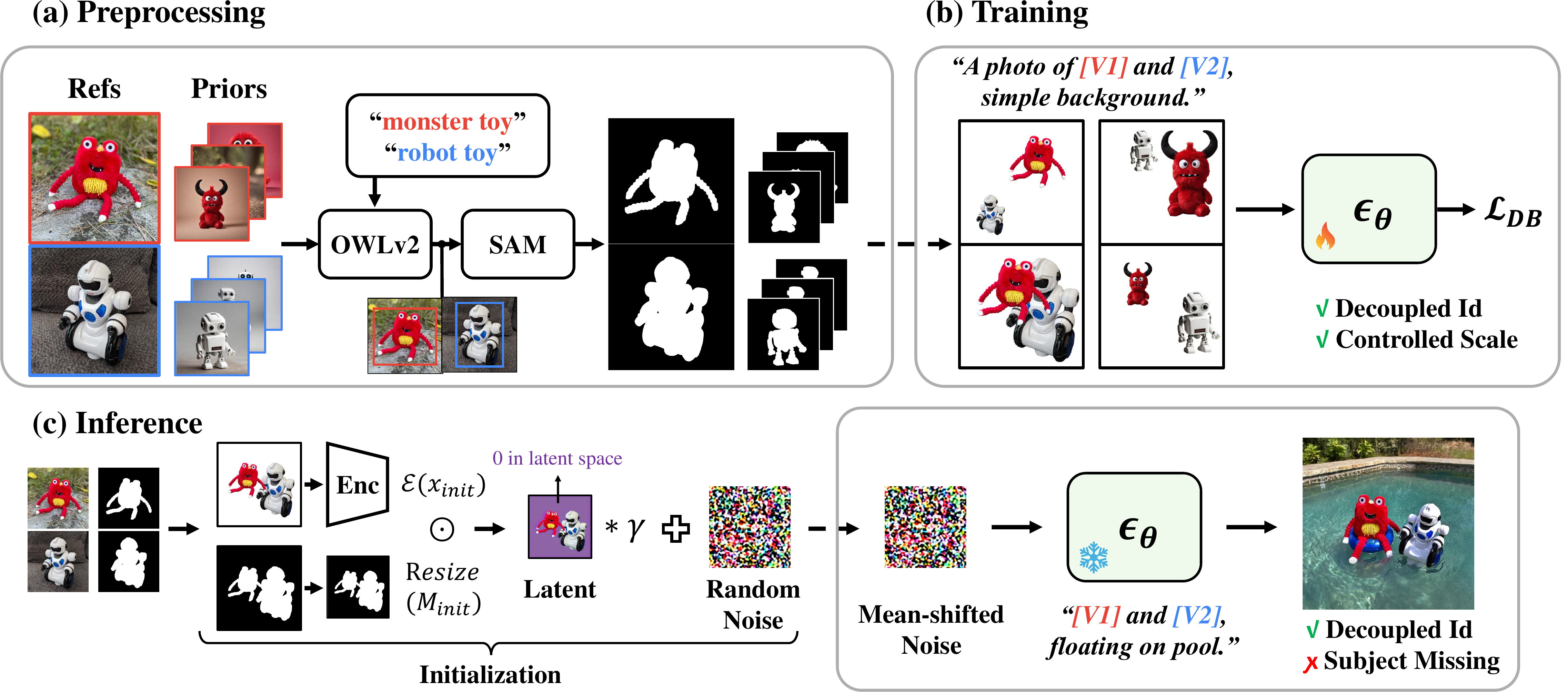}
\vspace{-0.15in} 
    \caption{\textbf{Overview of \metabbr}. 
    (a) We automatically obtain segmented subjects using SAM~\citep{kirillov2023sam} and OWLv2~\citep{minderer2024owlv2} in the preprocessing stage. 
    (b) We augment the training data by randomly positioning segmented subjects with controllable scales to train the diffusion model $\epsilon_{\theta}$. We refer to this data augmentation method as Seg-Mix. 
    (c) We initialize the generation process with mean-shifted noise created from segmented subjects, which provides a signal for separating identities without missing.}
    \label{fig:method}
\vspace{-0.1in}
\end{figure*}

\section{\metabbr: Multi-subject personalization for decoupled identities}
\vspace{-0.05in}
In this section, we present \metabbr: \textbf{Mu}lti-subject personalization for \textbf{D}ecoupled \textbf{I}dentities, which leverages segmented subjects to separate identities. 
In Section~\ref{method:train}, we describe our training method, which augments training data through random compositions of segmented subjects.
We also introduce a simple inference method that initializes noise for sample generation based on subject segmentation in Section~\ref{method:inference}. 
Finally, we present a new metric to evaluate the multi-subject fidelity in Section~\ref{method:metric}.

\subsection{Training} \label{method:train}

\vspace{-0.05in}
\paragraph{Personalization with augmentation}
To address identity mixing in multi-subject personalization, we introduce a new data augmentation method for training the pre-trained text-to-image model called \emph{Seg-Mix}.
We aim to mitigate identity mixing by leveraging segmented subjects during personalizing text-to-image models. 
By isolating each subject from the background, Seg-Mix enables the model to learn to distinguish between different identities effectively.
We integrate Seg-Mix with DreamBooth~\citep{ruiz2023dreambooth}, which personalizes text-to-image models using unique identifiers (see Eq.~\eqref{eq:dreambooth}).

To implement Seg-Mix, we preprocess reference images by automatically extracting segmentation maps of user-provided subjects using the Segment Anything Model (SAM)~\citep{kirillov2023sam}.
Specifically, this process begins with the extraction of subject bounding boxes using the OWLv2~\citep{minderer2024owlv2}, an object detection model with an open vocabulary.
Subsequently, SAM segments the subjects based on these bounding boxes, as illustrated in \autoref{fig:method}(a).
After the preprocessing step, we create augmented images by randomly positioning the resized segmented subjects, as illustrated in \autoref{fig:method}(b). 
We provide the detailed procedures of our method in \autoref{alg:seg-mix}.
These augmented images are paired with a simple prompt "A photo of [$V_1$] and [$V_2$], simple background.", which is designed to explicitly remove identity-irrelevant information.
We also apply this augmentation to the prior dataset by creating images from segmented prior subjects.
Using augmented datasets, we fine-tune text-to-image models based on the DreamBooth objective function in Eq.\eqref{eq:dreambooth}.

One of the key advantages of Seg-Mix is its ability to train models without identity-irrelevant artifacts, due to the removal of backgrounds.
This process also mitigates unnatural artifacts, such as stitching artifacts observed in previous methods like Cut-Mix~\citep{han2023svdiff}.
Moreover, by allowing subjects to overlap during Seg-Mix, which is different from Cut-Mix, we prevent attributes from leaking to neighboring identities and enhance interactions among the subjects, which we analyze in Appendix~\ref{app:add:overlap}.

\vspace{-0.05in}
\paragraph{Descriptive class}
Intuitively, for two similar subjects in the same category, separating them solely with unique identifiers is a challenging task and prone to identity mixing. 
In single-subject personalization, \citet{chae2023instructbooth} observed that adding detailed descriptions in front of the class nouns helps in capturing the visual characteristics of rare subjects.
Inspired by this observation, we adopt specific class nouns (e.g., Weimaraner instead of dog) or add detailed descriptions in front of general class nouns (e.g., white robot toy instead of toy).
Instead of manually selecting appropriate classes, we leverage GPT4-v~\citep{gpt-4v} to automatically obtain these specific class nouns or descriptions.
We empirically validate that this simple modification improves the preservation of the details for multiple subjects leading to the decoupling of the identities of highly similar subjects.

\begin{figure*}[t!]
\vspace{-0.1in}
\centering
\begin{minipage}{0.67\linewidth}
    \includegraphics[width=1.0\linewidth]{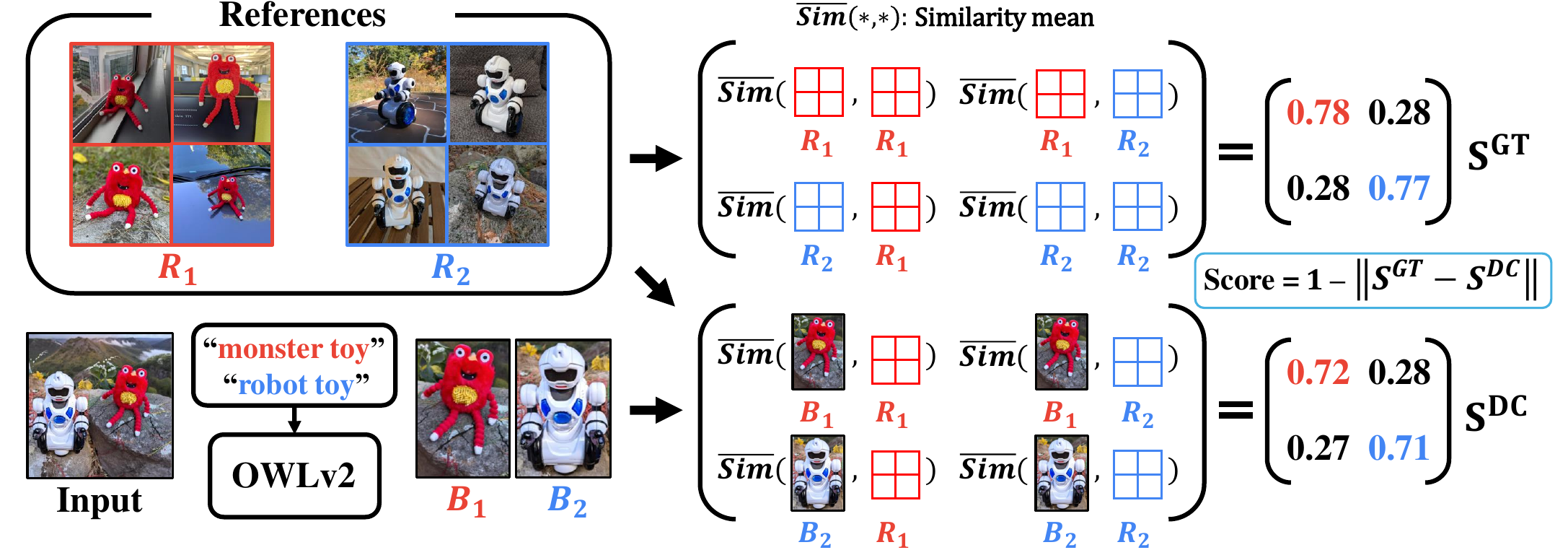}
\end{minipage}
\hfill
\begin{minipage}{0.32\linewidth}
    \resizebox{\linewidth}{!}{
\renewcommand{\arraystretch}{1.3}
\renewcommand{\tabcolsep}{2pt}
\begin{tabular}{l cc cc}
\toprule
Metrics & Spearman & AUROC \\
\midrule
DS~\citep{fu2023dreamsim} & 0.219 & 0.606 \\
DINOv2~\citep{oquab2023dinov2} & 0.156 & 0.559 \\
D\&C-DS & \textbf{0.618} & \textbf{0.850} \\
D\&C-DINO & 0.445 & 0.741 \\
\bottomrule
\end{tabular}}
\end{minipage}
\caption{\textbf{(Left) Overview of Detect-and-Compare}. 
We calculate the mean similarities between detected subjects and reference images to evaluate multi-subject fidelity. Specifically, we compare $\bm{S}^{GT}$ and $\bm{S}^{DC}$. We provide pseudo-code in \autoref{alg:dnc}.
\textbf{(Right) Correlation between metrics and human evaluation}. We report the Spearman's rank correlation coefficient and AUROC.}  
\label{fig:dnc_overview}
\vspace{-0.1in}
\end{figure*}

\subsection{Inference} \label{method:inference}
\vspace{-0.05in}
It has been observed that initial noises for the generation affect the overall quality of generated images~\citep{mao2023initial,samuel2024rare}, a finding that holds for personalized models with Seg-Mix as well.
Motivated by this observation, we propose a novel inference method to improve identity decoupling without additional training or computational overhead.
As illustrated in \autoref{fig:method}(c), we first create an image $\bm{x}_{init}$ of segmented subjects following Seg-Mix and extract its latent embedding from VAE encoder $\mathcal{E}$. 
We then add this latent embedding to a random Gaussian noise $\bm{\epsilon}$, scaled by a coefficient $\gamma$ as $\bm{z}_T = (\mathcal{E}(\bm{x}_{init}) \odot Resize(\bm{M}_{init})) * \gamma + \bm{\epsilon}$ where $Resize(\bm{M}_{init})$ denotes the resized version of segmentation mask $\bm{M}_{init}$.
This \emph{mean-shifted} noise $\bm{z}_T$ encodes coarse information about the subjects and their layout, serving as a good starting point in sample generation. 
We analyze the effect of coefficient $\gamma$ in Appendix~\ref{app:add:gamma} and validate the diversity of generated images from the initialization in Appendix~\ref{app:add:diversity}.
The proposed inference method is summarized in  \autoref{alg:seg-mix-init}. 
Additionally, instead of using randomly composed initial latent, we explore utilizing Large Language Models (LLMs)~\citep{Cho2023VisualP} to generate the layouts of bounding boxes for each subject aligned with the given prompt.
Such an LLM-guided initialization enhances the ability to render complex interactions between subjects (see \autoref{fig:appendix_LLM} in Appendix~\ref{app:samples:llm} for supporting results).

We remark that our initialization method also addresses the issue of subject dominance~\citep{tunanyan2023zero}, where certain subject dominates the generation while other subjects are ignored. 
By providing information through the initial composition, our inference method guides the model to consider all subjects without additional computation.
In Section~\ref{exp:ablation} and Appendix~\ref{app:add:init}, we validate that our inference method alleviates subject dominance, playing a crucial role when rendering many subjects simultaneously.

\subsection{New metric for multi-subject fidelity \label{method:metric}}
\vspace{-0.05in}
Existing metrics designed for measuring subject fidelity, such as CLIP-I~\citep{radford2021clip} or DINOv2~\citep{oquab2023dinov2}, are not suitable for evaluating multiple subjects because they do not account for identity mixing. 
Therefore, we introduce a new metric, called \emph{Detect-and-Compare} (D\&C), for evaluating multi-subject fidelity.

First, we utilize OWLv2~\citep{minderer2024owlv2} to detect the subjects in the generated image, with text queries as the supercategories of the subjects.
For the detected subjects $\{B_i\}^{N}_{i=1}$ and the reference subjects $\{R_j\}^{M}_{j=1}$, we construct similarity matrices by measuring the similarities between the subjects using subject fidelity metrics such as DreamSim~\citep{fu2023dreamsim} or DINOv2~\citep{oquab2023dinov2}.
Specifically, we first construct the \emph{D\&C similarities} matrix $\bm{S}^{DC}$, where $ij$-th entry represents the similarity between detected subject $B_i$ and reference $R_j$ (see \autoref{fig:dnc_overview}).
Similarly, we construct the \emph{ground-truth similarities} $\bm{S}^{GT}$, where $ij$-th entry represents the similarity between reference objects $R_i$ and $R_j$.
Since a mixed-identity subject yields high similarities to multiple references, we compare $\bm{S}^{DC}$ and $\bm{S}^{GT}$ to account for identity mixing.
Notably, the difference between $\bm{S}^{DC}$ and $\bm{S}^{GT}$ yields a matrix where diagonal entries denote similarities to the corresponding subject, while off-diagonal entries indicate similarities to other subjects which represent identity mixing. 
The closer $\bm{S}^{DC}$ is to $\bm{S}^{GT}$, the more accurately the detected subjects resemble the references, resulting in successful identity decoupling. 
Therefore, we define the \emph{D\&C score} as $1-\|\bm{S}^{GT}-\bm{S}^{DC}\|^2_F$.
We illustrate an overview of D\&C in \autoref{fig:dnc_overview}. 

To validate that our D\&C is capable of measuring multi-subject fidelity, we compare it with previous single-subject fidelity metrics extended to multi-subject settings.
These extended metrics compute the mean similarity to all the reference images~\citep{lui2023cones}. 
For 1600 images generated from various models, we assess the correlation between each metric and human evaluations using Spearman's rank correlation coefficient and the Area under the Receiver Operating Characteristic (AUROC) (see Appendix~\ref{app:exp:eval} for details). 
Table in \autoref{fig:dnc_overview} shows that D\&C with DreamSim (D\&C-DS) exhibits the highest correlation with human evaluation.
We provide qualitative examples in \autoref{fig:appendix_dnc_sample} and \autoref{fig:appendix_dnc_qualitative}.
These demonstrate that D\&C can capture multi-subject fidelity and is suitable as an evaluation metric.
\begin{figure*}[t!]
\vspace{-0.15in} 
\centering
    \includegraphics[width=1\linewidth]{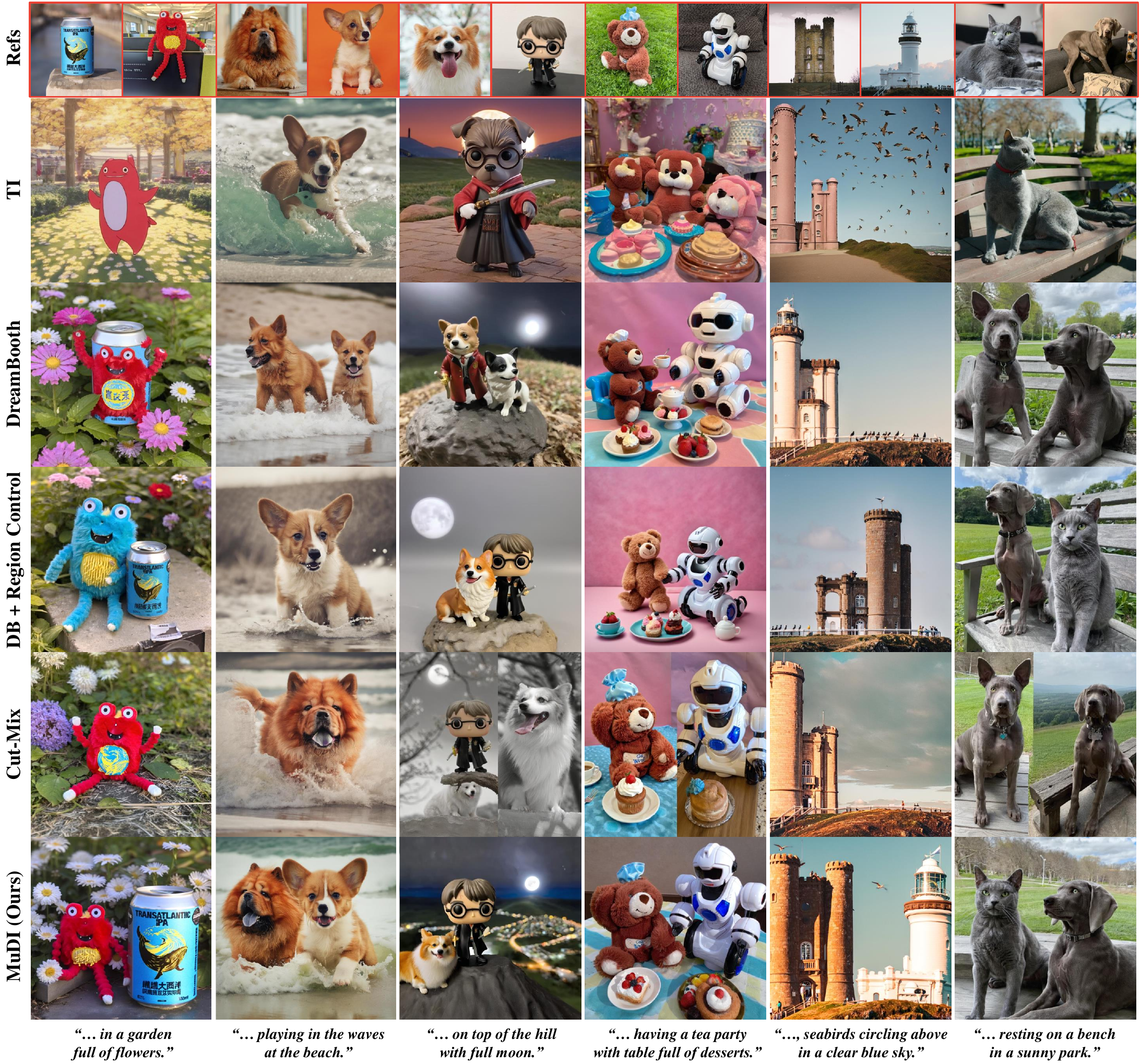}
\vspace{-0.2in} 
    \caption{\textbf{Qualitative comparison} of Textual Inversion (TI)~\citep{gal2022image}, DreamBooth (DB)~\citep{ruiz2023dreambooth}, DB with region control~\citep{gu2024mix}, Cut-Mix~\citep{han2023svdiff}, and \metabbr. Images in the same column are generated with the same random seed. We provide more examples in \autoref{fig:appendix_qual1}.}
    \label{fig:qual}
\vspace{-0.1in}
\end{figure*}

\vspace{-0.05in}
\section{Experiments}
\vspace{-0.05in}

\subsection{Experimental setup\label{exp:setup}}
\vspace{-0.05in}

\paragraph{Dataset\label{exp:dataset}}
We construct a new dataset to evaluate the performance of identity decoupling for multi-subject personalization methods. It consists of 8 pairs of similar subjects that are prone to identity mixing. 
We collected images from the DreamBench dataset~\citep{ruiz2023dreambooth} and the CustomConcept101 dataset~\citep{kumari2023multi}, consisting of diverse categories including animals, objects, and scenes.
For each pair of subjects, we generate 5 evaluation prompts using ChatGPT~\citep{chatgpt}, which describe scenes involving the subjects with simple actions and backgrounds. We provide more details of the dataset in Appendix~\ref{app:exp:dataset}.

\vspace{-0.05in}
\paragraph{Implementation details}
For all experiments, we use Stable Diffusion XL (SDXL)~\citep{podell2023sdxl} as the pre-trained text-to-image diffusion model and employ a LoRA~\citep{hu2021lora} with a rank of 32 for U-Net~\citep{ronneberger2015unet} module. 
We also present experiments with other Stable Diffusion models~\citep{rombach2022ldm} in Appendix~\ref{app:add:sdv2} and \ref{app:add:sdv1.5}.
For all methods, we pair the reference images with comprehensive captions obtained through GPT-4v~\citep{gpt-4v} which effectively mitigates overfitting to the background and shows better text alignment. 
We evaluate 400 generated images for each method, across 8 combinations with 5 evaluation prompts and 10 images of fixed random seeds. We provide more details in Appendix~\ref{app:exp:details}.

\vspace{-0.05in}
\paragraph{Baselines}
We evaluate our method against multi-subject personalization methods: {\em DreamBooth}~\citep{ruiz2023dreambooth}, DreamBooth with region control~\citep{gu2024mix}, DreamBooth using Cut-Mix~\citep{han2023svdiff} augmentation, namely {\em Cut-Mix}, and {\em Textual Inversion}~\citep{gal2022image}.
Note that we exclude Custom Diffusion~\citep{kumari2023multi} from the baselines due to its low quality when applied to SDXL (see Appendix~\ref{app:add:custom}).
For both Cut-Mix and Seg-Mix, we use a fixed augmentation probability of 0.3, and we do not use Unmix regularization~\citep{han2023svdiff} as it degrades the image quality for SDXL (see Appendix~\ref{app:add:ca}).
We describe further details in Appendix~\ref{app:exp:details}.

\subsection{Main results}

\vspace{-0.05in}
\paragraph{Qualitative comparison}
As shown in \autoref{fig:qual}, our approach successfully generates the subjects avoiding identity mixing, even for similar subjects such as two dogs (2nd column). 
On the contrary, DreamBooth results in mixed identities, and using region control proves ineffective for separating identities, as it seldom succeeds and frequently fails.
Cut-Mix also falls short of decoupling the identities while producing stitching artifacts. Textual Inversion fails to preserve the subjects' details.

\begin{figure*}[t!]
\vspace{-0.1in}
\caption{\textbf{(Left) Human evaluation results} on multi-subject fidelity and overall preference. \textbf{(Right) Quantitative results} on multi-subject fidelity and text fidelity. $\dagger$ denotes the text fidelity score considering the permutation of the subjects in the prompt to avoid position bias.}\label{fig:eval}
\vspace{-0.05in}
\begin{minipage}{0.46\linewidth}
    \includegraphics[width=1.0\linewidth]{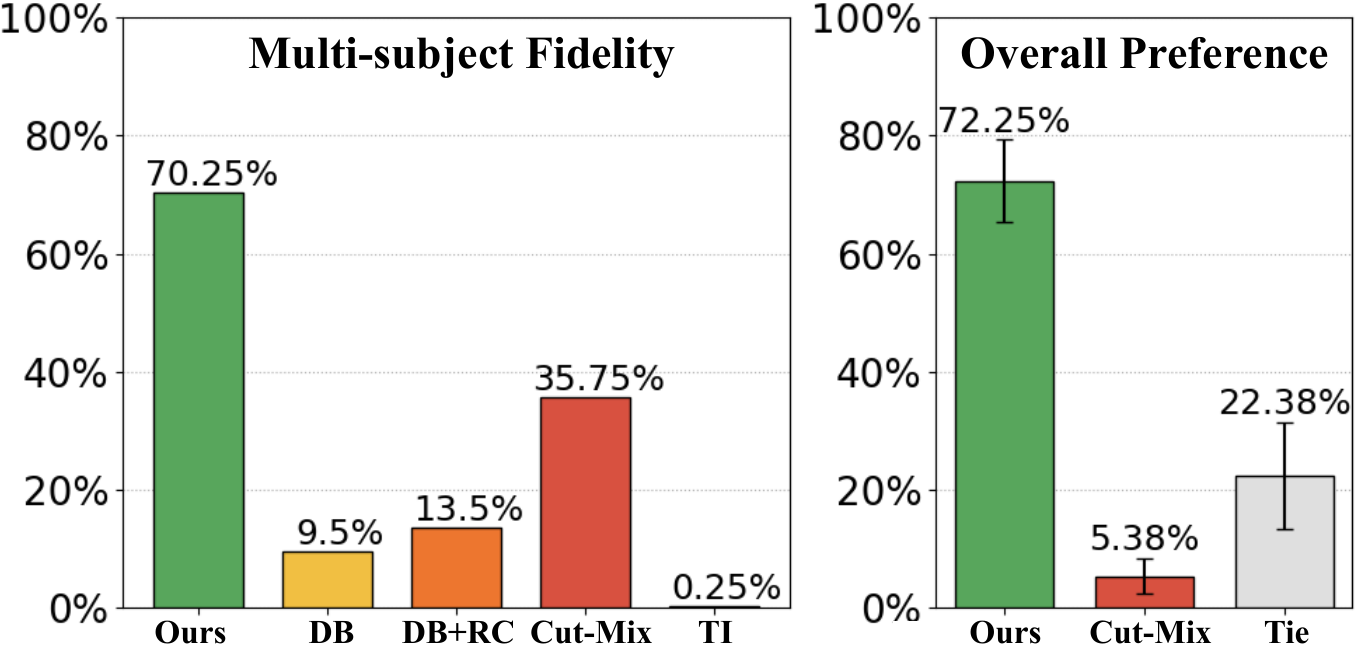}
\end{minipage}
\hfill
\begin{minipage}{0.53\linewidth}
    \centering
    \resizebox{\textwidth}{!}{
    \renewcommand{\arraystretch}{1.2}
    \renewcommand{\tabcolsep}{4pt}
\begin{tabular}{l c c c c}
\toprule
     & \multicolumn{2}{c}{Multi-Subject Fidelity} & \multicolumn{2}{c}{Text Fidelity} \\
\cmidrule(l{2pt}r{2pt}){2-3}
\cmidrule(l{2pt}r{2pt}){4-5}
    Method & D\&C-DS$\uparrow$ & D\&C-DINO$\uparrow$ & ImageReward$^\dagger$$\uparrow$ & CLIPs$^\dagger$$\uparrow$ \\
\midrule
    TI~\citep{gal2022image} & 0.116 & 0.132 & -0.149 & 0.227 \\
    DB~\citep{ruiz2023dreambooth} & 0.371 & 0.388 & 0.579 & 0.255 \\
    DB+Region~\citep{gu2024mix} & 0.340 & 0.379 & 0.349 & 0.245 \\
    Cut-Mix~\citep{han2023svdiff} & 0.432 & 0.460 & -0.287 & 0.225\\
\midrule
    \metabbr (Ours) & \textbf{0.637} & \textbf{0.610} & \textbf{0.770} & \textbf{0.263}\\
\bottomrule
\end{tabular}}


\end{minipage}
\vspace{-0.15in}
\end{figure*}

\begin{figure*}[t!]
\centering
    \includegraphics[width=0.9\linewidth]{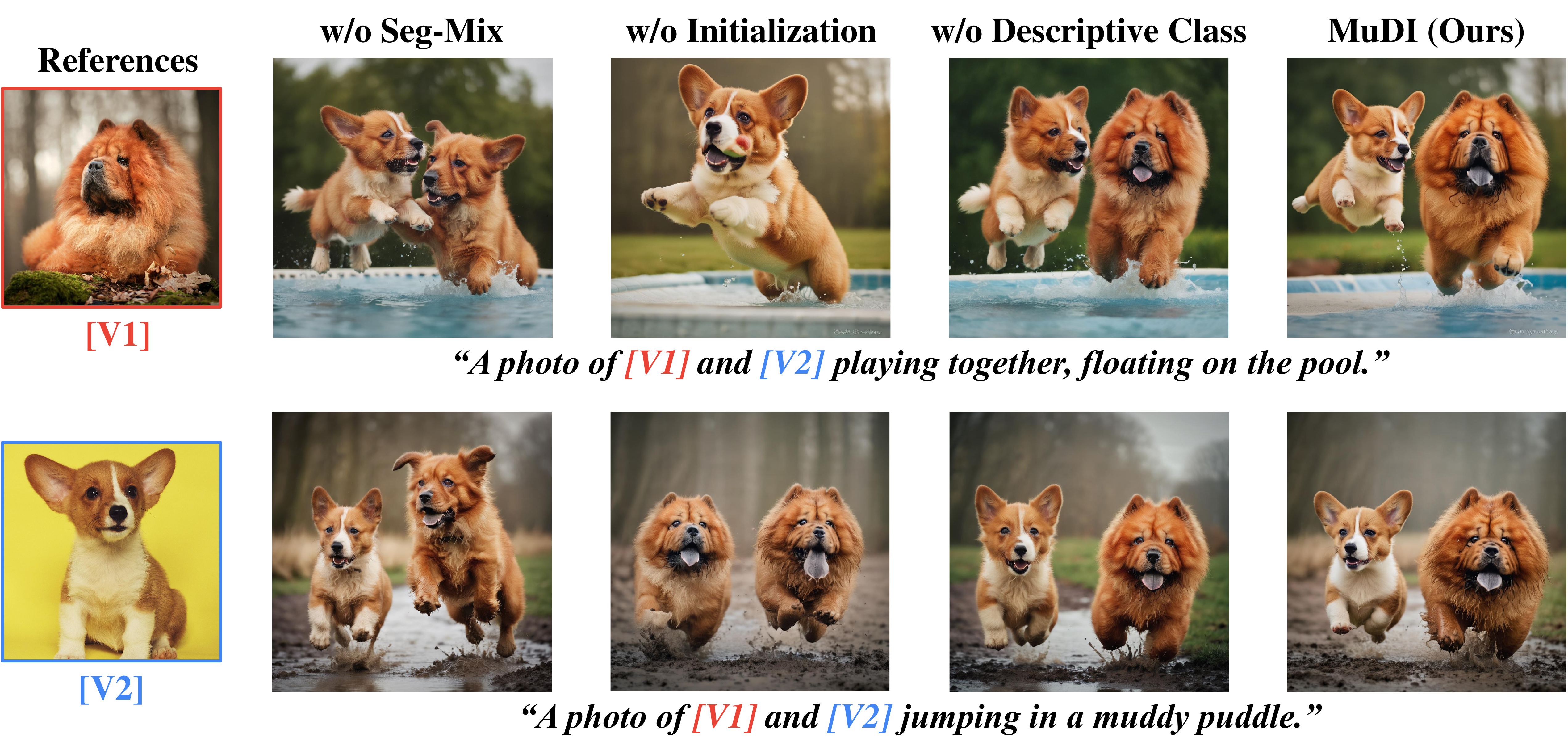}
\vspace{-0.1in}
    \caption{\textbf{Ablation Studies} on \metabbr. While our method successfully personalizes Corgi and Chow Chow, ablating Seg-Mix results in mixed identity dogs. Inference without our initialization generates images of the subject missing. Training without descriptive class fails to catch subject details.}
    \label{fig:ablation}
\vspace{-0.1in}
\end{figure*}

\vspace{-0.05in}
\paragraph{Human evaluation}
We conduct human evaluations to assess the quality of images generated by the baselines and our method. We first ask human raters to evaluate the multi-subject fidelity via binary feedback.
Additionally, we provide reference images of each subject along with two anonymized images: one from \metabbr and the other from Cut-Mix. 
Human raters are asked to indicate which
one is better, or tie based on three criteria: (1) similarity to the subjects in the reference images, (2) alignment with the given text, and (3) overall image fidelity. 
We provide more details in Appendix~\ref{app:exp:human_eval}.

As shown in \autoref{fig:eval} (Left), \metabbr significantly outperforms prior works in multi-subject fidelity, achieving twice the success rate in preventing identity mixing compared to Cut-Mix. 
Due to this, raters strongly prefer images generated by \metabbr in side-by-side evaluations.
These results confirm that \metabbr effectively decouples the identities of highly similar subjects without stitching artifacts.

\paragraph{Quantitative results}
We evaluate multi-subject personalization methods on two key aspects: {\em multi-subject fidelity}, which measures the preservation of subject details for multiple subjects, and {\em text fidelity}, which assesses how well the generated images align with the given text prompt. 
We use our D\&C scores to evaluate multi-subject fidelity.
For text fidelity, we report the results of \emph{ImageReward}~\citep{xu2024imagereward} and CLIP score (\emph{CLIPs})~\citep{radford2021clip}. 
To avoid position bias, we calculate scores for the two different orders and average them, for example "[$V_1$] and [$V_2$]" and "[$V_2$] and [$V_1$]."

As shown in the Table of \autoref{fig:eval} (Right), our framework achieves the highest scores in both multi-subject and text fidelity, significantly outperforming previous methods.
These results are consistent with qualitative assessments and human evaluations, where \metabbr preserves subject details effectively without identity mixing, unlike prior methods.
The superior text fidelity also indicates that our method generates images that closely follow the given prompt without mixing the subjects.

\begin{figure*}[t!]
\vspace{-0.1in}
\centering    
    \caption{\textbf{Personalizing more than two subjects.} (a) \metabbr successfully personalizes more than two subjects without identity mixing. (b) Success rates when varying the number of subjects.}\label{fig:many}
\includegraphics[width=0.95\linewidth]{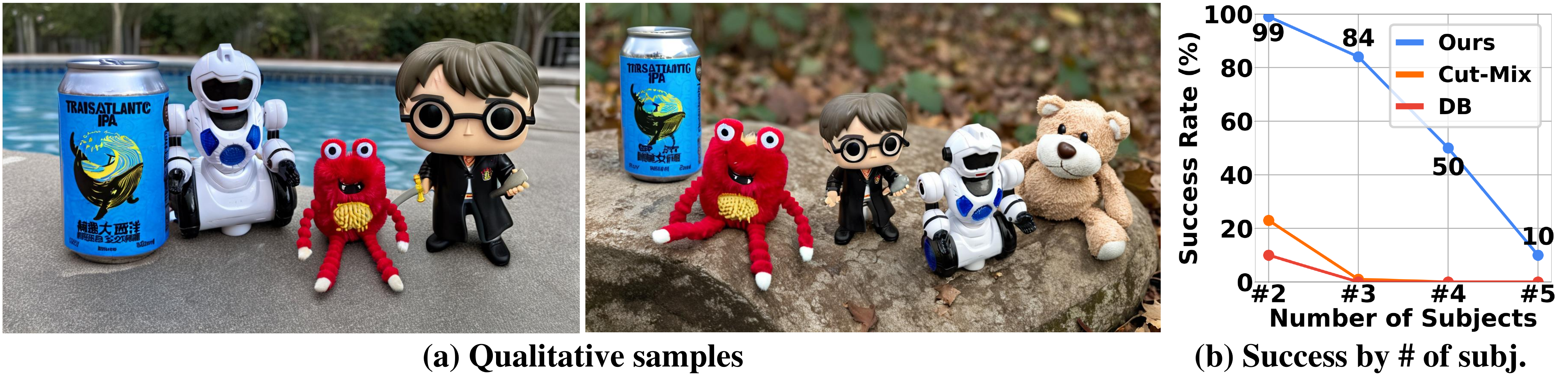}
\vspace{-0.1in}
\end{figure*}
\begin{figure*}[t!]
    \includegraphics[width=1.0\linewidth]{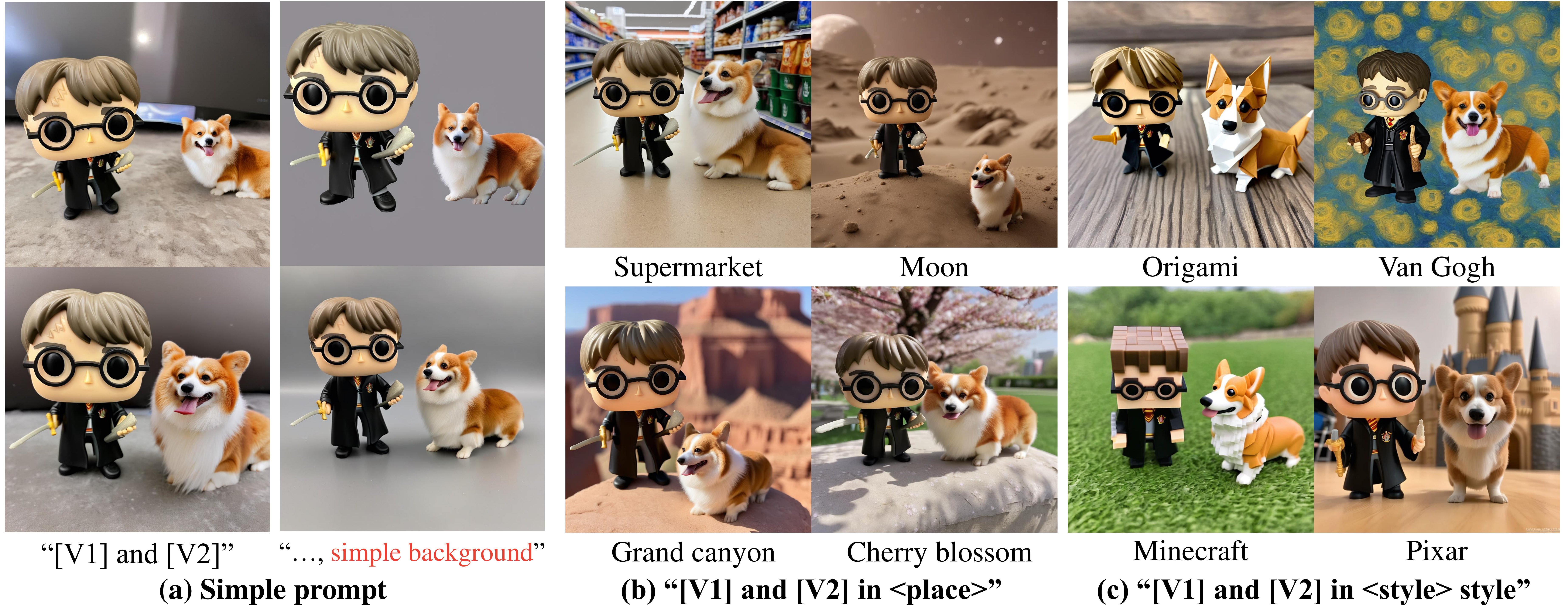}
\vspace{-0.15in}
    \caption{\textbf{Diverse backgrounds generated by MuDI}. Our Seg-Mix does not have a bias with backgrounds due to the training prompt "A photo of [$V_1$] and [$V_2$], simple background". (a) Inference with simple prompts. (b) Inference with various places. (c) Inference with various styles.}
    \label{fig:background}
\vspace{-0.1in}
\end{figure*}

\subsection{Ablation studies \label{exp:ablation}}

\begin{wraptable}{h}{0.4\textwidth}
\vspace{-0.2in}
    \centering
\caption{Results on ablation studies.}
\vspace{-0.05in}
    \resizebox{0.4\textwidth}{!}{
    \renewcommand{\arraystretch}{1.0}
    \renewcommand{\tabcolsep}{2pt}
\begin{tabular}{l c c}
\toprule
     & \multicolumn{2}{c}{Multi-Subject Fidelity} \\
\cmidrule(l{2pt}r{2pt}){2-3}
    Method & D\&C-DS$\uparrow$ & D\&C-DINO$\uparrow$ \\
\midrule
    w/o Seg-Mix & 0.475 & 0.481 \\
    w/o Initialization & 0.477 & 0.480 \\
    w/o Desc. Class & 0.556 & 0.558 \\
\midrule
    \metabbr (Ours) & \textbf{0.637} & \textbf{0.610} \\
\bottomrule
\end{tabular}}
\vspace{-0.2in}
\label{tab:abl_quantitative}
\end{wraptable}

\vspace{-0.05in}
\paragraph{Necessity of Seg-Mix}
To validate that Seg-Mix is crucial for decoupling the subjects' identities, we compare \metabbr against its variant without it. 
As shown in \autoref{tab:abl_quantitative}, ablating Seg-Mix results in low multi-subject fidelity due to identity mixing.
\autoref{fig:ablation} demonstrates that the attributes of the Corgi and Chow Chow are completely mixed without Seg-Mix. 
In particular, we show in \autoref{fig:appendix_controlnet} that using additional spatial conditioning, e.g., ControlNet~\citep{zhang2023controlnet}, without Seg-Mix still suffers from identity mixing.

\begin{figure*}[t!]
    \centering    
    \includegraphics[width=1.0\linewidth]{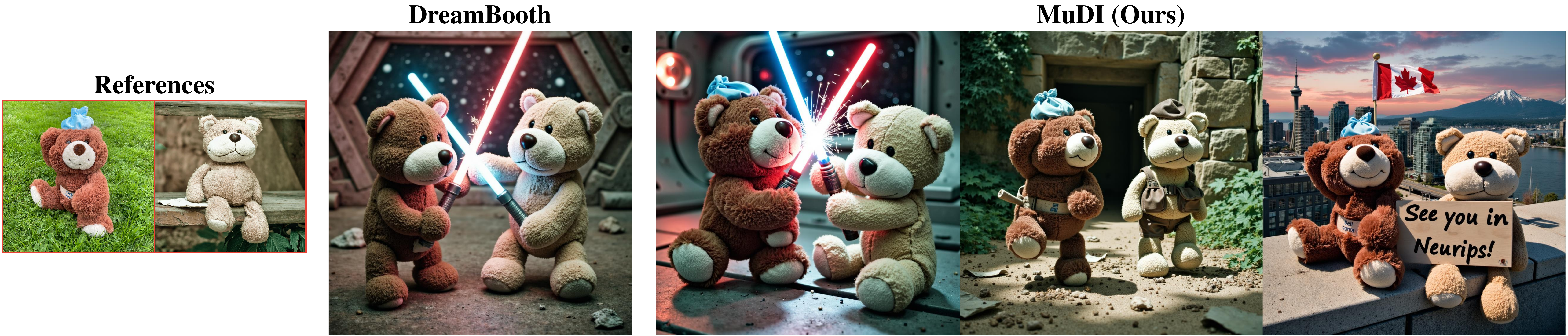}
    \centering    
    \includegraphics[width=1.0\linewidth]{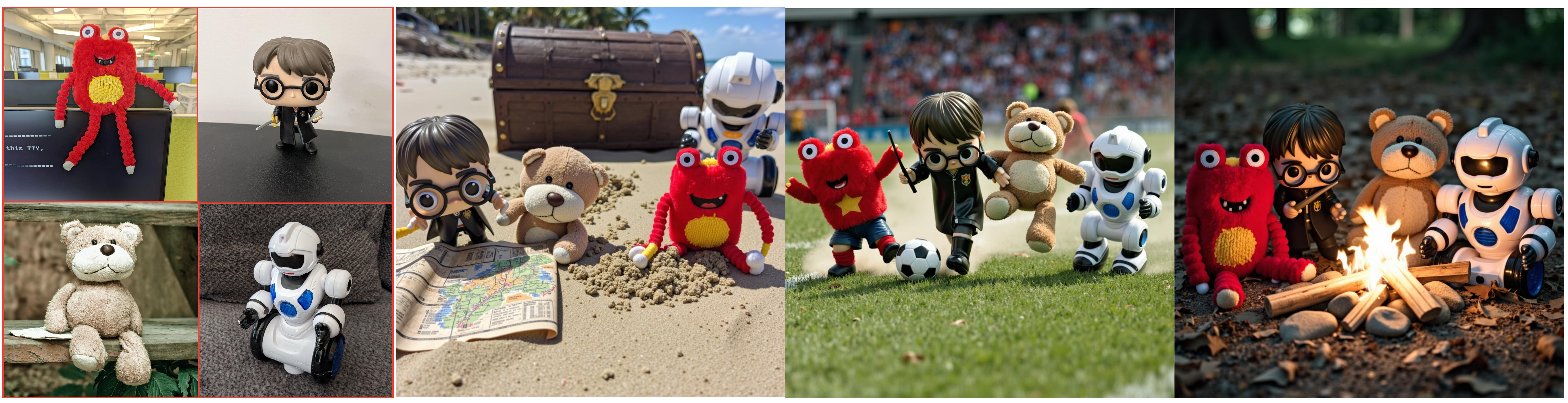}
\vspace{-0.15in}
    \caption{\textbf{Images generated using FLUX~\citep{flux} as a pre-trained text-to-image diffusion model}. (Top row) DreamBooth produces mixed-identity teddy bears while MuDI generates distinct bears. (Bottom row) MuDI can personalize many subjects without identity mixing on diverse backgrounds.}
    \label{fig:flux}
\vspace{-0.1in}
\end{figure*}

\paragraph{Importance of our initialization}
We show in \autoref{fig:ablation} that the inference initialization improves identity separation and alleviates subject dominance. \autoref{tab:abl_quantitative} validates that images generated without initialization result in lower subject fidelity. 
We empirically find that our initialization provides significant benefits in three scenarios: (1) personalizing unusual subjects that pre-trained models struggle to generate (e.g., the cloud man of \autoref{fig:fig1} bottom-right), (2) personalizing more than two subjects, and (3) using complex prompts, which we explain in detail in Appendix~\ref{app:add:init}.

\vspace{-0.05in}
\paragraph{Descriptive class}
We show in \autoref{fig:ablation} that using descriptive classes to represent the subjects improves the preservation of the subjects' detail, and \autoref{tab:abl_quantitative} further shows that this method enhances subject fidelity. 
Despite the improvement, relying only on descriptive classes may occasionally lead to some subjects being ignored. This is effectively addressed by applying our initialization which results in significantly improved outcomes.

\vspace{-0.05in}
\paragraph{More than two subjects}
\autoref{fig:many}(b) shows the success rates of \metabbr, DreamBooth~\citep{ruiz2023dreambooth}, and Cut-Mix~\citep{han2023svdiff} as the number of subjects varies.
Our method achieves significantly high success rates, while previous approaches~\citep{ruiz2023dreambooth,han2023svdiff} fail to personalize even two subjects effectively.
In particular, our method shows over 50\% success for generating four objects together (see \autoref{fig:many}(a)).
However, we observe that the performance of \metabbr decreases as the number of personalized subjects increases, particularly for highly similar subjects.
We provide further details in Appendix~\ref{app:add:many}.

\vspace{-0.05in}
\paragraph{Diverse background}
As shown in \autoref{fig:background}, our Seg-Mix does not have a bias with white backgrounds and can generate diverse backgrounds. This is because the prompt “A photo of [$V_1$] and [$V_2$], simple background” is used during training for the image of segmented subjects composed on a white background. This effectively disentangles the background from the identities through the text “simple background”, preventing overfitting.

\vspace{-0.05in}
\paragraph{Model agnostic}
Notably, MuDI is a model-agnostic personalization method as it is based on data augmentation during training that does not require model-specific techniques, such as utilizing attention maps~\citep{avrahami2023break, xiao2023fastcomposer} or choosing where to fine-tune~\citep{kumari2023multi, tewel2023perfusion}.
We validate this by using FLUX~\citep{flux} as the pre-trained text-to-image model based on DiT~\citep{peebles2023dit}, which is different from SDXL~\citep{podell2023sdxl} based on UNet~\citep{ronneberger2015unet} backbone. 
As shown in \autoref{fig:flux} top row, DreamBooth produces mixed identity teddy bears while MuDI successfully generates distinct bears without identity mixing. We show in \autoref{fig:flux} bottom row that MuDI can personalize multiple subjects using FLUX in diverse backgrounds.

\subsection{Other use cases \label{exp:applications}}

\begin{figure*}[t!]
    \includegraphics[width=1.0\linewidth]{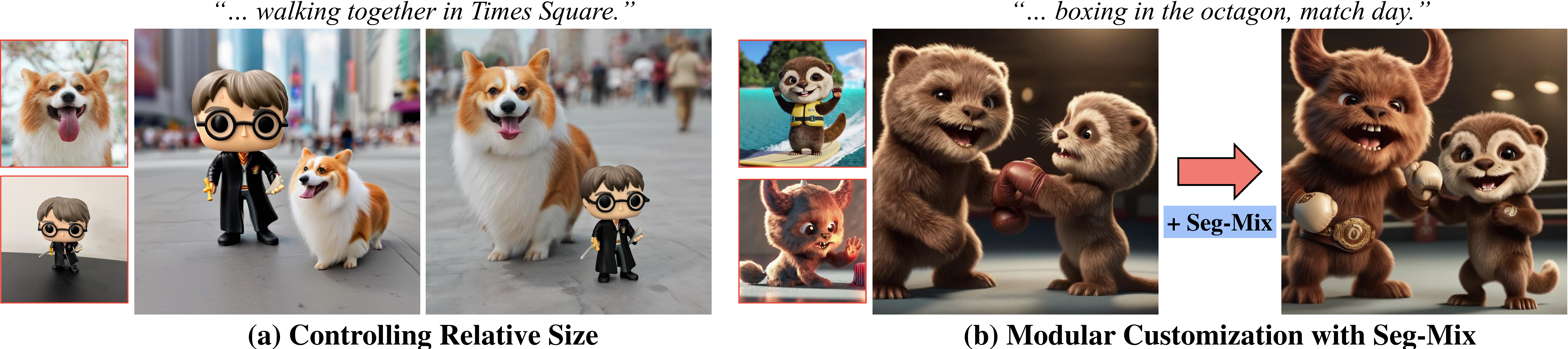}
\caption{{Examples of other use cases of our method.} \textbf{(a) Controlling relative size with Seg-Mix.} We visualize samples generated by \metabbr using size-controlled Seg-Mix. \textbf{(b) Modular customization.} Applying Seg-Mix after merging LoRAs significantly improves identity decoupling.}
\label{fig:applications}
\vspace{-0.1in}
\end{figure*}

\paragraph{Controlling relative size \label{exp:applications:size}}
Our framework offers an intuitive way to control the relative size between the personalized subjects. 
By resizing the segmented subjects according to user intents in Seg-Mix, we find that personalized models generate subjects with the desired relative sizes. 
This showcases another benefit of our method unlike previous methods~\citep{ruiz2023dreambooth,kumari2023multi,han2023svdiff}, which often result in inconsistent relative sizes due to a lack of size information during fine-tuning.
As shown in \autoref{fig:applications}(a), our method allows the model to be personalized to generate either a larger dog compared to the toy or vice versa, by setting their relative sizes during Seg-Mix. 
The generated images show a consistent relative size which we provide more examples in \autoref{fig:appendix_relative_size}.
Additionally, controlling the relative size of the segmented subjects during inference initialization can further improve the size consistency.

\paragraph{Modular customization \label{exp:applications:modular}}
The proposed Seg-Mix can also be applied to modular customization, where the subjects are independently learned in advance by single-subject LoRAs~\citep{hu2021lora}. We then efficiently combine these LoRAs to generate multi-subject images. To integrate Seg-Mix with modular customization, we first generate images for each subject using their respective single-subject LoRA, which serve as reference images.
Next, we merge the single-subject LoRAs using an existing method such as gradient fusion~\citep{gu2024mix}. After merging, we apply Seg-Mix with the generated single-subject images for 200-300 iterations. This approach effectively reduces identity mixing and avoids the need for training from scratch by reusing the single-subject LoRAs. We illustrate the process of using Seg-Mix in \autoref{fig:appendix_modular}.

\autoref{fig:applications}(b) shows samples generated by gradient fusion~\citep{gu2024mix}, a modular customization method, applied to two-subject personalization. Without spatial conditioning, it produces mixed identities for the characters of the otter and the monster (left). 
However, if we fine-tune fused model with Seg-Mix only for a few iterations, the fine-tuned model produces a high-quality image of clearly separated subjects (\autoref{fig:applications}(b), right).
We note that it is important to incorporate Kullback-Leibler (KL) divergence as regularization~\citep{fan2024dpok} in fine-tuning in order to prevent saturation and overfitting.

\subsection{Iterative training \label{exp:sft}}
\vspace{-0.05in}
To further improve the quality, we investigate a fully automatic iterative training (IT) method~\citep{sohn2023styledrop}, which fine-tunes the personalized model using high-quality samples obtained from an earlier training stage.
Specifically, we first generate multi-subject images with MuDI and select high-quality images based on the D\&C score, which closely aligns with the human evaluation. 
These selected images are then used to fine-tune the personalized model, with KL regularization~\citep{fan2024dpok} added to Eq.~\eqref{eq:dreambooth}.
By applying IT to the images of Corgi and Chow Chow, the D\&C-DS score is improved from 0.613 to 0.672, achieving a higher success rate (see \autoref{fig:appendix_SFT}).
We provide further details in Appendix~\ref{app:samples:sft}.
\section{Conclusions}
In this work, we present \metabbr, a novel personalizing framework for multiple subjects that addresses identity mixing.
We leverage segmented subjects automatically obtained from the foundation model for image segmentation for both training and inference, through data augmentation for training pre-trained models and initializing the generation process.
We experimentally validate our approach on a new dataset comprising combinations of subjects prone to identity mixing, for which ours successfully prevents mixing even for highly similar subjects. 
We describe the limitations and societal impacts of our work in Appendix~\ref{app:limitation}.
We hope that our work can serve as a starting point to develop personalizing methods for multiple concepts in more challenging scenarios.

\section{Acknowledgements}
We thank Juyong Lee, and Jaewoo Lee for providing valuable feedback.
This work was supported by National Research Foundation of Korea (NRF) grant funded by the Korea government (MSIT) (No. RS-2023-00256259), Institute for Information \& communications Technology Promotion(IITP) grant funded by the Korea government(MSIT) (No.RS-2019-II190075 Artificial Intelligence Graduate School Program(KAIST)), Institute of Information \& communications Technology Planning \& Evaluation(IITP) grant funded by the Korea government(MSIT) (No. RS-2024-00509279 Global AI Frontier Lab), Artificial intelligence industrial convergence cluster development project funded by the Ministry of Science and ICT(MSIT, Korea)\&Gwangju Metropolitan City, and KAIST-NAVER Hypercreative AI Center.

\bibliography{references}

\newpage
\appendix
\begin{center}{\bf {\LARGE Appendix}}\end{center}
\vspace{0.1in}

\paragraph{Organization} The Appendix is organized as follows: In Section~\ref{app:exp}, we describe the details of the experiments and our framework. We provide additional experimental results in Section~\ref{app:add}, and further discussion on other use cases of \metabbr in Section~\ref{app:applications}.
Lastly, in Section~\ref{app:limitation}, we discuss the limitations and societal impacts of our work.

\section{Experimental details \label{app:exp}}

\subsection{Implementation details \label{app:exp:details}}

\paragraph{Training details}
In our experiment, we use Stable Diffusion XL (SDXL)~\citep{podell2023sdxl} as the pre-trained text-to-image diffusion model. We employ LoRA~\citep{hu2021lora} with a rank of 32 for the U-Net~\citep{ronneberger2015unet} module, instead of training the full model weights. We do not train the text encoder.

For all methods, we pair the reference images with comprehensive captions obtained through GPT-4v~\citep{gpt-4v}, instead of using a simple prompt like "A photo of a [V]". This effectively mitigates overfitting to the background and shows better text alignment for the baselines and our method. 

We construct the prior dataset $\mathcal{D}_{ref}$ in Eq.~\eqref{eq:dreambooth} by generating from the pre-trained text-to-image models using a prompt \emph{"A photo of <class>, simple background, full body shot."}. 
Since the generated prior images may contain more than one subject, we select images that contain a single subject. 

We determine the training iterations of Seg-Mix on each combination in the dataset based on the difficulty of personalizing the subjects individually using DreamBooth~\citep{ruiz2023dreambooth}.
For example, combinations including highly detailed subjects, such as "can" in \autoref{fig:qual}, require from 1400 to 1600 training iterations, while combinations containing subjects easy to learn, e.g., "dog," require about 1200 iterations.
We use a fixed augmentation probability of 0.3 for both Cut-Mix and our Seg-Mix with the same number of training iterations for a fair comparison. 
To prevent subject overfitting, we use 1000 training iterations for DreamBooth.


\begin{algorithm}[H]
\small
\caption{MuDI training (Seg-Mix)}\label{alg:seg-mix}
\begin{lstlisting}
# preprocess_data: list of all reference images & masks pair ([(imgs_0, masks_0), ...])
# max_m: max margin from both ends of the image (if large, allow overlap)
# scales: control the relative size else random resizing

def create_seg_mix(imgs, masks, out_size=(1024,1024), max_m=1, scales=None):
    imgs, masks = random.choice(preprocess_data, 2, replace=False) # sample 2 refs
    # randomly(or relative) resize each image & mask pair
    imgs, masks = resize(imgs, masks, out_size, scales)
    out, out_mask = np.zeros((*out_size, 3)), np.zeros(out_size) # blank image, mask
    if random.random() < 0.5: # random order swap
        imgs, masks = imgs[::-1], masks[::-1]
    # random margin from ends of the image
    m = [random.randint(0, max_margin) _ for in range(2)]
    out, out_mask = paste_left(out, out_mask, imgs[0] * masks[0], m[0])
    out, out_mask = paste_right(out, out_mask, imgs[1] * masks[1], m[1])
    return out, out_mask

def train_loss(seg_mix_prob=0.3, **kwargs):
    img, mask, class, prompt = dataloader.next()
    if random.random() < seg_mix_prob: # do augmentation
        # sample another class for seg-mix
        new_class = random.choice(class_list - class)
        new_img, new_mask = sample_img_mask(new_class)
        imgs, masks = [img, new_img], [mask, new_mask]
        img, mask = create_seg_mix(imgs, masks, **kwargs)
    # DreamBooth training (Eq. 1)
    loss = LDM_loss(img, prompt)
    return loss.mean()  
\end{lstlisting}
\label{a}
\end{algorithm}


    
\begin{algorithm}[t]
\small
\caption{MuDI inference (Initialization)}\label{alg:seg-mix-init}
\begin{lstlisting}
# class_list: classes in prompt
# gamma: guidance strength of our initialization
# kwargs: same arguments from Algorithm 1

def latent_initialize(class_list, gamma=1.0, **kwargs):
    # sample reference images
    imgs, masks = zip(*[sample_img_mask(cls) for cls in class_list])
    # create_seg_mix from Algorithm 1
    out, mask = create_seg_mix(imgs, masks, **kwargs)
    # encode image to latent
    out_latent = encoder(out) 
    # resize to latent size
    out_mask = resize(out_mask) 
    # segmented latent
    init_latent = out_mask * out_latent
    noise = torch.rand_like(init_latent)
    # forward process with gamma scaling
    init_latent = add_noise(init_latent * gamma, strength=1, noise=noise)
    return init_latent

def inference(prompt, class_list, gamma=1.0, **kwargs):
    # execute computation only once
    init_latent = latent_initialize(class_list, gamma, **kwargs)
    # existing inference pipeline
    img = inference_pipe(prompt, init_latent=init_latent, **kwargs)
    return img
\end{lstlisting}
\end{algorithm}


    

\begin{figure*}[t!]
\centering
    \includegraphics[width=1\linewidth]{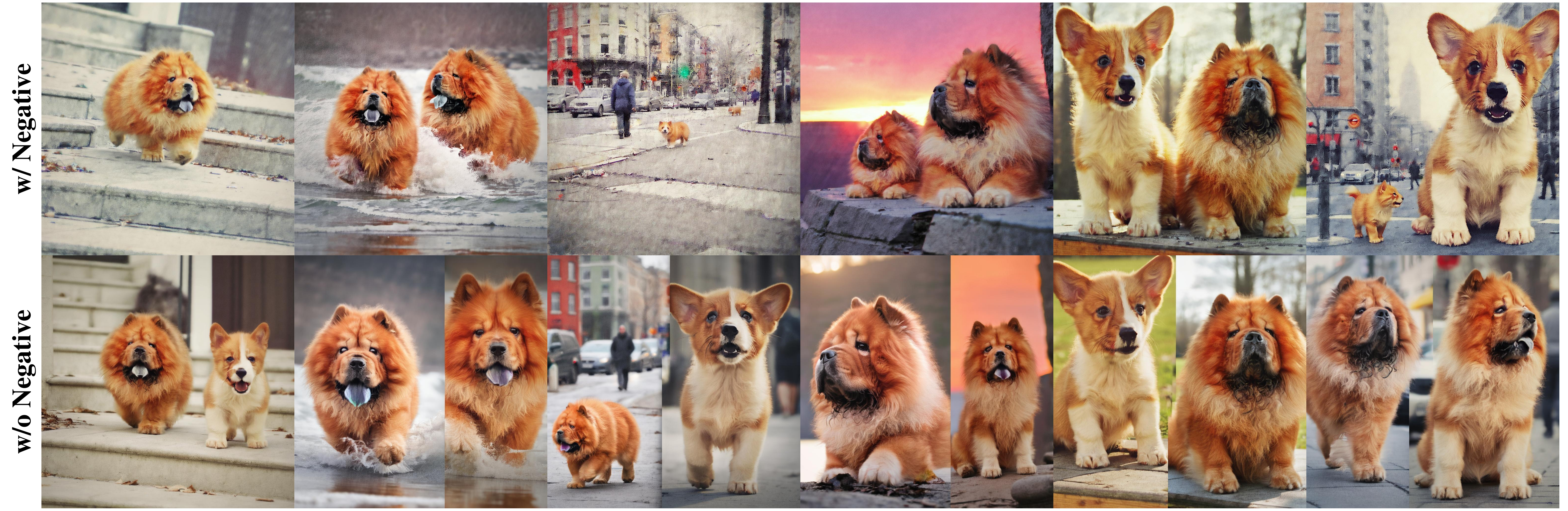}
\vspace{-0.1in} 
    \caption{\textbf{Cut-Mix with and without negative prompt}. We observe that using the negative prompt "A dog and a dog" leads to reduced artifacts but results in over-saturation as shown in the first row.}
    \label{fig:cutmix_negative}
\end{figure*}

\paragraph{MuDI training details}
We provide a pseudocode of our \metabbr training in \autoref{alg:seg-mix}. 
For Seg-Mix, we randomly rescale the segmented subjects and randomly choose the location of the subjects to create random compositions of the segmented subjects. Note that the margin from the boundaries of the image is also set as random, where larger margins allow for the possibility of subjects overlapping.

In particular, when creating the prior dataset $\mathcal{D}_{ref}$ for training, we use a descriptive class (Section~\ref{method:train}) to serve as the prior class.
We automatically create segmentation masks for the images in the prior dataset using OWLv2~\citep{minderer2024owlv2} and SAM~\citep{kirillov2023sam}, similar to segmenting the reference images described in Section~\ref{method:train}. We illustrate the segmentation of the prior dataset in \autoref{fig:method} (denoted as Priors). 
Note that we select images that contain a single subject and also result in a single segmentation mask. We generated 50 images for the prior dataset.

After the preprocessing step, the training of \metabbr takes almost the same duration as DreamBooth~\citep{ruiz2023dreambooth}, taking about 90 minutes to personalize two subjects on a single RTX 3090 GPU. 
We use AdamW optimizer~\citep{LoshchilovH19adamw} with $\beta_1=0.9$, $\beta_2=0.999$, weight decay of $0.0001$, and a learning rate of 1e-4, following the setting of DreamBooth~\citep{ruiz2023dreambooth}, and set the batch size to 2.

\paragraph{\metabbr inference details}
We provide a pseudocode of our inference initialization in \autoref{alg:seg-mix-init}. 
We first create images of randomly composed segmented subjects using the reference images and the extracted segmentation masks. 
The composition can be either random, manually set, or obtained by using LLM as described in Section~\ref{method:inference}. 
The images are then encoded into a latent with the VAE of the SDXL~\citep{podell2023sdxl}, which is scaled by a factor of $\gamma$. 
The scaled latent is perturbed by the forward noising process from time 0 to T, which results in the initial latent for the inference process.
We control the magnitude of the $\gamma$ scale and the relative size between the segmented subjects to address identity mixing as well as subject dominance.


\paragraph{Inference using negative prompt}
While \citet{han2023svdiff} propose using negative prompts to reduce stitching artifacts, we observe that this produces over-saturated samples. As shown in the top row of \autoref{fig:cutmix_negative}, using negative prompts results in low-quality images.
Therefore, we opt not to use negative prompts in the evaluation of Cut-Mix~\citep{han2023svdiff}. 
For our framework, we use a simple negative prompt \emph{"sticker, collage."} that alleviates sticker-like artifacts caused by over-training with the segment-and-mixed images.

\begin{figure*}[t!]
\centering
    \includegraphics[width=1\linewidth]{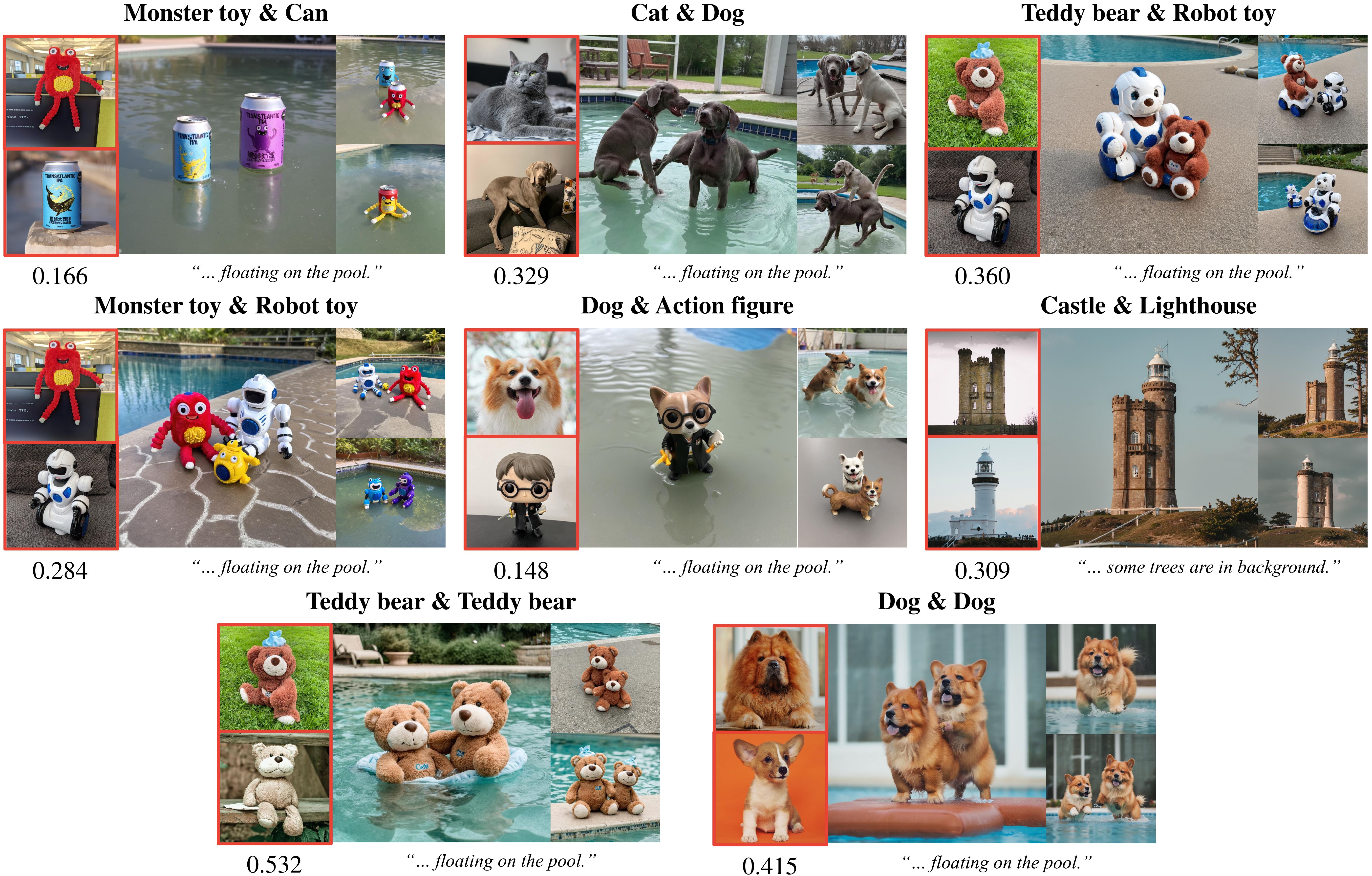}
\vspace{-0.20in} 
    \caption{\textbf{Dataset.} We introduce a new dataset comprising eight combinations of similar subjects. For each combination, we visualize one image per subject (red boxes) and three images generated by DreamBooth. The score below the subjects denotes the DreamSim~\citep{fu2023dreamsim} similarity score between 0 and 1, where a larger value indicates higher similarity. 
    The bottom-most two combinations have the highest similarity which makes them challenging to personalize without mixing the identities.}
    \label{fig:appendix_dataset}
\vspace{-0.05in}
\end{figure*}

\subsection{Dataset \label{app:exp:dataset}}
We introduce a new dataset to facilitate evaluation for multi-subject personalization, comprising 8 combinations of similar subjects prone to identity mixing. 
We collected images from the datasets widely used, namely the DreamBench dataset~\citep{ruiz2023dreambooth} and the CustomConcept101 dataset~\citep{kumari2023multi}.  
We construct the dataset to comprise diverse categories including animals, objects, and scenes. 
We visualize the subjects and the identity-mixed samples from DreamBooth in \autoref{fig:appendix_dataset}.

\clearpage
\begin{algorithm}[t!]
    \small
    \caption{Detect \& Compare (D\&C)}\label{alg:dnc}
        \textbf{Input:} Embeddings of reference subjects $(R_1, R_2, ... R_N)$, Embeddings of boxes $(B_1, B_2, ... B_M)$  \\
        \textbf{Output:} D\&C score (0 $\sim$ 1, it depends on similarity between references)
    \begin{algorithmic}[1]
        \IF[Count Error, return 0] {M != N}
            \STATE \textbf{return} {$0$}
        \ELSE
            \STATE $\bm{S}^{GT}, \bm{S}^{DC} = [0]_{N\times N}, [0]_{N\times N}$ \COMMENT{Initialize square matrix}
            \FOR {$i=1,2,\ldots N$}
                \FOR {$j=1,2,\ldots N$}
                    \STATE $\bm{S}^{GT}_{ij} = \text{mean}(\text{matmul}(R_i, R_j))$ \COMMENT{This can be pre-calculated before, and has symmetric property}
                    \STATE $\bm{S}^{DC}_{ij} = \text{mean}(\text{matmul}(B_i, R_j))$
                \ENDFOR
            \ENDFOR
        \ENDIF
        \STATE $\bm{S}^{DC}$ = row-wise-sort$(\bm{S}^{DC})$ \COMMENT{Sort the rows based on the maximum value of each column sequentially}
        \STATE score $=1-\|\bm{S}^{GT}-\bm{S}^{DC}\|^2_F$
        \STATE \textbf{return} {score}
    \end{algorithmic}
\end{algorithm}

\begin{figure*}[t!]
\vspace{-0.1in} 
\centering
    \includegraphics[width=\linewidth]{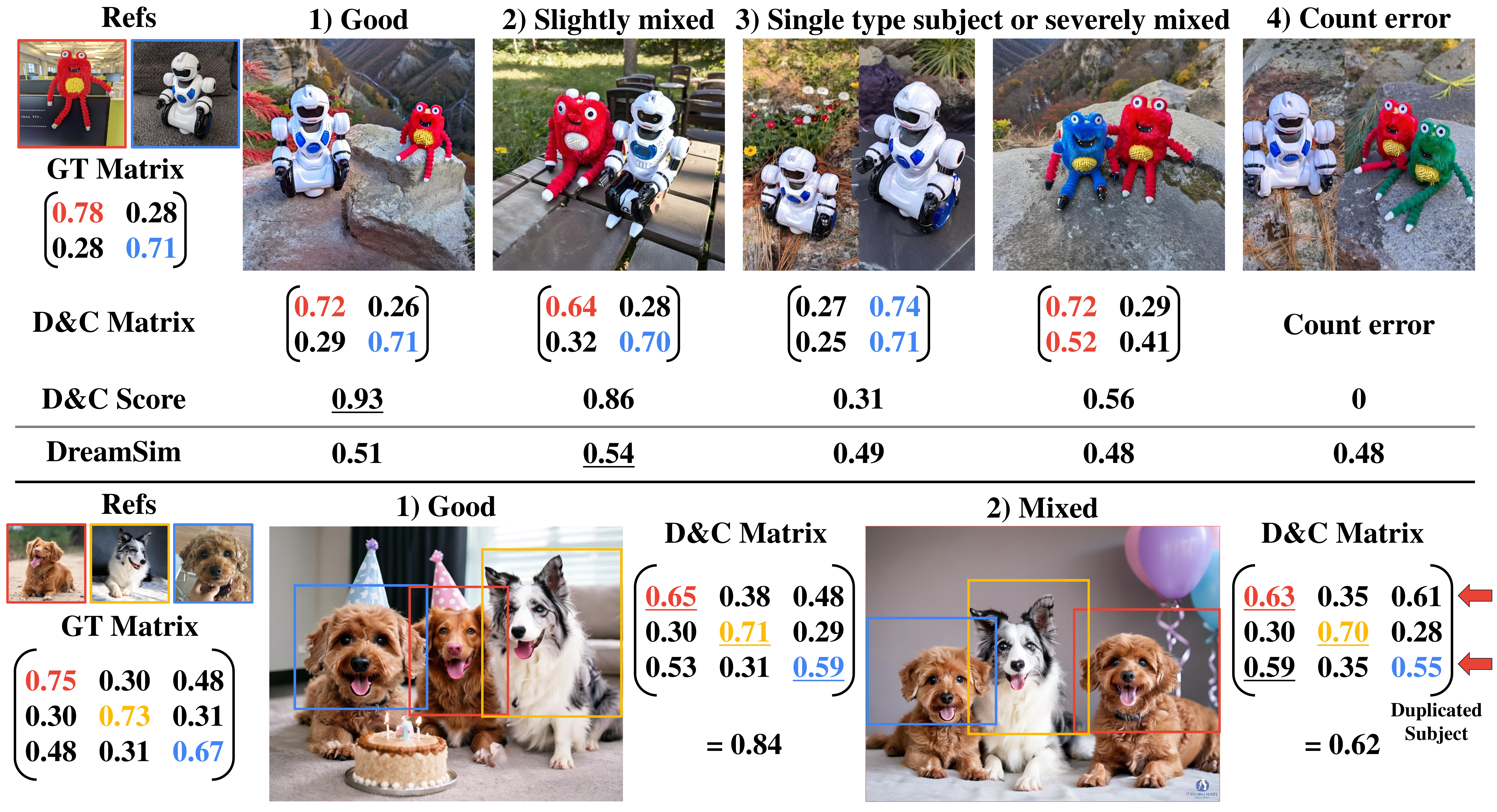}
    \caption{\textbf{(Top row)} We visualize the D\&C-DS scores and DreamSim scores for various cases.
    \textbf{(Bottom row)} We provide examples of D\&C-DS scores for images of three subjects. 
    }
    \label{fig:appendix_dnc_sample}
\end{figure*}

\subsection{Detect-and-Compare \label{app:exp:eval}}
We summarize the process of measuring the D\&C score in \autoref{alg:dnc}.

\paragraph{Correlation with human evaluation}
In the table of \autoref{fig:dnc_overview}, we assess the correlation between human evaluation and the metrics (D\&C scores and DreamSim scores).
We generate a total of 1600 images, 400 images from Textual Inversion~\citep{gal2022image}, DreamBooth~\citep{ruiz2023dreambooth}, Cut-Mix~\citep{han2023svdiff}, and our \metabbr, respectively, and ask human raters to evaluate the multi-subject fidelity via binary feedback. 
We then measure the correlation between human evaluation results and the scores using Spearman's rank correlation coefficient and Area under the Receiver Operating Characteristic (AUROC).
Note that the Spearman's rank correlation was computed using the normalized sum of all human evaluation answers (e.g., if 3 out of 5 raters answered "good," the score is 0.6). The AUROC was computed based on the majority voting results (0 or 1).

\paragraph{Examples}
In the top row of \autoref{fig:appendix_dnc_sample}, we categorize the generated images from DreamBooth~\citep{ruiz2023dreambooth}, Cut-Mix~\citep{han2023svdiff}, and \metabbr into four cases, and provide the D\&C matrix, D\&C score, and DreamSim score for each image.
For successful images, the difference between $\bm{S}^{GT}$ and $\bm{S}^{DC}$ is significantly small and results in high D\&C scores. 
In cases where the identities are severely mixed or show two identical subjects, the difference becomes considerably larger, and the D\&C scores decrease.
For example, the fourth generated image features two monster toys, where one is blue and the other is red. The blue monster toy resembles both the reference monster toy and the robot toy, leading to a significant difference in the second row of $\bm{S}^{GT}$ and $\bm{S}^{DC}$. 
However, DreamSim~\citep{fu2023dreamsim} extended to multi-subject settings, which compute the mean similarity to all the reference images~\citep{lui2023cones}, fails to distinguish between these cases effectively. 

We also provide an example of D\&C matrices and scores for three subjects in the bottom row of \autoref{fig:appendix_dnc_sample}.
Our D\&C can be easily applied to evaluate identity mixing for many subjects.

\paragraph{Qualitative comparison of D\&C and DreamSim}
Additionally, we analyze the alignment of D\&C to the human evaluation by comparing with DreamSim in \autoref{fig:appendix_dnc_qualitative}. 
We sort 24 images generated by \metabbr based on the D\&C-DS scores and DreamSim scores, respectively, and compare the ranking with the human evaluation.
D\&C perfectly aligns with the human evaluation, giving lower scores to failed images with mixed identities.
However, the single-subject metric DreamSim fails to align with human evaluation, giving high scores to images with mixed identities or the wrong number of subjects.
This indicates that single-subject metrics are ill-suited to be used for evaluating multi-subject fidelity.
The qualitative comparison agrees with the quantitative analysis in \autoref{fig:dnc_overview}, where we show that D\&C achieves a high correlation with the human evaluation of multi-subject fidelity.

\begin{figure*}[t!]
\centering
    \includegraphics[width=1.0\linewidth]{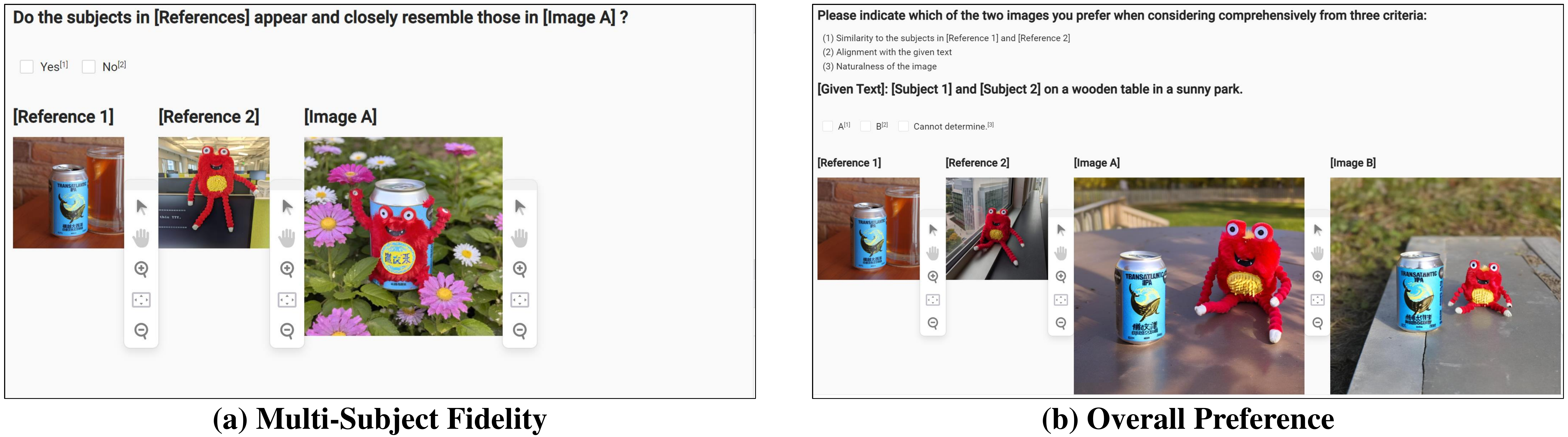}
    \caption{A screenshot of questionnaires from our human evaluation on (a) multi-subject fidelity and (b) overall preference. }
    \label{fig:appendix_human_eval}
\end{figure*}
\begin{figure*}[t!]
\centering
    \includegraphics[width=1.0\linewidth]{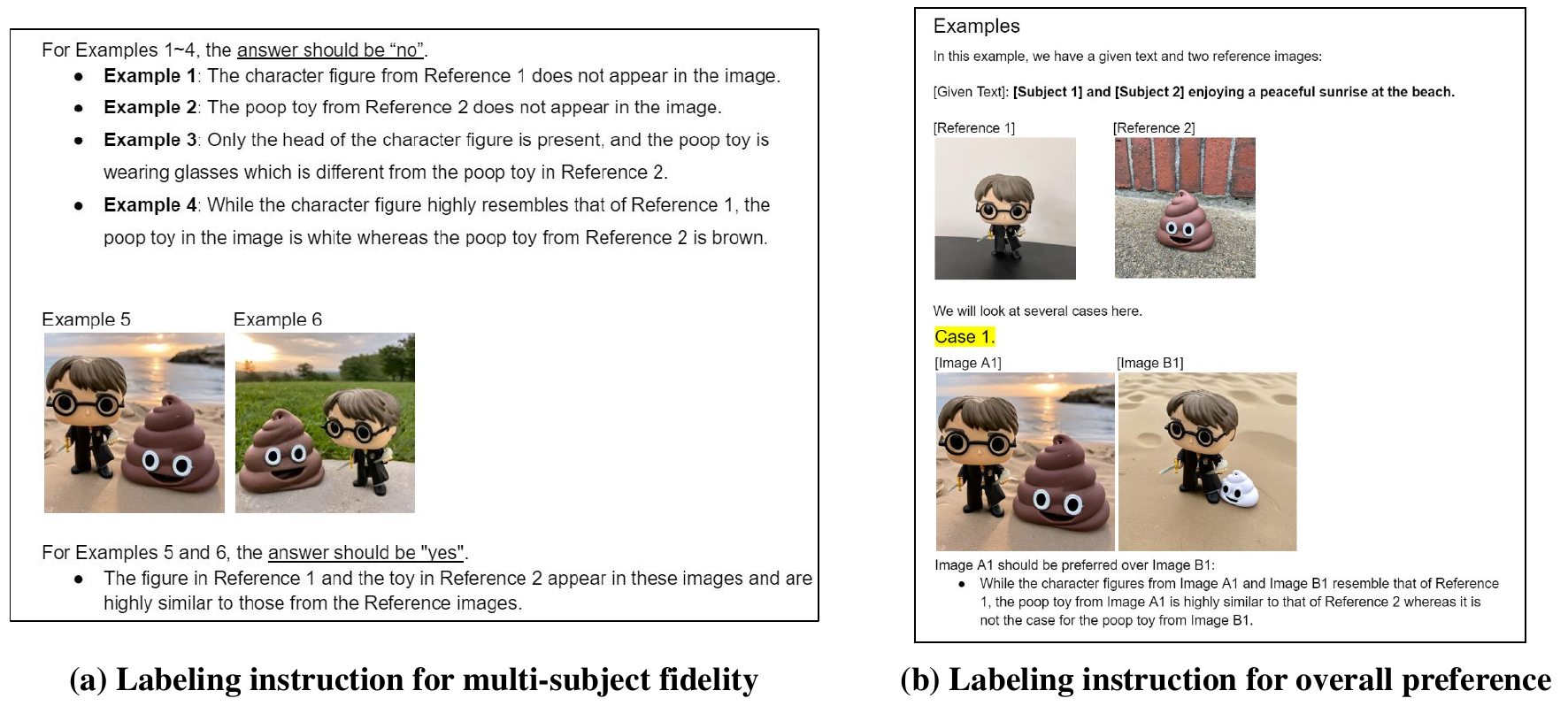}
    \caption{A screenshot of labeling instruction from our human evaluation on (a) multi-subject fidelity and (b) overall preference.}
    \label{fig:appendix_label_inst}
\end{figure*}

\subsection{Evaluation details} \label{app:eval-details}
We evaluate 400 generated images for each method, across 8 combinations with 5 evaluation prompts and 10 images of fixed random seeds.
\begin{figure*}[t!]
\centering
    \includegraphics[width=1.0\linewidth]{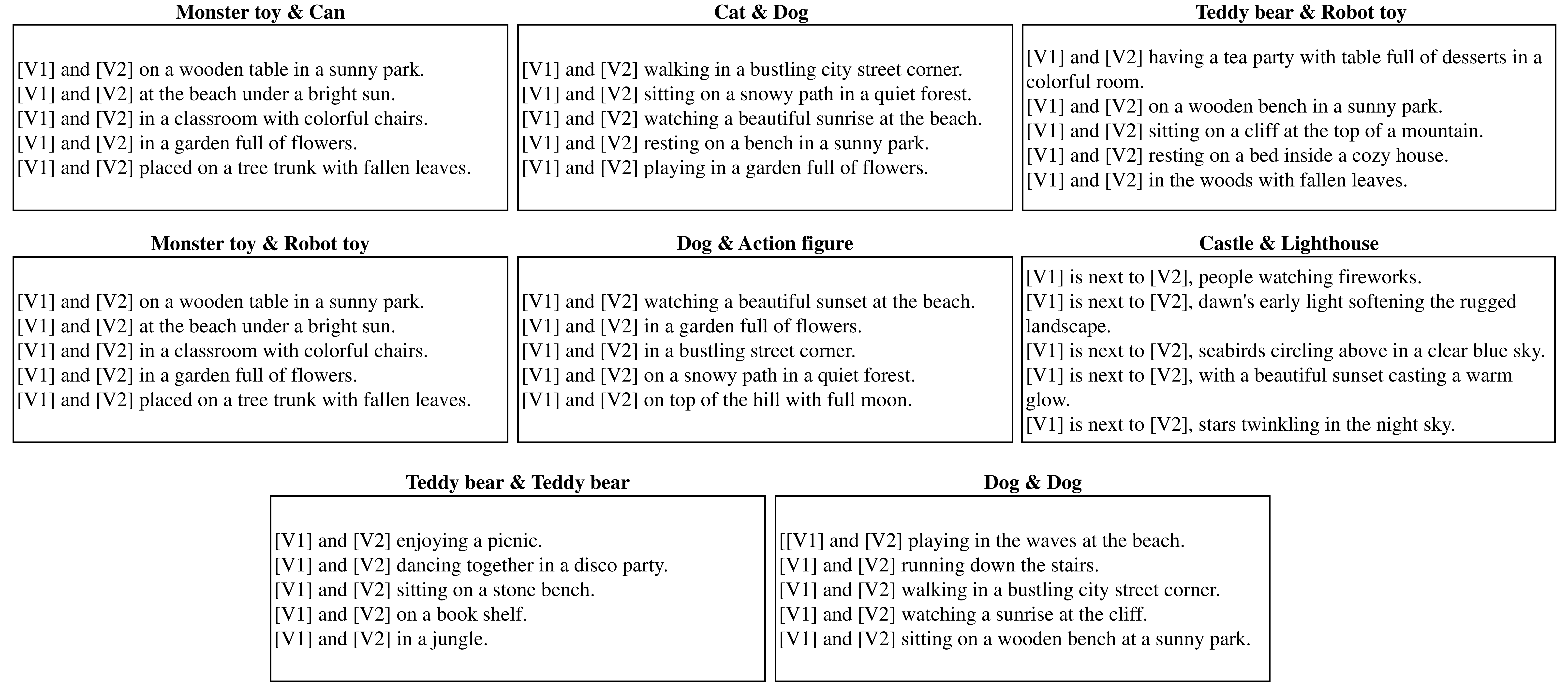}
    \caption{Our evaluation prompts for each concept.}
    \label{fig:appendix_prompts}
\end{figure*}
\paragraph{Evaluation prompts}
To evaluate the personalization methods, we generate 5 evaluation prompts for each combination in the dataset using ChatGPT~\citep{chatgpt}.
Each prompt describes a scene of the subjects with simple action such as \emph{"... in a classroom with colorful chairs,"} or \emph{"... walking in a bustling city street corner."}. 
We avoid using complex prompts as models fail to generate images that align with such prompts, regardless of the identity mixing. We note that the evaluation prompts were unseen during training. Details of our evaluation prompts for each concept are in \autoref{fig:appendix_prompts}.

\paragraph{Quantitative evaluation}
For evaluating the text fidelity, we made fair comparisons by using descriptive classes for both evaluation and text prompts. 
To avoid positional bias from the order of subjects in the prompts, we measured the scores for all possible orders of the subjects, for example, "monster toy and can" and "can and monster toy", and reported the average of the scores. 
In the case of ImageReward~\citep{xu2024imagereward}, the score differences between the different subject orders were not small.

\paragraph{Human evaluation\label{app:exp:human_eval}}
Our human evaluation was conducted in two main aspects: (1) multi-subject fidelity and (2) overall preference. For multi-subject fidelity, a random subset containing an equal number of instances from each method was created and provided to human raters for binary feedback. For overall preference, images from both Cut-Mix and ours were provided in random order, with all images generated from the same seed. We provide reference images of each subject along with two anonymized images, i.e., one from \metabbr and the other from Cut-Mix
We ask human raters to evaluate which image they prefer based on three criteria: (1) similarity to the subjects in the reference images, (2) alignment with the given text, and (3) image fidelity. 
If both images fail to depict the subjects in the reference images, raters are instructed to select "cannot determine". 
We provide screenshots of questionnaires and labeling instructions in \autoref{fig:appendix_human_eval} and \autoref{fig:appendix_label_inst}, respectively.

\begin{figure*}[ht!]
\centering
    \includegraphics[width=0.95\linewidth]{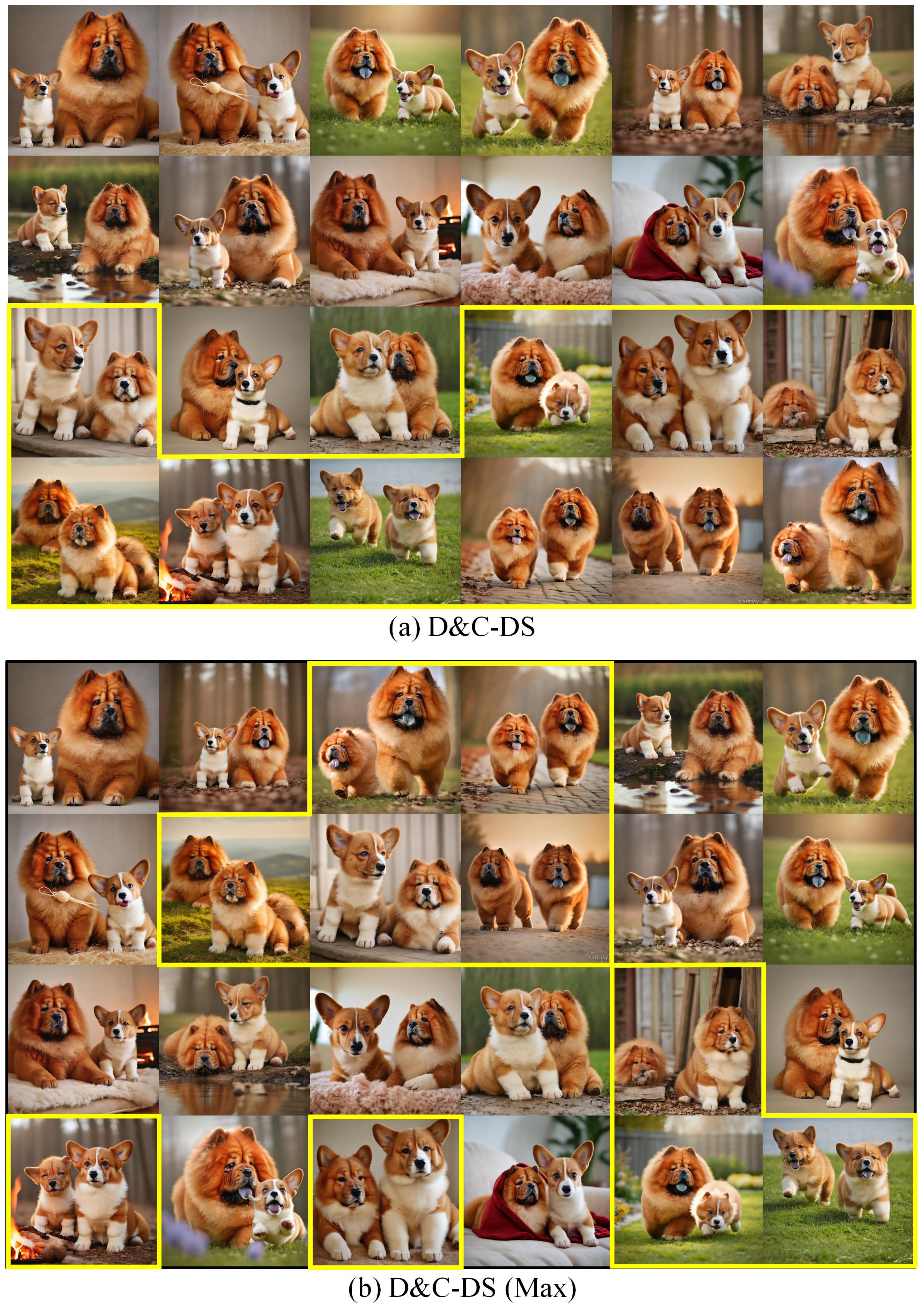}
\vspace{-0.10in} 
    \caption{\textbf{Qualitative comparison of D\&C-DS and DreamSim~\citep{fu2023dreamsim}.} 
    We sort 24 images generated by \metabbr based on (a) D\&C-DS and (b) DreamSim. The highest-scored image is placed at the top left, with scores decreasing progressively towards the bottom right. Note that for DreamSim similarity, we take the average of the similarities to all the reference images.
    The yellow boxes indicate failed images evaluated by human raters, for instance, mixed identity dogs or a dog is missing.
    We observe that D\&C score perfectly aligns with the human evaluation, yielding the failed images lower scores than the successful images.
    On the other hand, DreamSim does not align with human evaluation where the failed images are ranked high.
    }
    \label{fig:appendix_dnc_qualitative}
\vspace{-0.15in}
\end{figure*}

\subsection{Additional generated examples \label{app:exp:qual1}}
\begin{figure*}[ht!]
\centering
    \includegraphics[width=0.97\linewidth]{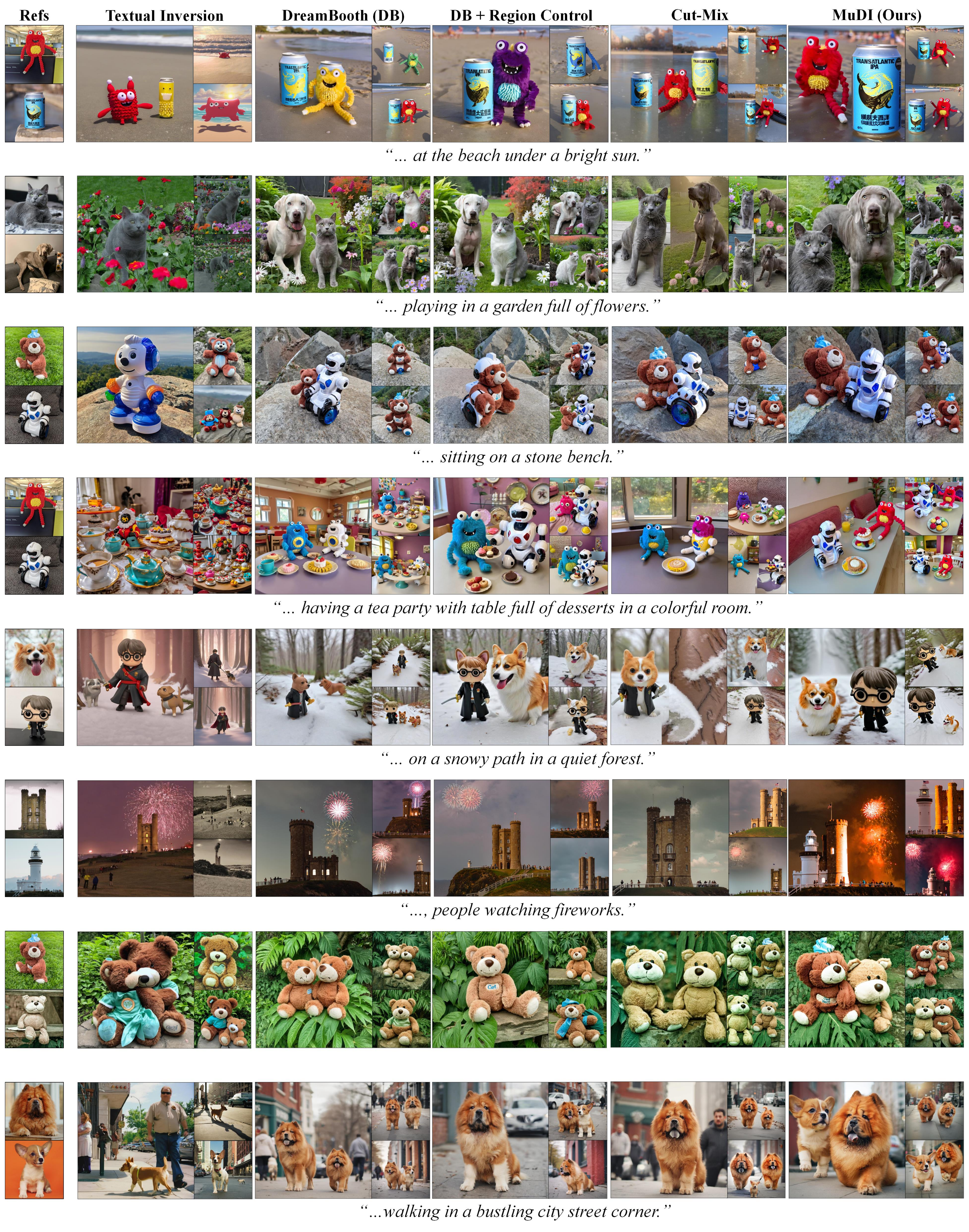}
\vspace{-0.1in} 
    \caption{\textbf{Qualitative comparison} of images generated by Textual Inversion~\citep{gal2022image}, DreamBooth (DB)~\citep{ruiz2023dreambooth}, DreamBooth with region control~\citep{gu2024mix}, Cut-Mix~\citep{han2023svdiff}, and our \metabbr. We visualize non-curated images generated with the same random seed.}
    \label{fig:appendix_qual1}
\vspace{-0.15in}
\end{figure*}

We provide additional non-curated generated examples in \autoref{fig:appendix_qual1}.

\clearpage
\section{Additional experimental results \label{app:add}}

\begin{figure*}[t!]
\centering
    \caption{\textbf{Qualitative comparison} of images generated by DreamBooth~\citep{ruiz2023dreambooth}, Custom Diffusion~\cite{kumari2023multi}, and \metabbr. 
    Similar to DreamBooth, Custom Diffusion results in identity mixing.}
    \label{fig:custom}
\vspace{-0.1in}
    \includegraphics[width=1\linewidth]{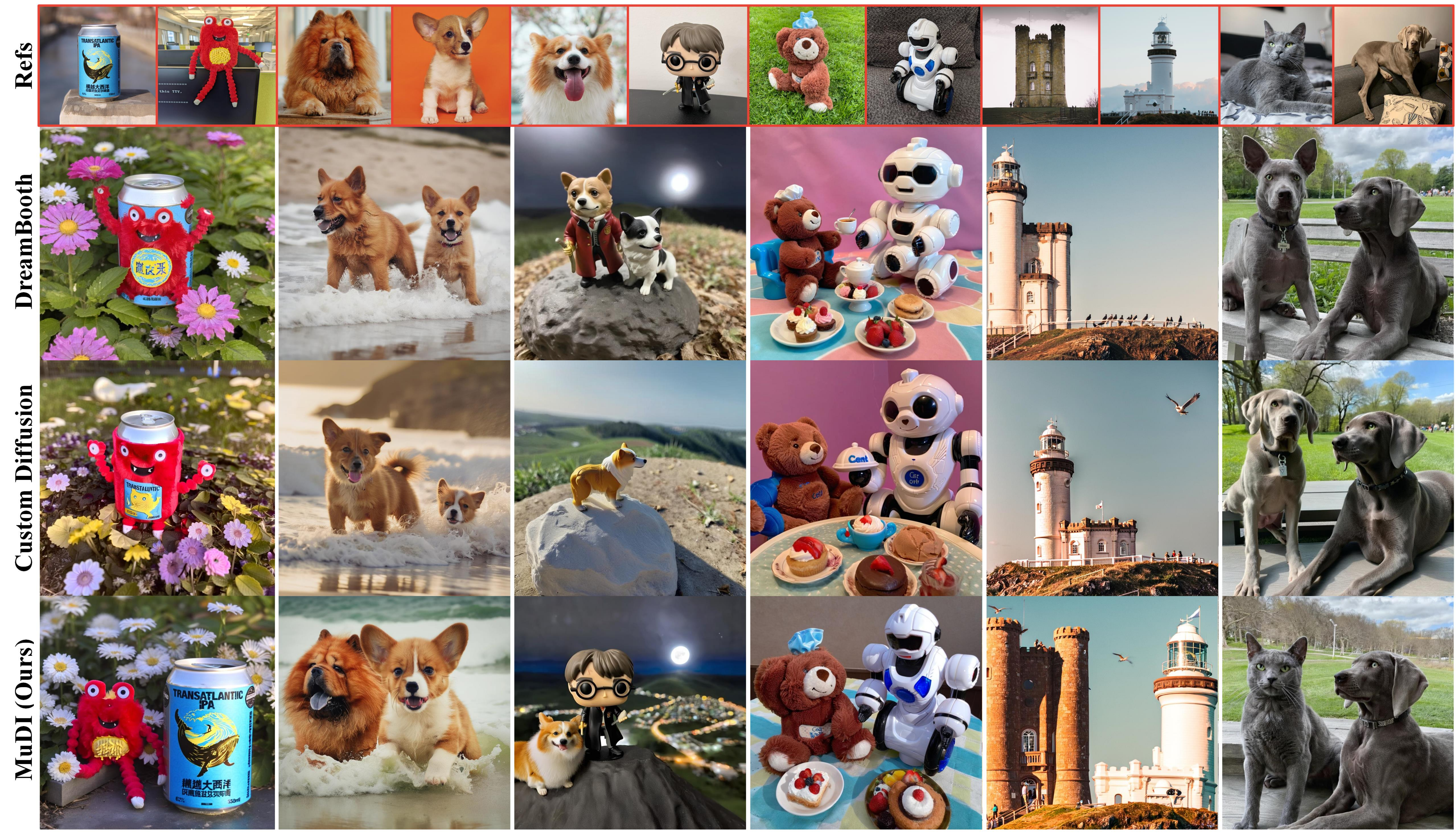}
\end{figure*}

\begin{table}[t!]
    \centering
    \resizebox{0.8\textwidth}{!}{
    \renewcommand{\arraystretch}{1.2}
    \renewcommand{\tabcolsep}{8pt}
\begin{tabular}{l c c c c}
\toprule
     & \multicolumn{2}{c}{Multi-Subject Fidelity} & \multicolumn{2}{c}{Text Fidelity} \\
\cmidrule(l{2pt}r{2pt}){2-3}
\cmidrule(l{2pt}r{2pt}){4-5}
    Method & D\&C-DS$\uparrow$ & D\&C-DINO$\uparrow$ & ImageReward$^\dagger$$\uparrow$ & CLIPs$^\dagger$$\uparrow$ \\
\midrule
    Textual Inversion~\citep{gal2022image} & 0.116 & 0.132 & -0.149 & 0.227 \\
    DreamBooth~\citep{ruiz2023dreambooth} & 0.371 & 0.388 & 0.579 & 0.255 \\
    Custom Diffusion~\citep{kumari2023multi} & 0.353 & 0.389 & 0.144 & 0.243 \\
    Cut-Mix~\citep{han2023svdiff} & 0.432 & 0.460 & -0.287 & 0.225\\
\midrule
    \metabbr (Ours) & \textbf{0.637} & \textbf{0.610} & \textbf{0.770} & \textbf{0.263}\\
\bottomrule
\end{tabular}}
\vspace{0.1in}
\caption{\textbf{Quantitative results} on multi-subject fidelity and text fidelity. $\dagger$ denotes the text fidelity score considering the permutation of the subjects in the prompt to avoid position bias.}\label{tab:eval_cd}
\vspace{-0.1in}
\end{table}


\subsection{Comparison with Custom Diffusion \label{app:add:custom}} 
Here, we provide the results of Custom Diffusion~\citep{kumari2023multi} that uses SDXL~\citep{podell2023sdxl} as the pre-trained text-to-image diffusion model. 
Due to GPU constraints, we fine-tune the weights of LoRA~\citep{hu2021lora} instead of directly fine-tuning the model weights. 
We evaluate two different models, one that uses a high rank (i.e., rank 128) and the other that uses the same rank as ours (i.e., rank 32). 
However, we do not observe significant differences between them.

As shown in \autoref{fig:custom}, Custom Diffusion demonstrates degradation in the subject fidelity compared to DreamBooth~\citep{ruiz2023dreambooth}.
The quantitative results in \autoref{tab:eval_cd} similarly show that Custom Diffusion results in lower multi-subject fidelity as well as lower text fidelity compared to DreamBooth and \metabbr.
Due to the degradation, we exclude Custom Diffusion from our baseline in the main experiments.

\begin{figure*}[t!]
\centering
    \caption{\textbf{Ablation study on Subject Overlap for our Seg-Mix}. We compare \metabbr against its variant trained with Seg-Mix which does not allow subject overlap during data augmentation, i.e., Seg-Mix w/o subject overlap. 
    \metabbr successfully personalizes the subjects with natural interaction. However, Seg-Mix without subject overlap results in identity mixing and subject ignorance.}
\vspace{-0.05in}
    \includegraphics[width=1\linewidth]{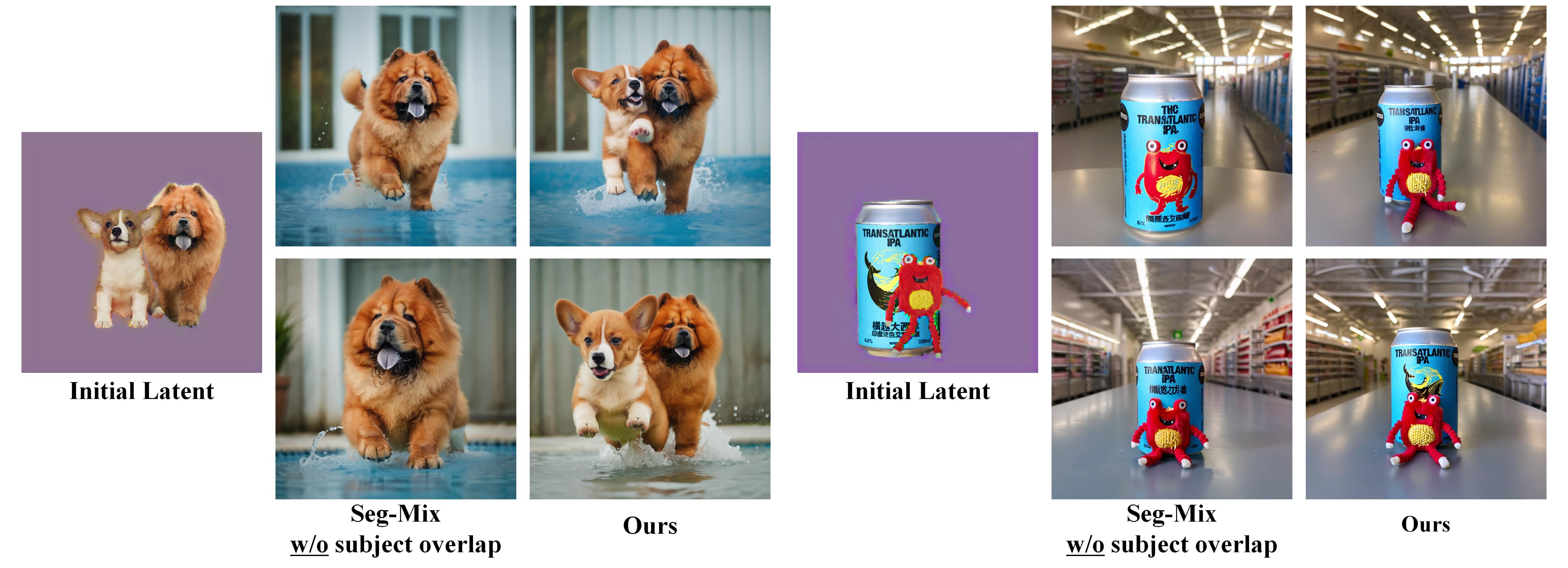}
\vspace{-0.1in} 
    \label{fig:smt_overlap}
\end{figure*}

\subsection{Importance of subject overlap in Seg-Mix \label{app:add:overlap}}
As described in Section~\ref{method:train}, our data augmentation method, Seg-Mix, allows subjects to be overlapped when randomly positioning the segmented subjects (see \autoref{fig:method} upper right). 
This differs from Cut-Mix~\citep{han2023svdiff} which is restricted to augmenting images of non-overlapped subjects.
Here, we verify that training the text-to-image models with images of overlapped subjects is crucial for generating interaction between the subjects.
In \autoref{fig:smt_overlap}, we qualitatively compare our \metabbr with its variant that is trained with Seg-Mix which does not allow subject overlap during data augmentation, namely \emph{Seg-Mix w/o subject overlap}.
Our \metabbr successfully personalizes the subjects distinctly with natural close-distance interaction, for example, two dogs playing in the pool.
In contrast, Seg-Mix w/o subject overlap produces mixed identities for neighboring subjects (e.g., monster toy in the can) or 
subject ignorance (e.g., generating only the Chow Chow while ignoring the Corgi).

\begin{figure*}[t!]
\centering
    \includegraphics[width=1\linewidth]{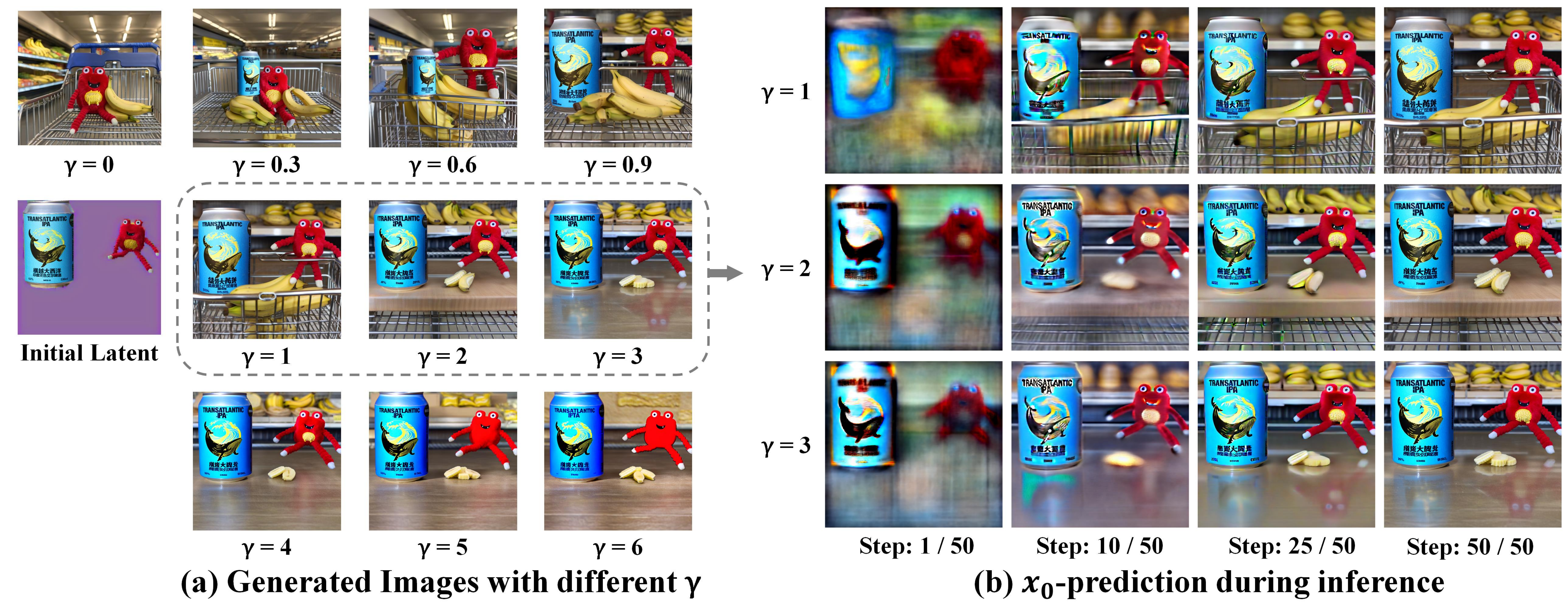}
\vspace{-0.1in} 
    \caption{\textbf{Analysis on $\bm{\gamma}$-scaling for inference initialization.}
    \textbf{(a) Generated Images for varying $\gamma$.} 
    A larger scale $\gamma$ results in more information preserved from the initial latent.  
    \textbf{(b) $\bm{x_0}$-prediction through inference steps.} 
    The image except for the fine details is determined in the first 10 steps. Thus providing information from the start via inference initialization plays a critical role in generating successful multi-subject composition.
    }
    \label{fig:appendix_gamma_ablation}
\vspace{-0.1in}
\end{figure*}

\subsection{Analysis on $\gamma$-scaling for inference initialization \label{app:add:gamma}}
Here, we analyze the effect of $\gamma$-scaling for our initialization by varying the magnitude of $\gamma$ for generating samples.
As demonstrated in \autoref{fig:appendix_gamma_ablation}(a), without inference initialization (i.e., $\gamma=0$) it results in an image independent of the initial latent, while the larger scale of $\gamma$ yields images with layouts of subjects similar to the initial latent.
Empirically, we observe that $\gamma$ exceeding 4 produces a highly saturated image with the same posture and layout as the initial latent.

In particular, we validate the reason for the effectiveness of our initialization by investigating the predicted clean image (i.e., $\bm{x}_0$) through the inference steps.
As shown in \autoref{fig:appendix_gamma_ablation}(b), we observe that the overall image except for the fine details is determined within the first 10 steps of the inference.
Therefore, our initialization is critical in generating successful multi-subject composition by providing information at the start.
Using a larger scale $\gamma$ yields more information that strongly affects the final image but also fixes the fine details such as postures and layouts.


\begin{figure*}[t!]
\vspace{-0.1in}
\centering
    \caption{\textbf{Diversity of images generated by \metabbr}. 
    \metabbr can generate images of personalized subjects in diverse postures not restricted to the initial latent of our inference initialization.
    Each visualized initial latent arranged in a 3$\times$3 grid corresponds to the generated image of the same position in the 3$\times$3 grid.
    All the images are generated from the same random seed using the prompt \emph{"... having a tea party with a table full of desserts in a colorful room."}
    }
    \label{fig:appendix_smp_posture}
\vspace{-0.05in}
    \includegraphics[width=0.95\linewidth]{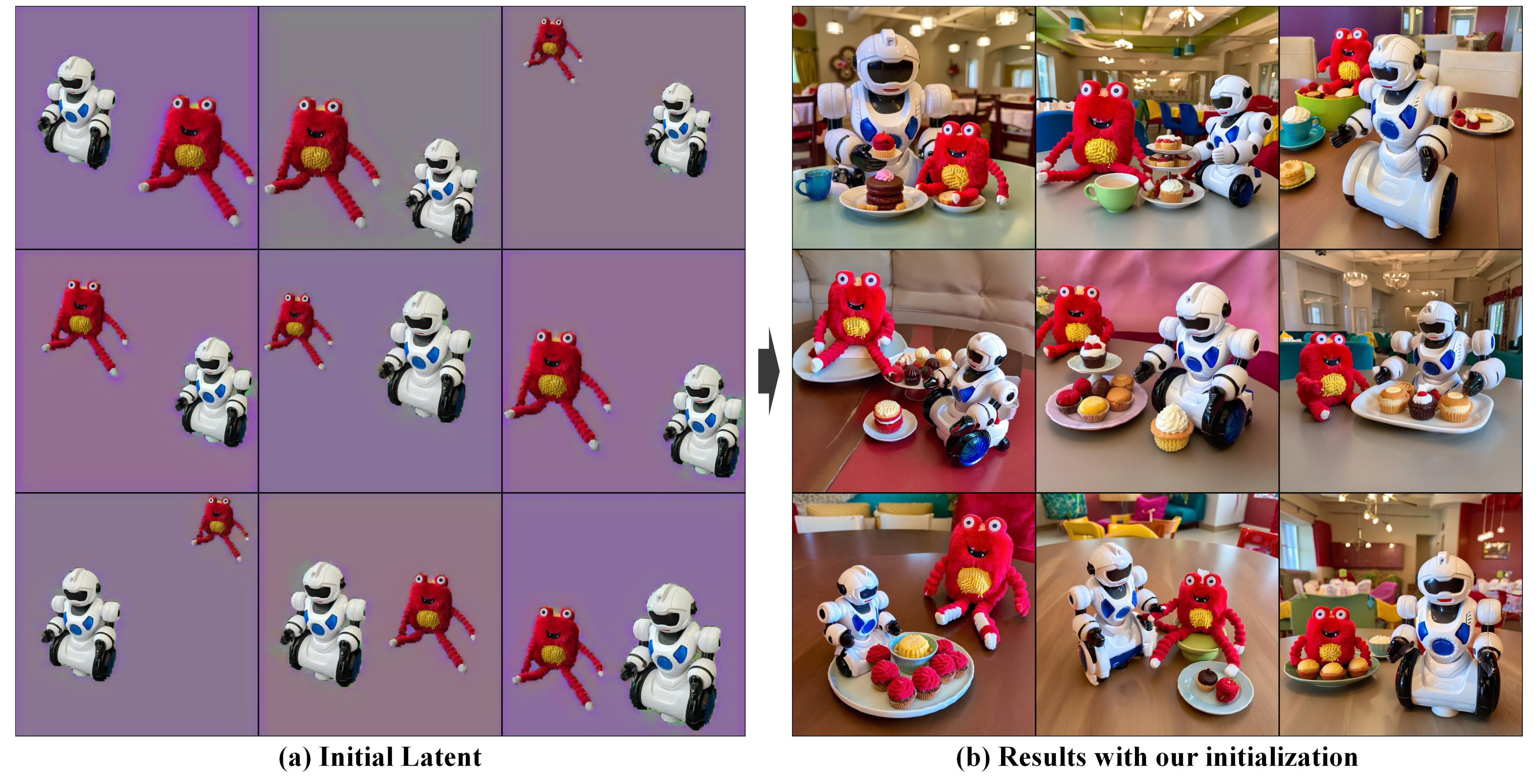} 
\vspace{-0.1in}
\end{figure*}

\subsection{Diversity of images generated by \metabbr \label{app:add:diversity}}
In \autoref{fig:appendix_smp_posture}, we demonstrate that \metabbr is able to generate images of personalized subjects in diverse postures not restricted by the initial latent of our inference initialization.
In particular, the generated subjects integrate smoothly with the background and exhibit natural interactions between the subjects.


\begin{figure*}[t!]
\centering
    \includegraphics[width=1\linewidth]{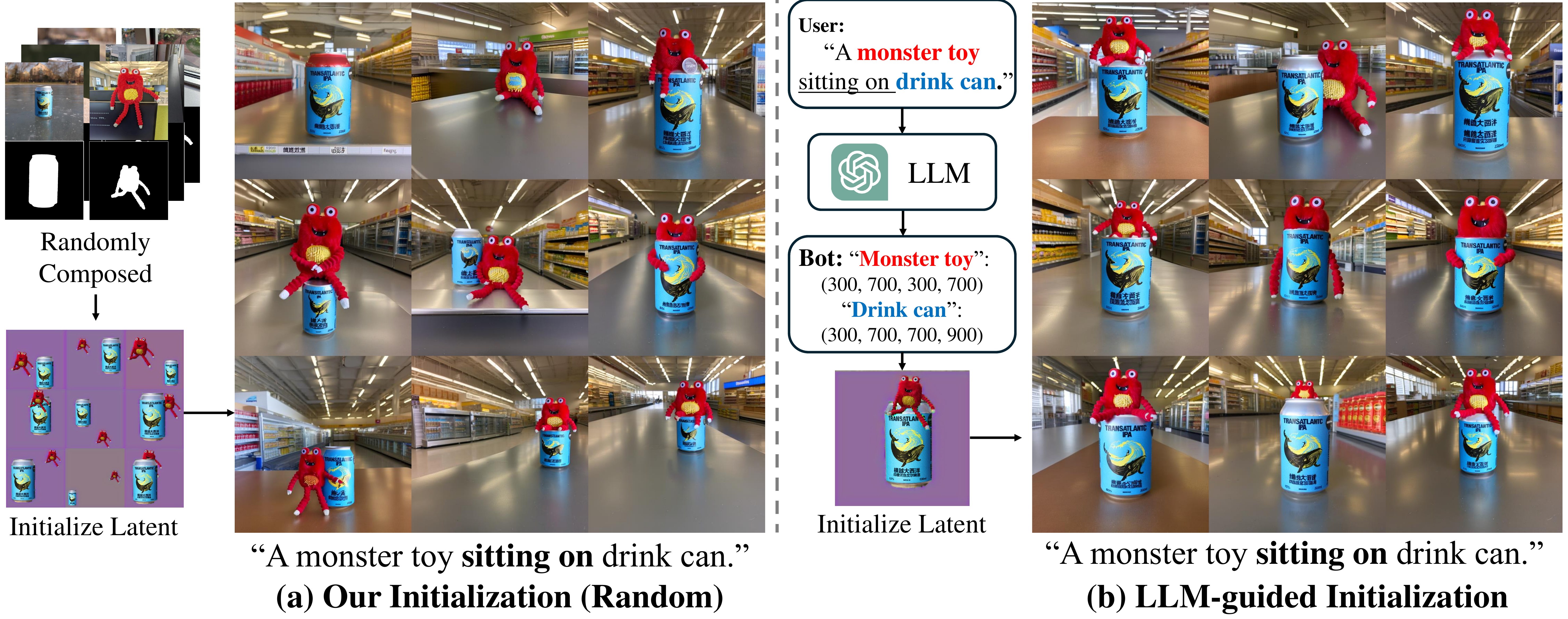}
\vspace{-0.1in} 
    \caption{\textbf{Examples of LLM-guided initialization for interactions.} 
    Latents and images located at the same position in each 3$\times$ 3 grid are paired. All images are generated from the same random seed. 
    \textbf{(a) Our initialization (Random).} 
    Prompts describing interactions, for example, "monster toy sitting on a drink can," may not fit the randomly created initial layouts. 
    Even though our initialization prevents identity mixing, the generated images may fail to reflect the interaction. 
    \textbf{(b) LLM-guided initialization.} 
    Instead of randomly positioning the segmented subjects, we utilize LLM to automatically generate prompt-aligned layouts for the inference initialization. 
    We find that LLM-guided initialization enables the generation of complex interactions between subjects.
    }
    \label{fig:appendix_LLM}
\end{figure*}

\subsection{LLM-guided initialization for interactions\label{app:samples:llm}}
Generating complex interactions involving relations like "[$V_1$] toy sitting on [$V_2$] can" can be challenging for personalized subjects. 
However, our initialization method can significantly assist this by providing a well-aligned layout reflecting the prompt, such as placing the toy above the can.

Inspired by \citet{Cho2023VisualP}, we utilize Large Language Models (LLMs) to generate prompt-aligned layouts of the segmented subjects.
The generated layouts are used instead of randomly created layouts for the inference initialization.
Such LLM-guided initialization enhances the ability to render complex interactions between subjects which we visualize in \autoref{fig:appendix_LLM}.

\begin{figure*}[t!]
\centering
    \caption{We describe three scenarios where our inference initialization provides significant advantages. 
    For each scenario, we visualize images generated with and without initialization on the left and the right columns respectively, where the image pairs (left and right) are generated using the same random seed.
    \textbf{(a) personalizing unusual subjects} that pre-trained models struggle to generate, such as cloud man. 
    \textbf{(b) personalizing more than two subjects}, in particular for similar subjects.
    \textbf{(c) using complex prompts} like \emph{"... as astronaut, floating on the moon, crater, space shuttle..."}.}
    \label{fig:appendix_smi_dominance}
    \includegraphics[width=1\linewidth]{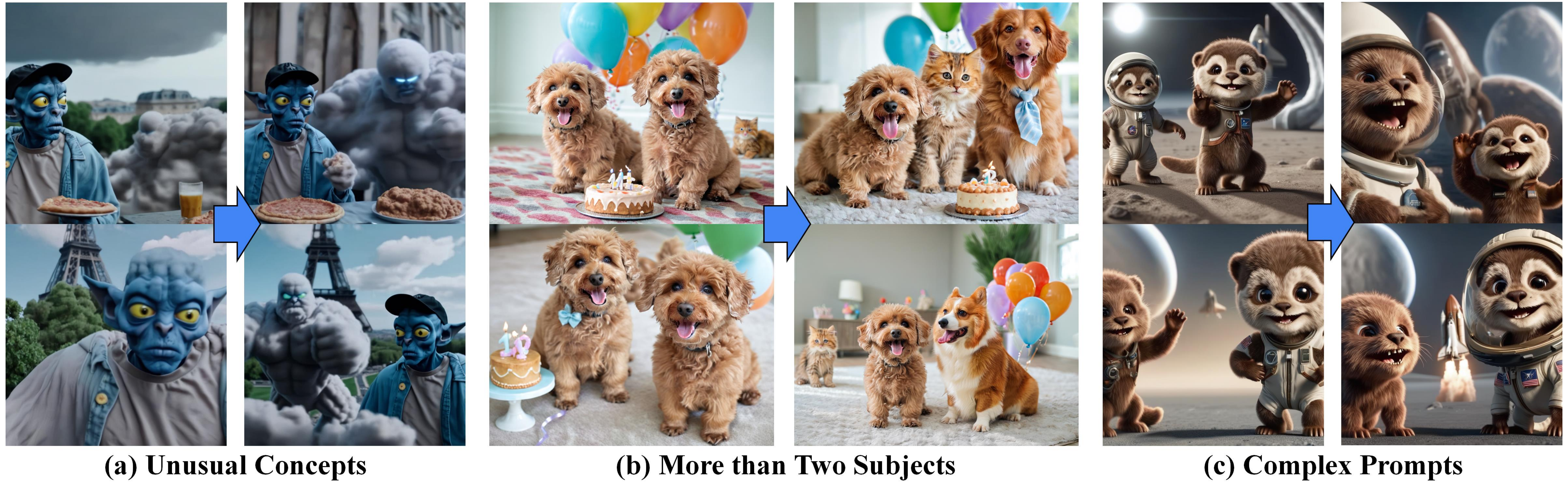}
\end{figure*}

\begin{figure*}[t!]
\centering
    \includegraphics[width=0.95\linewidth]{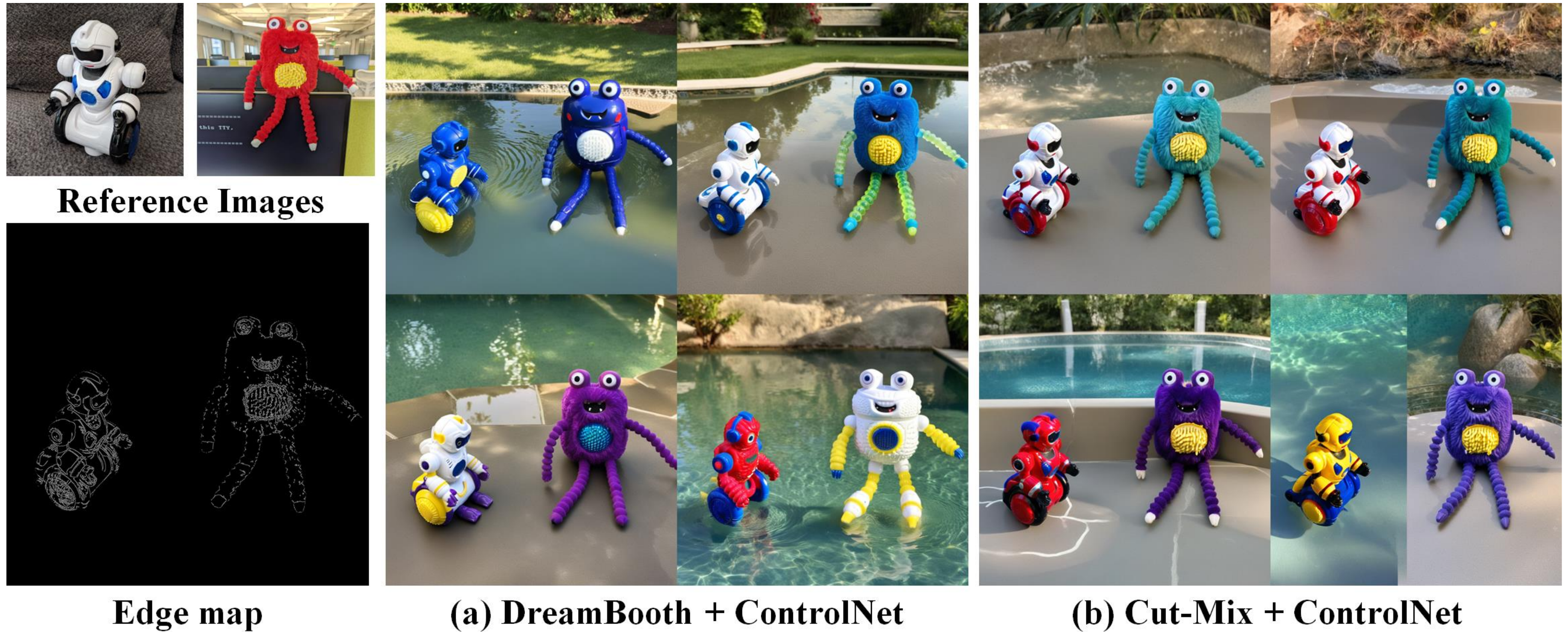}
\vspace{-0.0in} 
    \caption{Even with the strong spatial conditioning of ControlNet~\citep{zhang2023controlnet}, existing methods~\citep{ruiz2023dreambooth,han2023svdiff} suffer from identity mixing. The generated images show a monster toy with the robot-like body or a robot toy with the color of the monster toy.}
    \label{fig:appendix_controlnet}
\end{figure*}

\subsection{Importance of inference initialization \label{app:add:init}}
We explain in detail the scenarios where our inference initialization provides significant benefits.
First, as shown in \autoref{fig:appendix_smi_dominance}(a), unusual subjects that pre-trained models struggle to generate (e.g., the cloud man) are frequently ignored during generation without initialization. 
However, our initialization alleviates the ignorance of unusual subjects by guiding the model to consider all subjects starting from the initial latent.
Furthermore, initialization is crucial when personalizing more than two subjects as demonstrated in \autoref{fig:appendix_smi_dominance}(b). Generating images of more than two subjects without initialization often results in some subjects missing. It is almost impossible to compose many subjects together in an image without initialization.
Lastly, we observe that initialization plays an important role when the given prompt is complex as shown in \autoref{fig:appendix_smi_dominance}(c). Personalized diffusion models fail to generate images of the subjects when the prompt describes uncommon or highly detailed scenes, resulting in subjects of mixed identities or some subjects missing.
The inference initialization mitigates this problem by providing information on the subjects through the initial latent for which the model can focus more on rendering the scene described by the prompts. 
Our approach allows us to create images of personalized subjects, in particular new characters, in novel scenes.

\subsection{Multiple subject composition with ControlNet}
We validate that leveraging ControlNet~\citep{zhang2023controlnet} for previous approaches, for example, DreamBooth~\citep{ruiz2023dreambooth} and Cut-Mix~\citep{han2023svdiff}, fails to address identity mixing. 
\autoref{fig:appendix_controlnet} demonstrates that DreamBooth and Cut-Mix produce mixed identity toys even with the spatial conditioning of ControlNet.
We note that other types of layout conditioning based on cross-attention maps, for instance, the region control~\citep{gu2024mix}, do not alleviate identity mixing when using SDXL as the pre-trained model, as explained in Section~\ref{app:add:ca}.

\begin{figure*}[t!]
\vspace{-0.1in}
\centering
    \caption{\textbf{Inference initialization for pre-trained text-to-image models.} \textbf{(a)} For the unseen subjects, using initialization without personalization fails to preserve the details of the subjects, even initializing with a high gamma value (i.e., $\gamma=4$). 
    \textbf{(b)} For the known subjects, where the images are generated by the model, initialization still results in identity mixing.}
    \label{fig:appendix_init_wo_segmix}
    \vspace{-0.05in} 
    \includegraphics[width=1\linewidth]{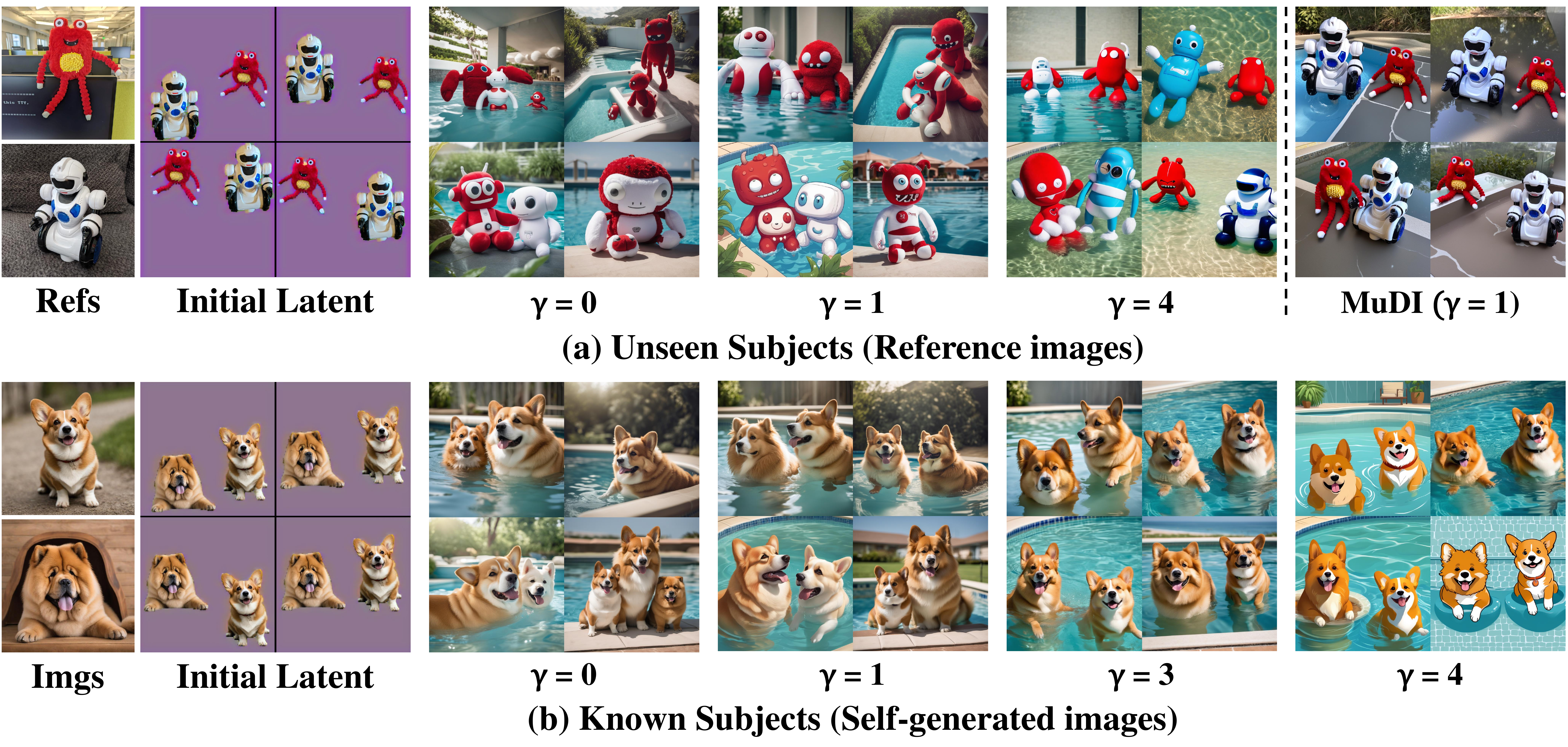}
\vspace{-0.3in}
\end{figure*}

\subsection{Inference initialization for pre-trained text-to-image model}
In \autoref{fig:appendix_init_wo_segmix}, we provide examples of inference initialization applied to the pre-trained text-to-image model.
When the initial latents are created from subjects that were not seen by the pre-trained model (monster toy and robot toy in \autoref{fig:appendix_init_wo_segmix}(a)), the model fails to generate the details of the subjects.
Only the layouts are preserved when using a high $\gamma$ scale for the initialization.
When the initial latent is created from known subjects (Corgi and Chow Chow in \autoref{fig:appendix_init_wo_segmix}(b)), the model results in identity mixing, even with a high $gamma$ scale. 
Therefore, we can observe that our inference initialization can only be effectively used to address identity mixing when the model is fine-tuned by our Seg-Mix.

\begin{figure*}[t!]
\vspace{-0.05in}
\centering
    \caption{\textbf{Qualitative comparison} of personalizing three subjects. 
    DreamBooth~\citep{ruiz2023dreambooth} suffers from severe identity mixing, especially for the two dogs. 
    Cut-Mix~\citep{han2023svdiff} also fails to generate three subjects often ignoring some subjects and producing stitching artifacts.
    In contrast, \metabbr successfully generates three personalized subjects without identity mixing that align with the given prompts.}
    \label{fig:appendix_three}
\vspace{-0.05in}
    \includegraphics[width=1\linewidth]{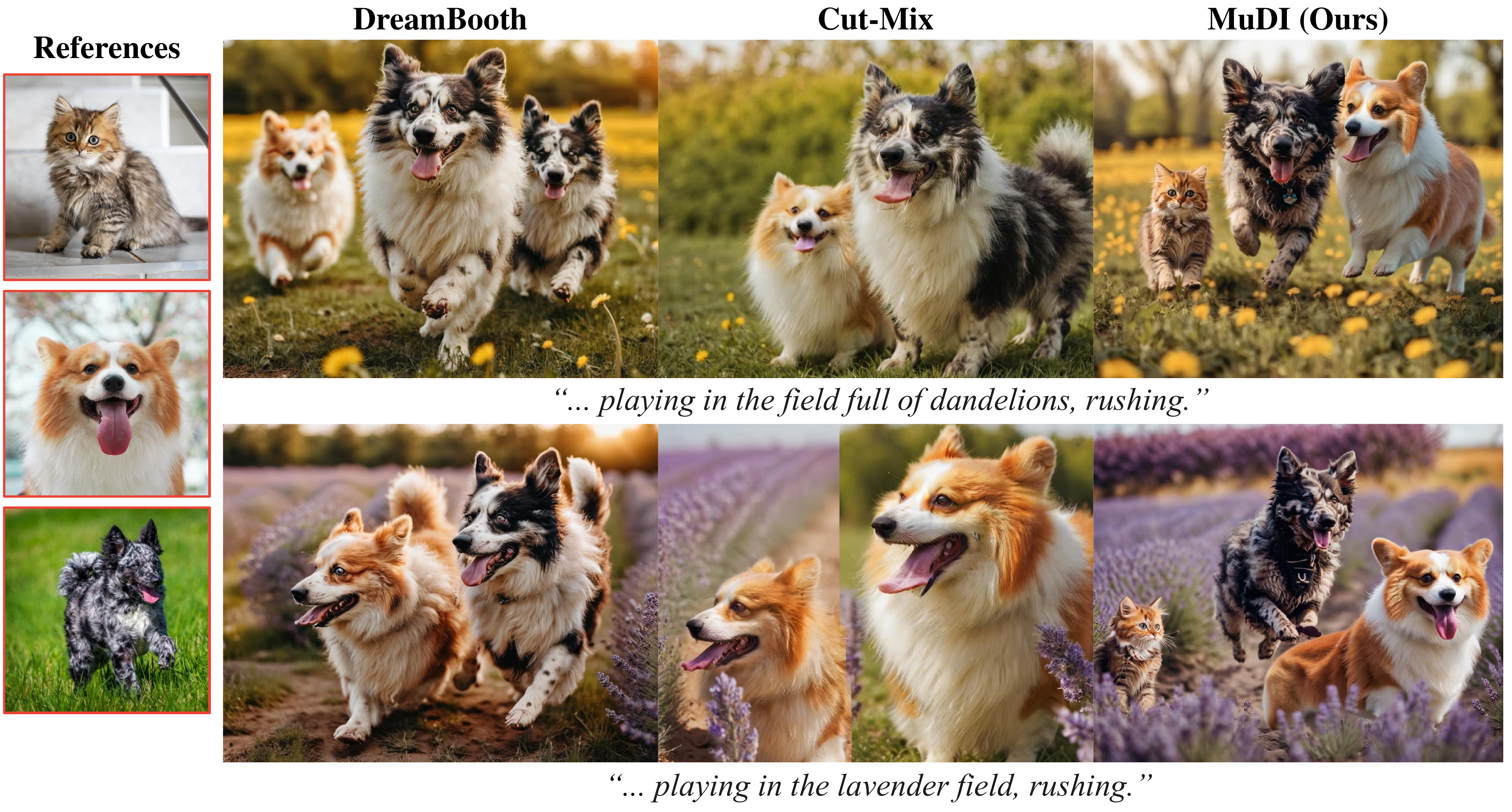}
\end{figure*}

\begin{figure*}[t!]
\centering
    \includegraphics[width=1\linewidth]{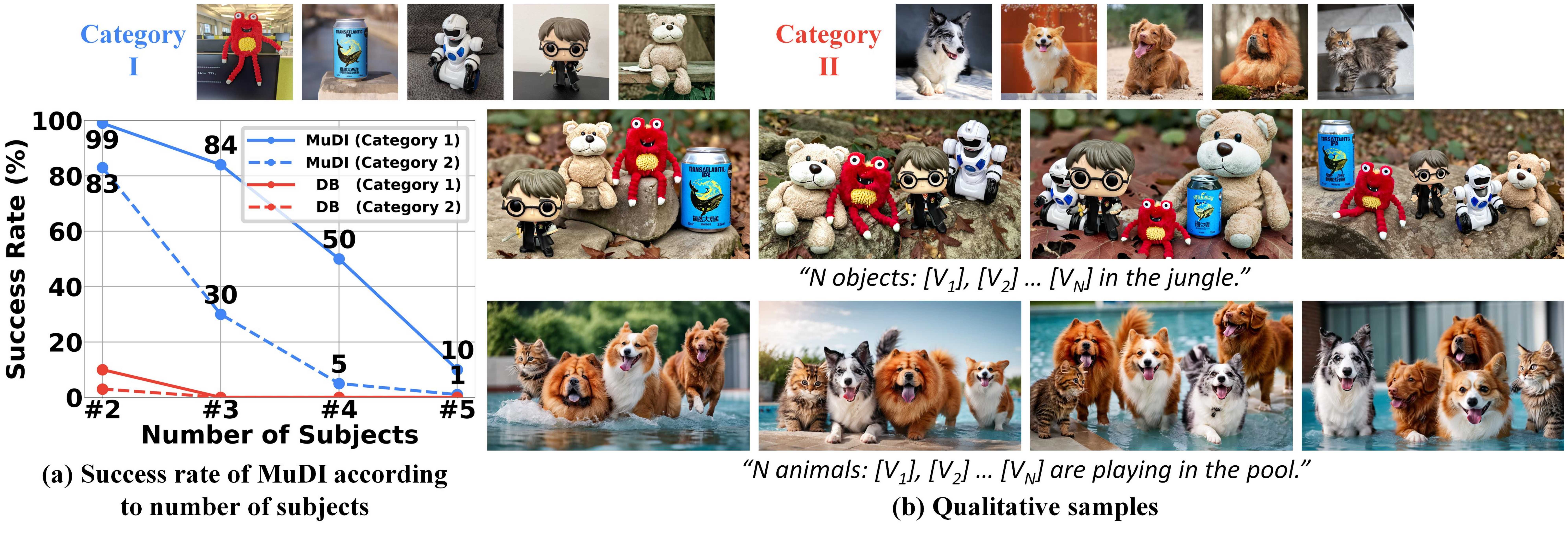}
\vspace{-0.2in} 
    \caption{\textbf{Analysis of the number of personalized subjects.}
    \textbf{(a) Success rate of MuDI according to number of subjects.} MuDI shows a higher success rate compared to the baseline, DreamBooth.
    \textbf{(b) Qualitative samples.} MuDI generates images with 4 and 5 subjects without identity mixing.
    }
    \label{fig:appendix_count_ablation}
\vspace{-0.05in}
\end{figure*}

\subsection{More than two subjects \label{app:add:many}}

\vspace{-0.05in}
\paragraph{Qualitative comparison}
In \autoref{fig:appendix_three}, we provide a qualitative comparison of previous approaches~\citep{ruiz2023dreambooth,han2023svdiff} and our \metabbr on personalizing three subjects. 
As DreamBooth~\citep{ruiz2023dreambooth} suffers from identity mixing even for two subjects, it fails to generate a composition of the three personalized subjects.
Cut-Mix~\citep{han2023svdiff} also produces mixed-identity dogs and often generates images of some subjects missing.
In contrast, \metabbr can successfully generate high-quality images of the dogs and the cat that align with the given prompts.

\vspace{-0.05in}
\paragraph{Number of personalized subjects}
We analyze the performance of \metabbr with respect to the number of subjects in \autoref{fig:appendix_count_ablation}. 
We used two types of datasets composed of five subjects where the first category consists of five objects (monster toy, drink can, robot toy, Harry Potter toy, and teddy bear), while the second category consists of five animals (four types of dogs and one type of cat). 
After fine-tuning the pre-trained text-to-image model for each category, we generated 600 images composing $N$ subjects for $N\in\{2,3,4,5\}$.
The images were generated using the prompts "[$V_1$], [$V_2$], ... [$V_N$] are in the jungle," for the objects and "[$V_1$], [$V_2$], ... [$V_N$] are playing together in the pool," for the animals. 
We consider all combinations and permutations of the subjects' order in the prompts uniformly.
In particular, we observe that adding "\emph{$N$ objects:}" or "\emph{$N$ animals:}" at the start of the prompt achieves a higher success rate.
We use a $\gamma$ scale of 2 for cases with two or three subjects, and a $\gamma$ scale of 3 for cases with more than three subjects.
The success rate was measured by first filtering the images using the D\&C scores and then evaluating the success by humans.

We report the success rate of the generated images with respect to the number of subjects in \autoref{fig:appendix_count_ablation}(a). 
DreamBooth completely fails to personalize more than two subjects for both categories.
In contrast, \metabbr achieves a significantly high success rate with the objects (Category I), showing over 50\% success for generating four subjects together without identity mixing.
While \metabbr shows a relatively lower success rate with the animals (Category II) due to the high similarities of the subjects, \metabbr can generate high-quality images of five animals with lively actions that align perfectly with the background. 
We visualize the successful images generated by \metabbr in \autoref{fig:appendix_count_ablation}(b).
However, we observe that the performance of \metabbr decreases as the number of personalized subjects increases, particularly for highly similar subjects.

\vspace{-0.05in}
\paragraph{Empirical findings}
We end this section by providing empirical findings for personalizing multiple subjects.
First, during training, we find it to be sufficient to augment images by composing only pairs of subjects, rather than composing three or more subjects together.
Furthermore, when personalizing more than three subjects, a higher augmentation probability is required during training compared to the case with two subjects. Also, a higher $\gamma$ value is needed during inference to generate all the subjects together.
Lastly, the prompt is crucial for generating multiple subjects. Prompts describing a detailed background or challenging actions may likely yield unsuccessful images.
Notably, adding detailed descriptions like "\emph{$N$ objects:}" at the start of the prompt results in a higher success rate.

\begin{figure*}[t!]
\vspace{-0.1in}
\centering
    \caption{\textbf{Visualization of cross-attention maps in SDXL}. \textbf{(a)} The token for the bear demonstrates a high value in the region corresponding to the bird, and the token for the bird takes a high value in an irrelevant location (right bottom). Note that this figure can be compared to Figure 4 of \citet{hertz2022prompt}.
    \textbf{(b)} The cross-attention maps of the identifier token (e.g., olis) do not show consistent results with the corresponding subject (i.e., monster toy in this example). We highlight the maps with black rectangles that have low values for the subject compared to the subject-irrelevant regions.}
    \label{fig:appendix_sdxl_attention}
\vspace{-0.05in}
    \includegraphics[width=1\linewidth]{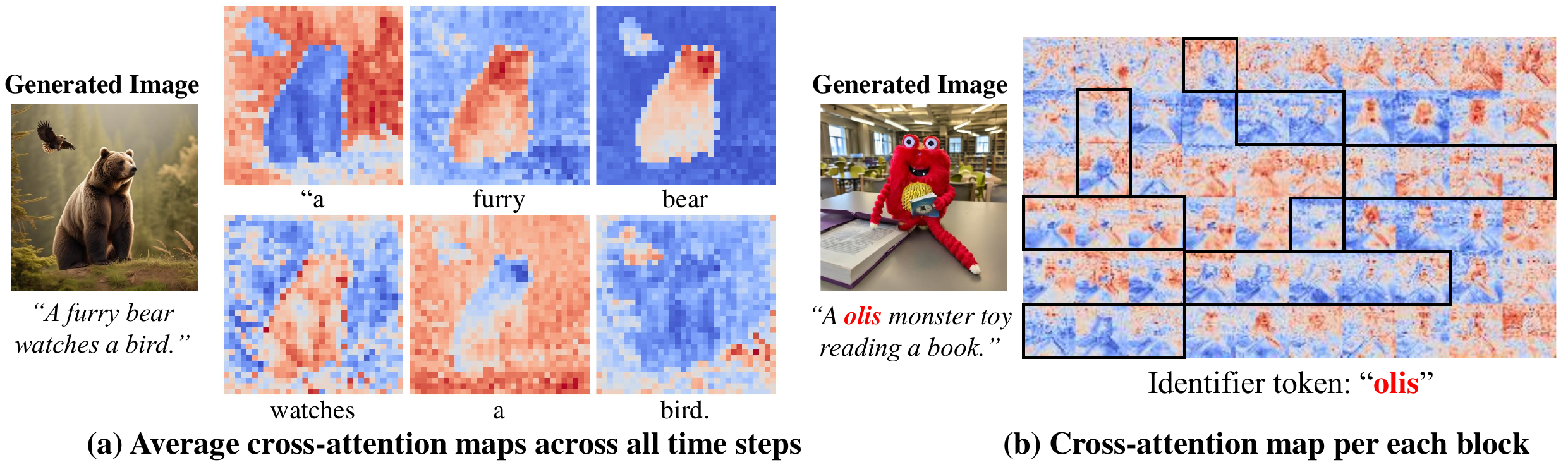}
\end{figure*}

\subsection{Analysis on cross-attention maps of SDXL \label{app:add:ca}}
Cross-attention maps have been widely used in prior works on image editing~\citep{hertz2022prompt}, layout-guided generation~\citep{liu2023cones2,gu2024mix}, single-subject personalization~\citep{wei2023elite,avrahami2023break}, and zero-shot multi-subject personalization~\citep{xiao2023fastcomposer} due to their controllability on the relation between the spatial layouts and the words in the prompt~\citep{hertz2022prompt}.
While the cross-attention maps worked successfully on previous diffusion models like Stable Diffusion (SD)~\citep{rombach2022ldm}, it is not the case for recent diffusion models such as Stable Diffusion XL (SDXL)~\citep{podell2023sdxl}.
The architectural design of SDXL, where an additional text condition is added to the time embedding~\citep{balaji2022ediff}, significantly reduces the consistency of the cross-attention maps which we demonstrate in \autoref{fig:appendix_sdxl_attention}.
Therefore, previous approaches based on the cross-attention maps~\citep{liu2023cones2,gu2024mix} are not directly applicable when using SDXL as a pre-trained text-to-image diffusion model.
For example, \citet{han2023svdiff} propose Unmix regularization, a technique that utilizes cross-attention maps to reduce stitching artifacts of the generated images, which we observe it to be ineffective for SDXL.

\begin{figure*}[t!]
\vspace{-0.05in}
\centering
    \includegraphics[width=0.9\linewidth]{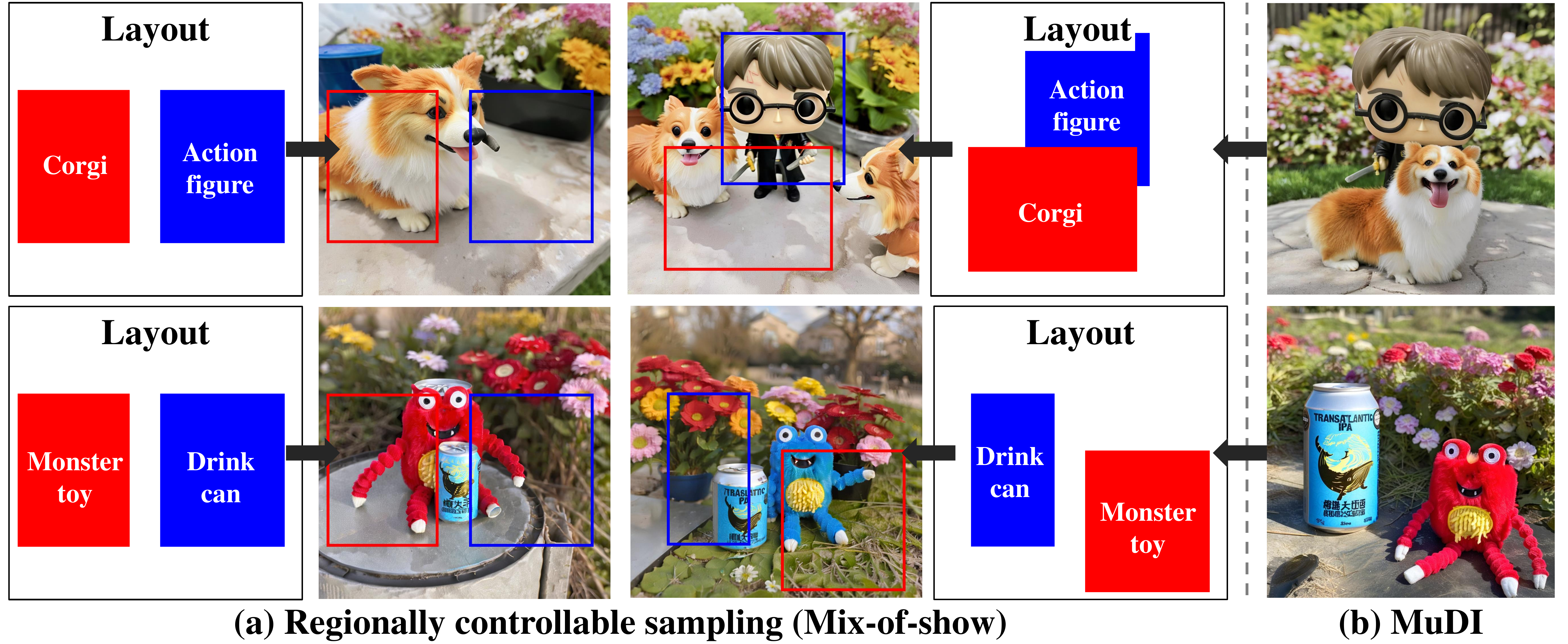}
    \caption{\textbf{Comparison of \metabbr and regionally controllable sampling~\citep{gu2024mix} for SDXL.} 
    \textbf{(a)} Regionally controllable sampling often results in missing subject or identity mixing. \textbf{(b)} \metabbr prevents identity mixing as well as subject missing.}
    \label{fig:appendix_mix_sdxl}
\vspace{-0.05in}
\end{figure*}

\subsection{Regionally controllable sampling for SDXL \label{app:add:region}}
To address identity mixing, \citet{gu2024mix} propose regionally controllable sampling, i.e., region control, which leverages multiple regional prompts and the corresponding cross-attention maps during inference.
However, manipulating the cross-attention map is not effective in preventing identity mixing for recent diffusion models like SDXL~\citep{podell2023sdxl}. 
We empirically observe that region control is highly likely to produce images with some subjects missing or having mixed identities, as demonstrated in \autoref{fig:appendix_mix_sdxl}(a).

On the other hand, our \metabbr can prevent both subject missing or identity mixing without using cross-attention maps, even in cases involving highly similar subjects or overlapping layouts. 
Notably, our initialization does not require additional computational overhead, in contrast to regional control which requires 1.6 times the inference time due to the high number of cross-attention blocks.

\begin{figure*}[t!]
\vspace{-0.1in}
\centering
    \caption{\textbf{Qualitative comparison} of images generated by Custom Diffusion~\citep{kumari2023multi}, Cones2~\citep{liu2023cones2}, Mix-of-Show~\citep{gu2024mix}, and \metabbr that use Stable Diffusion v2~\citep{rombach2022ldm} as a pre-trained text-to-image model. $\ast$ denotes that it used ControlNet~\citep{zhang2023controlnet} to generate images.}
    \label{fig:appendix_sd2_qual}
\vspace{-0.05in}
    \includegraphics[width=\linewidth]{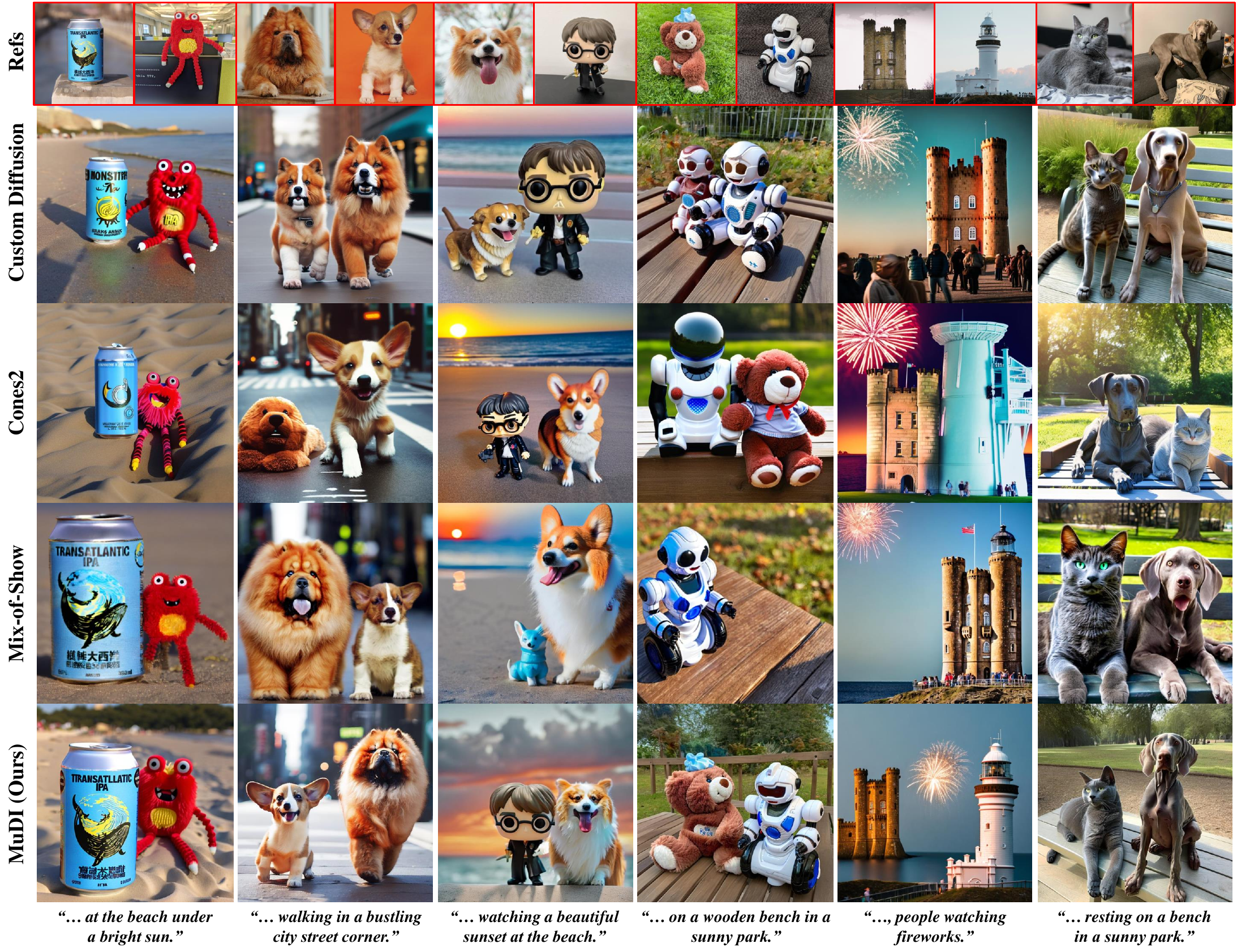}
\vspace{-0.2in}
\end{figure*}

\begin{table}[t!]
    \centering
    \resizebox{0.8\textwidth}{!}{
    \renewcommand{\arraystretch}{1.2}
    \renewcommand{\tabcolsep}{8pt}
\begin{tabular}{l c c c c c}
\toprule
     & \multicolumn{2}{c}{Multi-Subject Fidelity} & \multicolumn{2}{c}{Text Fidelity} & Speed \\
\cmidrule(l{2pt}r{2pt}){2-3}
\cmidrule(l{2pt}r{2pt}){4-5}
\cmidrule(l{2pt}r{2pt}){6-6}
    Method & D\&C-DS$\uparrow$ & D\&C-DINO$\uparrow$ & ImageReward$^\dagger$$\uparrow$ & CLIPs$^\dagger$$\uparrow$ & Time\\
\midrule
    Custom Diffusion~\citep{kumari2023multi} & 0.469 & 0.497 & 0.588 & 0.234 & \phantom{0.}1x\\
    Cones2~\citep{liu2023cones2} & 0.408 & 0.429 & 0.607 & \textbf{0.254} & 9.2x\\
    Mix-of-Show~\citep{gu2024mix} & 0.367 & 0.364 & 0.470 & 0.240 & 1.3x\\
    Mix-of-Show$^{*}$~\citep{gu2024mix} & 0.688 & \textbf{0.666} & 0.061 & 0.223 & 1.5x\\
    \midrule
    \textbf{\metabbr (Ours)} & \textbf{0.692} & 0.661 & \textbf{0.683} & 0.250 & 1.1x \\
\bottomrule
\end{tabular}}
\vspace{0.1in}
\caption{\textbf{Quantitative comparison using Stable Diffusion v2~\citep{rombach2022ldm}} as a pre-trained text-to-image model.
\textbf{$*$} indicates using ControlNet~\citep{zhang2023controlnet}. 
$\dagger$ denotes the text fidelity score considering the permutation of the subjects in the prompt to avoid position bias.
}
\label{tab:eval_sd2}
\vspace{-0.2in}
\end{table}

\subsection{Comparison with existing works using layout conditioning \label{app:add:sdv2}}
We compare our \metabbr with existing works on multi-subject composition using layout conditioning~\citep{gu2024mix,liu2023cones2}. 
We use Stable Diffusion v2~\citep{rombach2022ldm} as the pre-trained text-to-image diffusion model.
We create the layouts required for the baselines, Cones2~\citep{liu2023cones2} and Mix-of-Show~\citep{gu2024mix}, from the random initial latent of our inference initialization.
For a fair comparison, we did not use any image-based conditioning such as ControlNet~\citep{zhang2023controlnet} and T2I-adapter~\citep{mou2024t2iadapter} for Mix-of-Show.
We additionally report the results of Mix-of-Show using Canny edge ControlNet, namely Mix-of-Show$\ast$.
Note that Mix-of-Show demonstrates significant performance degradation when the size of the bounding boxes consisting of the layouts is not sufficiently large.
Therefore, we manually set the bounding boxes in the layout to be sufficiently large, for example, as the leftmost layout in \autoref{fig:appendix_mix_sdxl}.

As shown in \autoref{tab:eval_sd2}, \metabbr achieves the highest D\&C-DS scores as well as the ImageReward. 
Cones2 demonstrates low subject fidelity on unseen subjects, for example, the monster toy, as it is trained solely on text embeddings.
Mix-of-Show frequently generates images with some subjects missing, and results in low D\&C scores. 
When used with ControlNet, Mix-of-Show generally shows higher subject fidelity but often produces blurry backgrounds or images that are not aligned with the given prompts. 
We also report the relative inference time compared to the normal inference time using a pre-trained model for generating images with two subjects, as shown in \autoref{tab:eval_sd2}. 
We use the official codes available for each method and measure the inference time using a single RTX 3090 GPU. 
\metabbr achieves significantly faster inference speed compared to the methods based on cross-attention maps, namely Cones2 and Mix-of-Show.

Additionally, we provide a qualitative comparison in \autoref{fig:appendix_sd2_qual}. 
Custom diffusion~\citep{kumari2023multi} results in identity mixing for similar subjects, while Cones2 produces significantly low subject fidelity for unseen subjects such as the monster toy. 
Mix-of-Show generates images with some subjects missing
In contrast, our \metabbr successfully generates multi-subject images without identity mixing.

\begin{figure*}[t!]
\centering
    \includegraphics[width=1\linewidth]{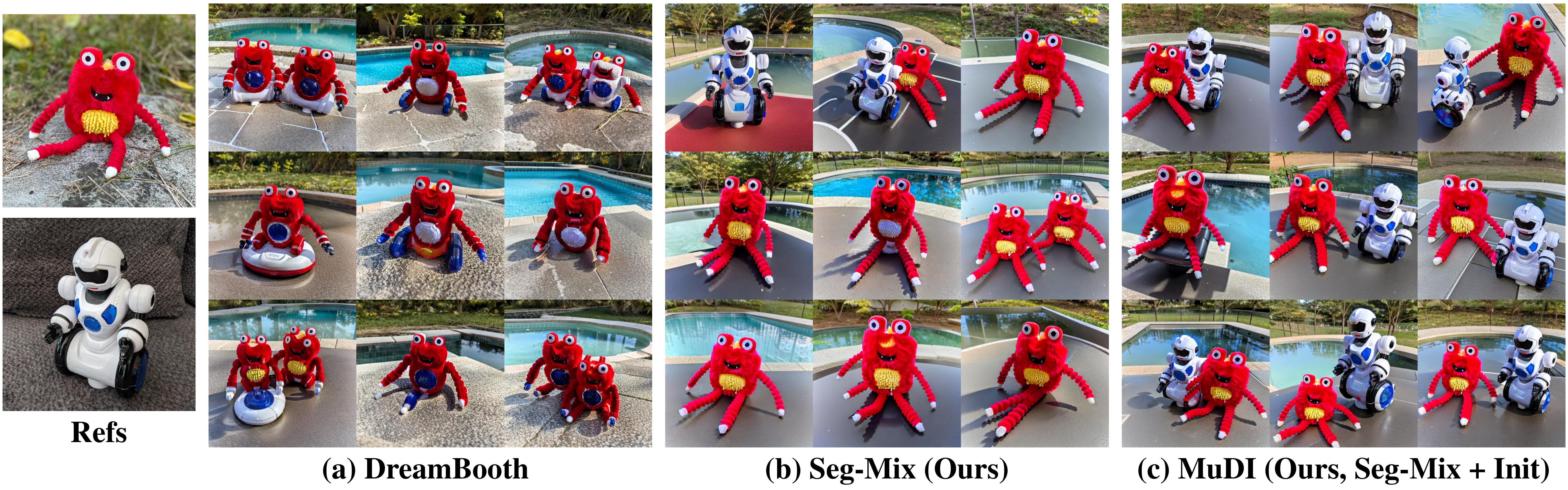}
    \caption{\textbf{Qualitative results using Stable Diffusion v1.5~\citep{rombach2022ldm}} as a pre-trained text-to-image model. 
    Similar to the case of using SDXL~\citep{podell2023sdxl} as a pre-trained model, DreamBooth~\citep{ruiz2023dreambooth} produces images of mixed-identity toys.
    Our Seg-Mix effectively addresses identity mixing but often generates images with the robot toy missing.
    In contrast, our \metabbr, which leverages both Seg-Mix and our inference initialization, successfully personalizes the subjects distinctly without identity mixing.
    Note that the images of the same positions in the 3$\times$3 grid are generated using the same random seed.
    }
    \label{fig:appendix_sd15}
\vspace{-0.1in}
\end{figure*}

\subsection{Stable Diffusion v1.5 as a pre-trained model \label{app:add:sdv1.5}}
In \autoref{fig:appendix_sd15}, we provide qualitative results of DreamBooth~\citep{ruiz2023dreambooth} and \metabbr using Stable Diffusion v1.5~\citep{rombach2022ldm} as the pre-trained text-to-image diffusion model.
Similar to the case when using SDXL~\citep{podell2023sdxl} as the pre-trained model, DreamBooth results in identity mixing.
Our Seg-Mix effectively addresses identity mixing but often generates images of a subject missing.
In contrast, our \metabbr which leverages both Seg-Mix and inference initialization successfully personalizes the subjects without identity mixing or subject ignorance.

\begin{figure*}[t!]
\centering
    \caption{\textbf{Examples of relative size control using Seg-Mix}. During training, we augment the images of segmented subjects with fixed relative sizes.
    \textbf{(a)} When the relative size of the toy and the can is equal (i.e., toy:can=1:1), the generated samples display a toy and a can of similar size. 
    \textbf{(b)} When we set the relative size of the toy to be smaller than the can (i.e., toy:can=1:2), the generated samples display a relatively small toy compared to the can.
    Note that the images of the same positions in the $3\time3$ grid are generated using the same random seed.
    }
    \label{fig:appendix_relative_size}
\vspace{-0.05in}
    \includegraphics[width=1\linewidth]{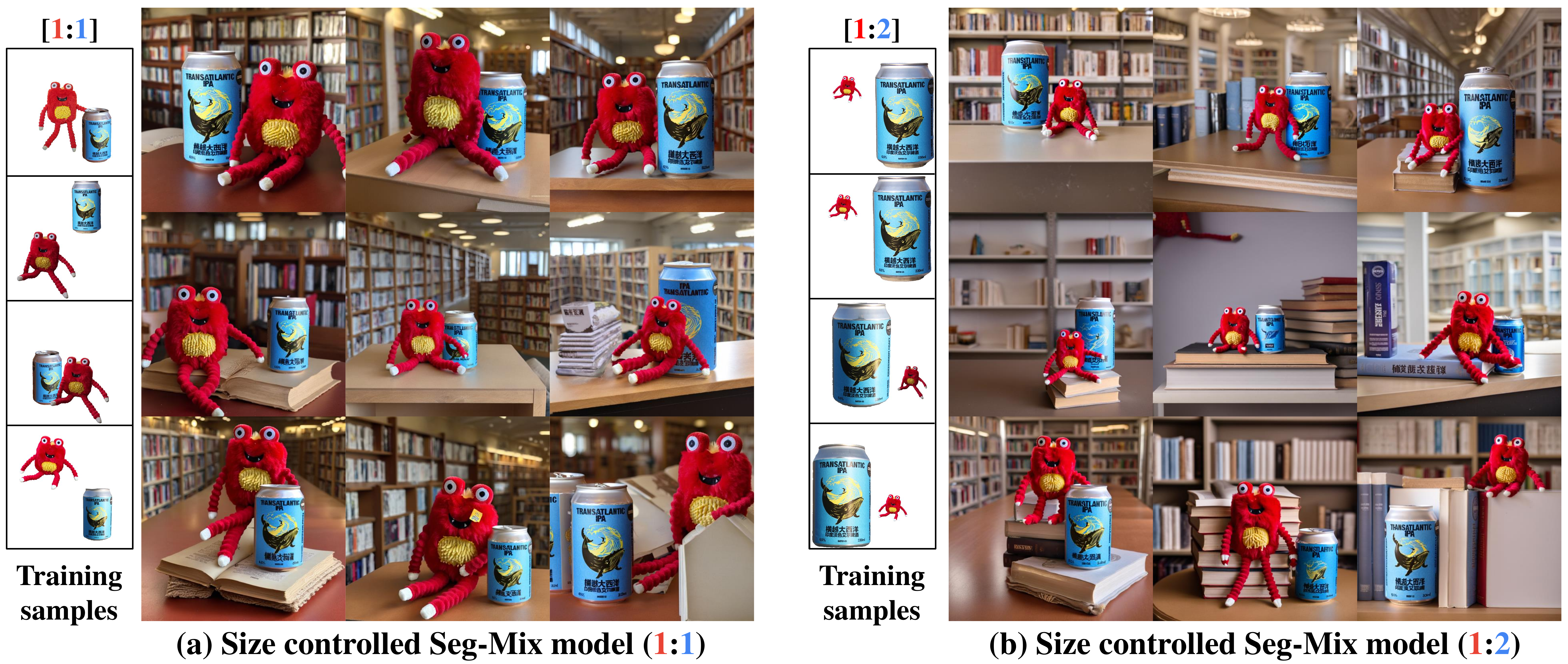}
\end{figure*}

\section{Other use cases \label{app:applications}}

\subsection{Relative size control \label{app:samples:size}}
\metabbr enables control of relative size between the personalized subjects.
During training, we can augment the images of segmented subjects with fixed relative sizes according to user intents, instead of random relative sizes.
This corresponds to setting the argument \emph{scales} of the function \emph{create\_seg\_mix} in \autoref{alg:seg-mix}.
As shown in \autoref{fig:appendix_relative_size}, we can personalize models to generate the toy to be larger than the can or vice versa.
We observe a consistent relative size of the personalized subjects in the generated images.

\begin{figure*}[t!]
\centering
    \includegraphics[width=1\linewidth]{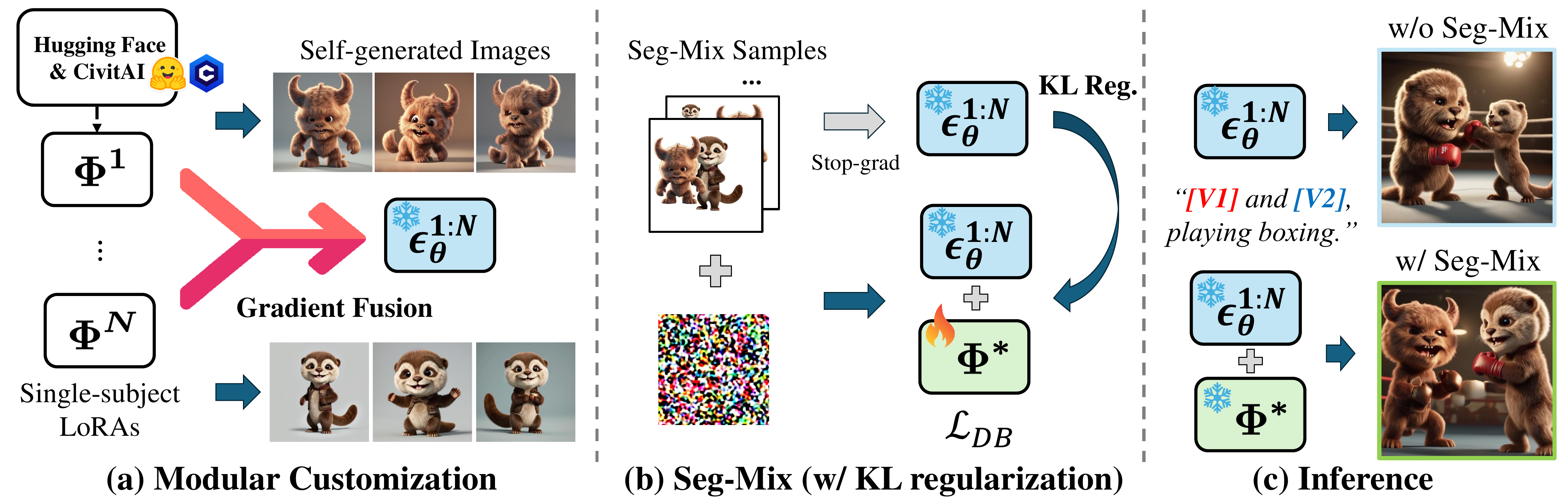}
\vspace{-0.1in} 
    \caption{\textbf{Modular customization with Seg-Mix}. \textbf{(a)} We first generate single-subject images using the pre-trained LoRAs ($\Phi^{i}$), and then merge the LoRAs using the gradient fusion~\citep{gu2024mix} to obtain a fused model $\epsilon^{1:N}_{\theta}$.
    \textbf{(b)} We use the self-generated images to train additional LoRA for identity decoupling via Seg-Mix. We add KL regularization to the training objective to prevent overfitting and saturation. 
    We only train for 200-300 iterations.
    \textbf{(c)} While the fused model results in mixed-identity characters, our Seg-Mix fine-tuning effectively addresses identity mixing.
    }
    \label{fig:appendix_modular}
\vspace{-0.05in}
\end{figure*}

\subsection{Modular customization \label{app:samples:modular}}
Our Seg-Mix can also be applied to modular customization, i.e., when we possess single-subject LoRAs that have been independently fine-tuned to each subject.
Instead of re-training the models each time for new combinations of subjects, we can efficiently merge the pre-trained models that are independently fine-tuned for each subject, avoiding the need for training from the beginning.

We first generate images of each subject using their corresponding LoRA, which are subsequently utilized as reference images. 
We then merge the single-subject fine-tuned LoRAs~\citep{hu2021lora} using an existing method such as gradient fusion~\citep{gu2024mix} to obtain a fused model (\autoref{fig:appendix_modular}(a)).
While the fused model can successfully generate each subject individually, composing multiple subjects results in severe identity mixing.
Therefore, we apply Seg-Mix with the generated single-subject images for 200-300 iterations (\autoref{fig:appendix_modular}(b)). 
When applying Seg-Mix, we apply a new LoRA to fine-tune the fused model and set the seg-mix probability to 1. This is because the fused model has already been trained with each subject and only needs to be trained with the composition of the subjects.
In particular, we add KL regularization~\citep{fan2024dpok} to the personalization objective (Eq.~\eqref{eq:dreambooth}) in order to prevent overfitting and saturation. 
Our approach effectively reduces identity mixing as shown in \autoref{fig:appendix_modular}(c).

Notably, we find that for certain subjects, using self-generated images instead of the original reference images for Seg-Mix fine-tuning achieved superior performance, especially alleviating posture overfitting.
For example, the reference images for the characters from Sora~\citep{Sora} (e.g., the otter) are obtained from the video frames that have highly limited postures.
Instead of using the reference images directly, we can generate diverse images of the subjects using the single-subject LoRAs, and use them when applying Seg-Mix.

\begin{figure*}[t!]
\centering
    \includegraphics[width=0.95\linewidth]{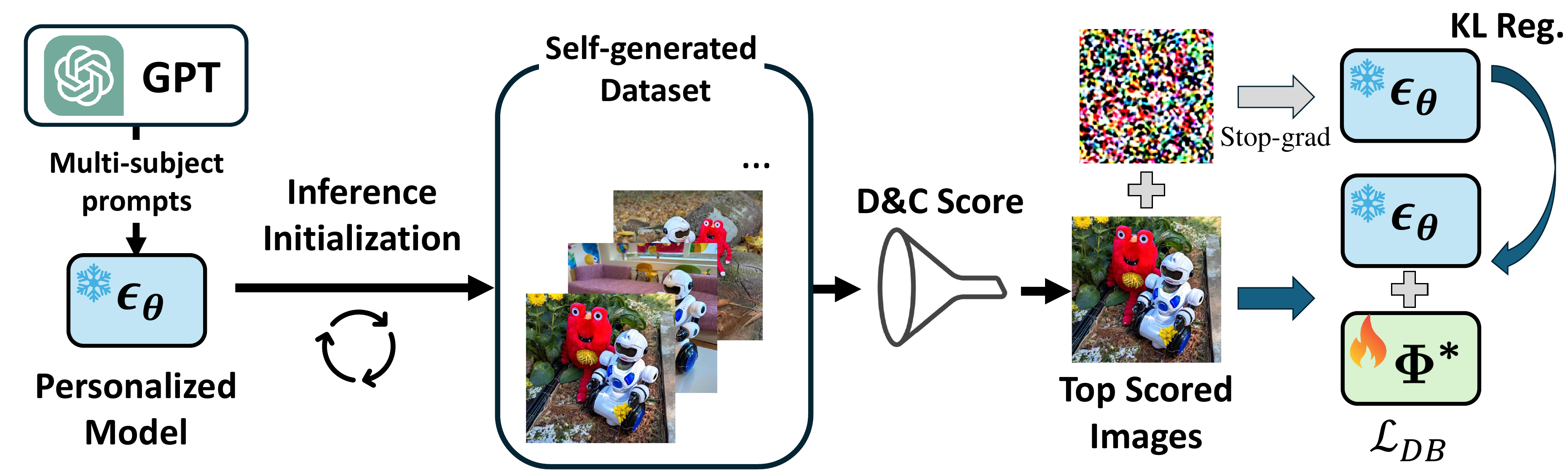}
    \caption{\textbf{Illustration of Iterative Training for \metabbr.} 
    } 
    \label{fig:appendix_sft_overview}
\end{figure*}

\begin{figure*}[t!]
\centering
    \includegraphics[width=1\linewidth]{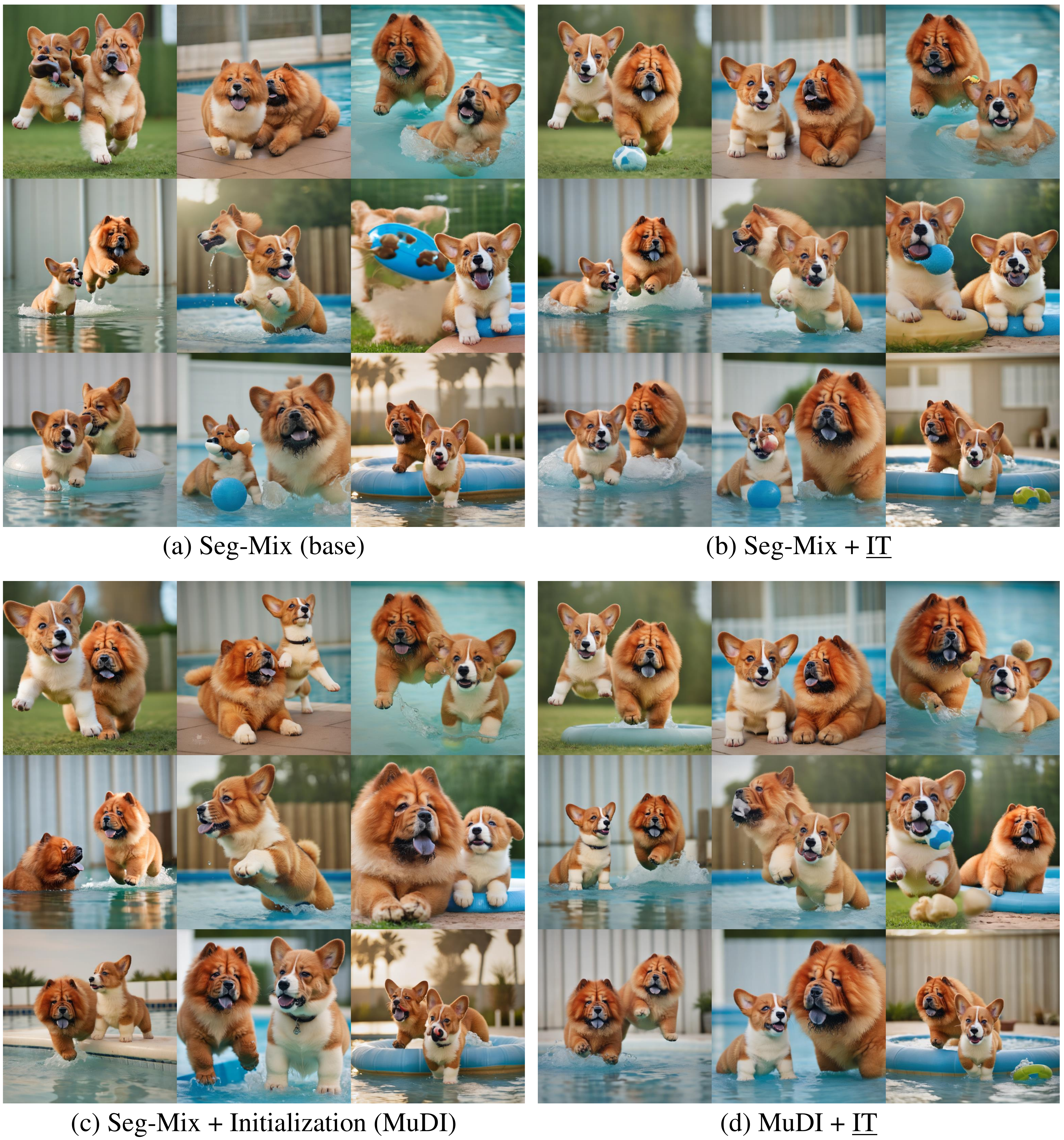}
\vspace{-0.15in} 
    \caption{\textbf{Qualitative comparison} of our iterative training (IT). 
    Images at the same position in each 3$\times$3 grid are generated from the same random seed. 
    \textbf{(a) Seg-Mix training} without initialization does not perfectly address identity mixing.
    \textbf{(b) Iterative training} without initialization shows improvement compared to the Seg-Mix training.
    \textbf{(c) \metabbr} addresses identity mixing and subject missing, but occasionally fails to decouple highly similar subjects. 
    \textbf{(d) \metabbr with iterative training} successfully personalizes multiple subjects that are highly similar.
    }
    \label{fig:appendix_SFT}
\end{figure*}


\subsection{Iterative training \label{app:samples:sft}}
We present an iterative training (IT) method~\citep{sohn2023styledrop} for \metabbr to further improve the image quality.
The key idea is to additionally fine-tune the personalized model using high-quality images generated with \metabbr. 
We introduce a fully automatic training based on our \metabbr and the D\&C score which closely aligns with human evaluation for the multi-subject fidelity.

To be specific, we generate 200 multi-subject images with \metabbr using simple prompts created by ChatGPT~\citep{chatgpt}, for example, "[$V1$] dog and [$V2$] dog watching birds from a window sill.". The top 50 images based on the D\&C-DS score are used to fine-tune the personalized model using LoRA~\citep{hu2021lora}.
We fine-tune the model with the KL regularization~\citep{fan2024dpok} added to the personalization objective $\mathcal{L_{DB}}$ of Eq.~\eqref{eq:dreambooth}. 
In particular, we observe that the KL regularization is crucial for preventing saturation and preserving the image quality of the self-generated images.
Empirically, setting the KL regularization weight as 1.0 results in a good trade-off between preventing saturation and multi-subject fidelity. 
We provide an overview of the iterative training framework in \autoref{fig:appendix_sft_overview}.

As shown in \autoref{fig:appendix_SFT}, IT considerably improves \metabbr on personalizing highly similar subjects.
We believe using other RL-based fine-tuning methods such as Direct Preference Optimization (DPO)~\citep{rafailov2024direct,wallace2023dpo} instead of our supervised fine-tuning approach, would be more robust against over-saturation and better to reflect human preferences. 
Additionally, combining different reward models with RL methods could further improve how well the system aligns with human preferences, which we leave for future work.

\begin{figure*}[t!]
\centering
    \includegraphics[width=0.90\linewidth]{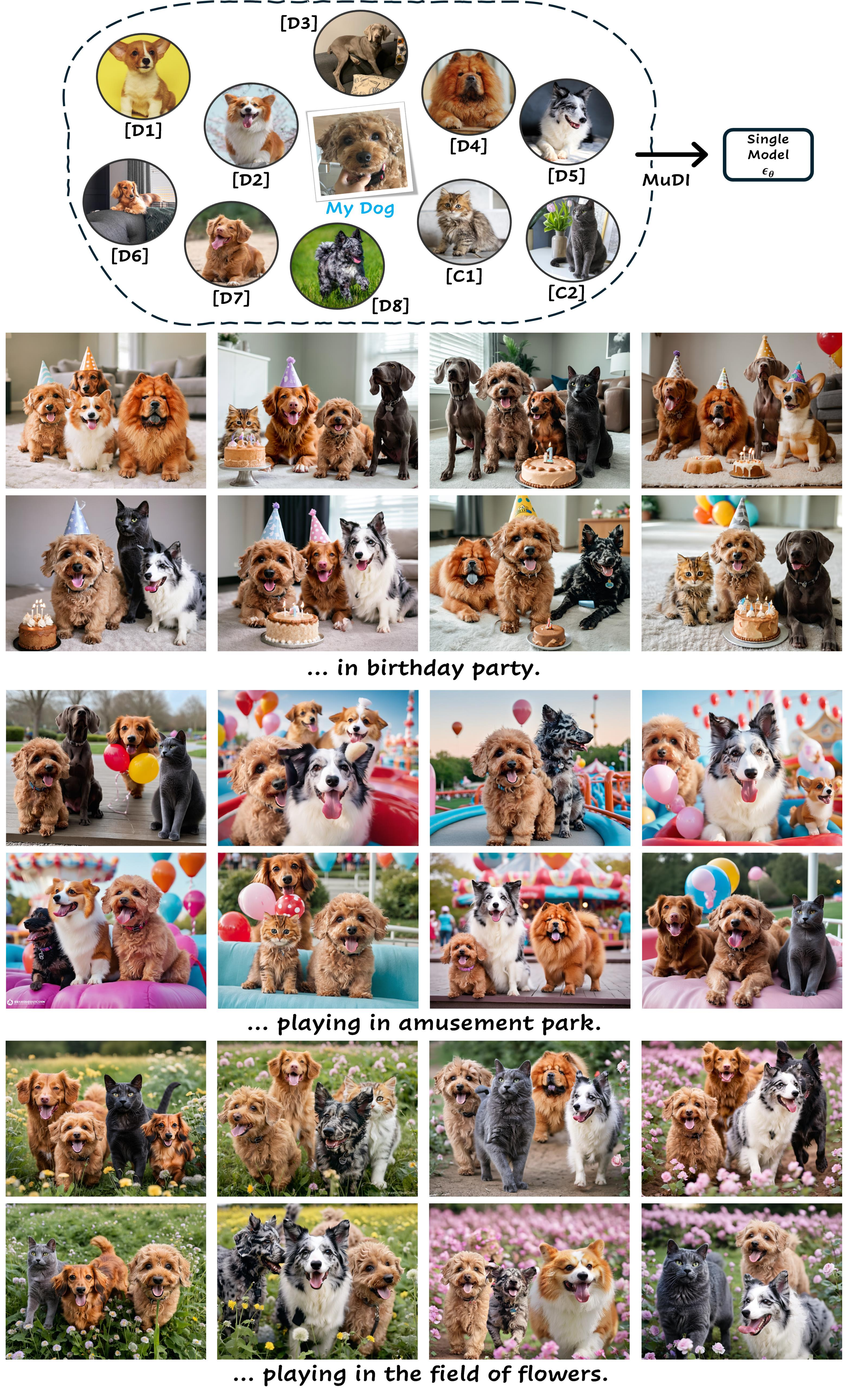}
    \caption{\textbf{Personalizing 11 concepts together with \metabbr using a single LoRA~\citep{hu2021lora}.} 
    We use descriptive classes for each dog and cat, for example, Weimaraner or Mudi, which enhances the ability to personalize multiple subjects that are highly similar. 
    }
    \label{fig:appendix_dog_friends}
\vspace{-0.15in}
\end{figure*}

\begin{figure*}[t!]
\centering
    \includegraphics[width=\linewidth]{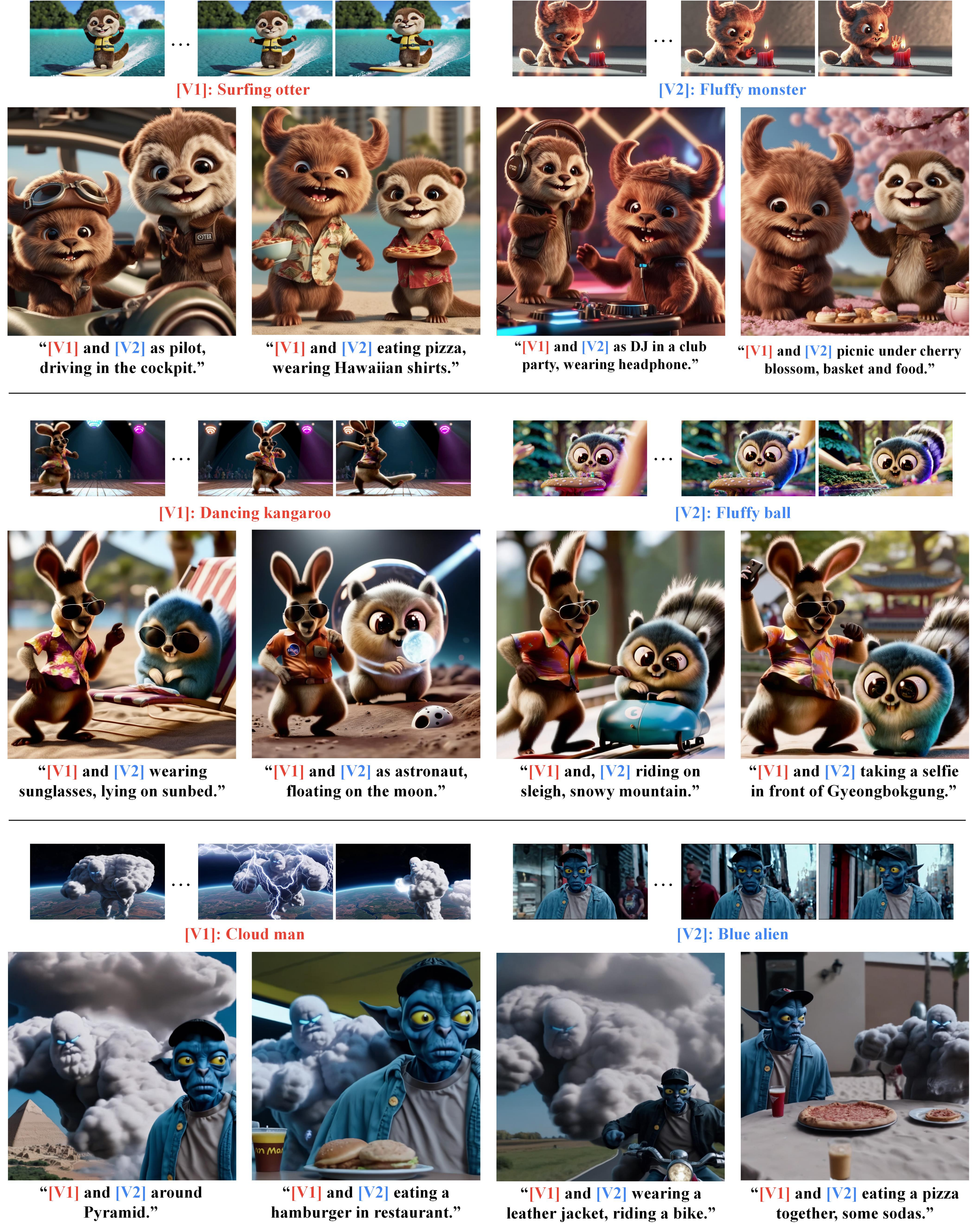}
    \caption{\textbf{Generated samples from \metabbr by personalizing characters from Sora~\citep{Sora}}.
    }
    \label{fig:appendix_sora}
\end{figure*}

\FloatBarrier
\newpage

\begin{figure*}[t!]
\centering
    \includegraphics[width=1\linewidth]{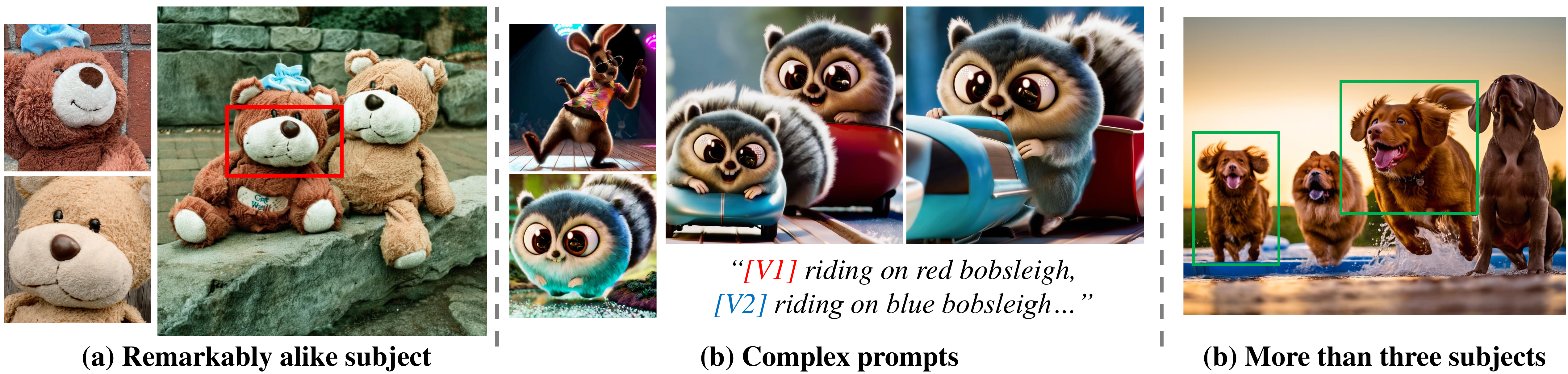}
\vspace{-0.1in} 
    \caption{\textbf{Limitations.} \textbf{(a) Remarkably alike subjects} are challenging to decouple perfectly as they lie very close in the latent space. 
    For example, two brown teddy bears can be easily mixed up as they have highly similar designs and colors.
    \textbf{(b) Complex prompts} that describe unusual or detailed scenes bring additional difficulty in preserving the details of the subjects. In this case, the subjects can be easily ignored during the generation.
    \textbf{(c) More than three subjects.}
    \metabbr significantly mitigates identity mixing but often duplicates the same subjects in the generated images.
    }
    \label{fig:limitation}
\end{figure*}

\section{Limitations and societal impacts \label{app:limitation}}
\paragraph{Limitations}
We find that decoupling the identities of remarkably alike subjects is still challenging even for our method, for example, two brown teddy bears in our dataset (see \autoref{fig:limitation}(a)).
Such subjects are very close in the image latent space which may be difficult to separate with the current text-to-image pre-trained models.
Furthermore, we observe that our method faces difficulties when the given prompt is complex. 
For example, we show in \autoref{fig:limitation}(b) that the generated images of personalized characters with the prompt "[V1] riding on red bobsleigh, [V2] riding on blue bobsleigh." do not display the kangaroo character.
This issue could be alleviated by optimizing to the specific prompt~\citep{arar2024palp}.
Lastly, although our framework effectively alleviates identity mixing for several subjects, we find that subject dominance becomes stronger as the number of personalized subjects increases. 
For instance, \metabbr may duplicate the same subjects in the generated images, as in \autoref{fig:limitation}(c) 
Adjusting the $\gamma$ scale in our initialization can address subject dominance but may yield image saturation. 
We believe our iterative training framework may potentially address these limitations and can be further developed by applying recent RLHF approaches~\citep{lee2023aligning, fan2024dpok, black2024ddpo, wallace2023dpo, clark2023draft}.

\paragraph{Societal impacts}
Our method allows for the synthesis of realistic images of multiple personalized subjects. 
However, there is a risk that our framework can be misused to generate harmful content for the public or to include subjects that are sensitive to privacy. 
To prevent this, it is necessary to apply measures such as watermarking to the generated images to prevent misuse, as well as protective watermarking specifically for privacy-sensitive subjects.


\end{document}